\documentclass[conference]{IEEEtran}
\IEEEoverridecommandlockouts
\usepackage{cite}
\usepackage{amsmath,amssymb,amsfonts}
\usepackage{algorithmic}
\usepackage{graphicx}
\usepackage{xcolor}

\usepackage{algorithmic}
\usepackage{array}
\usepackage{textcomp}
\usepackage{stfloats}
\usepackage{url}
\usepackage{verbatim}
\def\BibTeX{{\rm B\kern-.05em{\sc i\kern-.025em b}\kern-.08em
    T\kern-.1667em\lower.7ex\hbox{E}\kern-.125emX}}
\usepackage{balance}

\usepackage{rotating}
\usepackage{makecell}
\usepackage{multirow}
\usepackage{multicol}
\usepackage{pifont}
\usepackage{arydshln}
\usepackage{hyperref}
\usepackage{subcaption}

\begin{document}

\title{Deep Insights into Cognitive Decline: A Survey of Leveraging Non-Intrusive Modalities with Deep Learning Techniques}

\author{
    \IEEEauthorblockN{1\textsuperscript{st} David Ortiz-Perez}
    \IEEEauthorblockA{
        \textit{Dept. of Computer Science and Technology} \\
        \textit{University of Alicante}\\
        Alicante, Spain \\
        dortiz@dtic.ua.es}
    \and
    \IEEEauthorblockN{2\textsuperscript{nd} Manuel Benavent-Lledo}
    \IEEEauthorblockA{
        \textit{Dept. of Computer Science and Technology} \\
        \textit{University of Alicante}\\
        Alicante, Spain \\
        mbenavent@dtic.ua.es}
    \and
    \IEEEauthorblockN{3\textsuperscript{rd} Jose Garcia-Rodriguez}
    \IEEEauthorblockA{
        \textit{Dept. of Computer Science and Technology} \\
        \textit{University of Alicante}\\
        Alicante, Spain \\
        jgarcia@dtic.ua.es}
    \and
    \IEEEauthorblockN{4\textsuperscript{th} David Tomás}
    \IEEEauthorblockA{
        \textit{Dept. of Software } \\
        \textit{and Computing Systems} \\
        \textit{University of Alicante}\\
        Alicante, Spain \\
        dtomas@dlsi.ua.es}
    \and
    \IEEEauthorblockN{5\textsuperscript{th} M. Flores Vizcaya-Moreno}
    \IEEEauthorblockA{
        \textit{Unit of Clinical Nursing Research} \\
        \textit{Faculty of Health Sciences} \\
        \textit{University of Alicante}\\
        Alicante, Spain \\
        flores.vizcaya@ua.es}
}

\maketitle

\begin{abstract}
Cognitive decline is a natural part of aging. However, under some circumstances, this decline is more pronounced than expected, typically due to disorders such as Alzheimer’s disease. Early detection of an anomalous decline is crucial, as it can facilitate timely professional intervention. While medical data can help, it often involves invasive procedures. An alternative approach is to employ non-intrusive techniques such as speech or handwriting analysis, which do not disturb daily activities. This survey reviews the most relevant non-intrusive methodologies that use deep learning techniques to automate the cognitive decline detection task, including audio, text, and visual processing. We discuss the key features and advantages of each modality and methodology, including state-of-the-art approaches like Transformer architecture and foundation models. In addition, we present studies that integrate different modalities to develop multimodal models. We also highlight the most significant datasets and the quantitative results from studies using these resources. From this review, several conclusions emerge. In most cases, text-based approaches consistently outperform other modalities. Furthermore, combining various approaches from individual modalities into a multimodal model consistently enhances performance across nearly all scenarios.
\end{abstract}

\begin{IEEEkeywords}
Cognitive Decline, Deep Learning, Data Modality, Non-Intrusive Techniques
\end{IEEEkeywords}

\section{Introduction}

Cognitive abilities play a crucial role in daily life and are fundamental to our interactions with others. These abilities include reasoning, memory functions, attention, and language comprehension and communication~\cite{cognitive}. As individuals age, cognitive function typically undergoes a natural decline, with some abilities, such as processing speed, reasoning, and memory, showing more pronounced deterioration, while others remain relatively unaffected~\cite{age}. Although cognitive decline is a common and natural aspect of aging, some individuals may experience a more accelerated or pronounced deterioration in these abilities due to cognitive disorders, such as dementia~\cite{dementia}. The severity of this deterioration spans a broad spectrum, with Mild Cognitive Impairment (MCI) representing an intermediate stage between typical age-related decline and more severe forms, including different types of dementia~\cite{mci, mci2}. Age-related cognitive decline presents a significant societal challenge, especially given the growing trend toward aging populations~\cite{ageing}. Nonetheless, not all disorders affecting cognitive abilities are associated with age. For instance, aphasia, typically caused by strokes, can impair language functions across age groups~\cite{damasio1992aphasia}.

Recent technological advances, particularly in deep learning, have the potential to revolutionize numerous fields, including healthcare. The integration of automation into medical practice offers promising opportunities to improve the efficiency and productivity of healthcare professionals~\cite{s24092751,zhao2019deep}. In the context of cognitive decline detection, these technologies have demonstrated potential in facilitating early detection~\cite{venugopalan2021multimodal,liu2014early}, enabling timely intervention during the initial stages. Early detection is crucial, as it allows patients to receive professional treatment from the onset. This timely intervention is particularly important, as previous studies have shown that support from doctors, psychiatrists, or psychologists can significantly improve patients' quality of life and slow the progression of cognitive decline, ultimately benefiting their cognitive and behavioral functioning~\cite{early}.

These deep learning methods typically require large amounts of data and can benefit from multiple data sources, each offering a different balance between diagnostic contribution, cost, privacy, and degree of intrusiveness to the patient. Many studies rely on medical data, including Magnetic Resonance Images (MRI) of patients' brains, to diagnose these conditions~\cite{mri1,mri2,mri3,mri4,mri5,mri6,mri7}. While this type of data offers valuable insights into disorders and their impact on patients' brains, it typically requires highly invasive collection methods. Furthermore, these approaches face limitations in terms of feasibility in different contexts. As a result, this study focuses on non-invasive techniques that avoid disrupting patients' daily routines. These alternative approaches often involve the use of video recordings, integrating visual, audio, and textual data. In this context, key priorities include protecting patient privacy, ensuring ease of deployment, and minimizing interference with everyday activities.

Although various reviews and surveys have already addressed cognitive decline detection, most focus primarily on medical imaging, such as MRI or EHR (Electronic Health Records) \cite{pellegrini2018machine,graham2020artificial,choi2018predicting}. Imaging methods have been also explored specifically for certain conditions, including Alzheimer’s or MCI~\cite{grueso2021machine,warren2023functional,ansart2021predicting,mri7,ALBERDI20161}. In parallel, non-intrusive modalities have been used to study Alzheimer’s and Parkinson’s~\cite{10129131,yang2022deep,10.3389/fnagi.2023.1224723}. However, there remains a notable gap in the literature regarding comprehensive studies focused on general cognitive decline detection using non-intrusive data sources. This work aims to address that gap.

In this study, we present a comprehensive survey focused on the early detection of cognitive decline using deep learning methods and non-intrusive data sources. To achieve this, we systematically review current methodologies in this field and introduce the most relevant datasets for training and evaluating the performance of models designed to detect cognitive decline. Additionally, we discuss the effectiveness of each modality and identify the most suitable approaches for leveraging them. Furthermore, we explore various strategies for combining these modalities into multimodal models to exploit their complementary characteristics and features. Employing these modalities enables a more comprehensive understanding of the issue, ultimately leading to improved performance on a variety of tasks. This survey is primarily intended for researchers in machine learning, particularly those developing deep learning models for healthcare applications. However, we also aim to present clinically relevant insights to support interdisciplinary collaboration with healthcare professionals interested in cognitive assessment.

In summary, our main contributions are as follows:

\begin{itemize}
    \item To the best of our knowledge, this is the first survey focused on the general detection of cognitive decline using non-intrusive modalities with deep learning methods. The modalities included in this survey are audio, text, video, and image.
    \item We systematically review the most relevant works in this scope, categorizing them by modality and approach. Additionally, we present the combination of these modalities into various multimodal models.
    \item We discuss the effectiveness and characteristics of each methodology, both for individual modalities and multimodal fusions. This analysis allows us to identify the most promising challenges and future research directions. Furthermore, we highlight the key issues and difficulties related to cognitive decline detection tasks.
    \item We provide a comparative analysis of different approaches and their performance on standard datasets and benchmarks, accompanied by a summary of the findings.
\end{itemize}

The remainder of the paper is structured as follows: Section~\ref{sec:background} introduces background concepts related to cognitive decline disorders and assessments; Section~\ref{sec:review} details the methodology used to select the works included in this study; Section~\ref{sec:dataset} presents the most relevant datasets in this area; Section~\ref{sec:unimodal} explores unimodal approaches, while Section~\ref{sec:multimodal} discusses the combination of these approaches into multimodal frameworks; Section~\ref{sec:discussion} provides a brief discussion of the methodologies examined; finally, Section~\ref{sec:conclusions} presents the conclusions drawn from this research.

\section{Background Concepts}\label{sec:background}

This section defines background concepts essential for understanding how various disorders impact cognitive function in patients. It introduces the most common disorders in this context, along with the primary assessment methods used to evaluate the cognitive state of individuals.

\subsection{Cognitive disorders}

Although cognitive function deteriorates over time, severe deterioration can occur, often due to cognitive disorders. This subsection briefly introduces the main disorders included in this survey that can affect normal cognitive behavior.

\subsubsection{Mild Cognitive Impairment (MCI)} This syndrome is characterized by a more significant cognitive decline than is typical for an individual of a similar age and educational level. However, this decline is less severe than that experienced in dementia and does not significantly interfere with activities of daily living~\cite{gauthier2006mild,petersen2016mild}. MCI often leads to dementia and typically involves difficulties with one or more cognitive functions, such as memory, learning, reasoning, attention, language skills, or loss of interest or motivation~\cite{rosenberg2011neuropsychiatric}.

\subsubsection{Dementia} In contrast to MCI, dementia is characterized by a more pronounced decline in cognitive abilities, leading to significant challenges for patients in their daily lives~\cite{arvanitakis2019diagnosis}. Approximately 70\% of dementia cases are associated with Alzheimer’s disease, known for its characteristic memory loss \cite{geldmacher1996evaluation}. However, Alzheimer's disease also affects other cognitive functions, including attention and language skills. As a result, patients often experience difficulties in articulating their thoughts effectively, having trouble finding the appropriate words \cite{ferris2013language}.

\subsubsection{Aphasia} This condition is caused by damage to the brain regions, usually resulting from a stroke~\cite{damasio1992aphasia}. Aphasia primarily impairs an individual's ability to communicate, leading to difficulties in both speaking and understanding others' speech. Different types of aphasia affect speech in different ways depending on the affected brain regions, resulting in either simplified speech or overly complex sentences and the use of invented words~\cite{kirshner2021aphasia,alexander2008aphasia}.

\subsubsection{Parkinson} Parkinson's disease is a neurodegenerative disease that affects the motor function of people who suffer from it \cite{kalia2015parkinson}. It causes unintended or uncontrollable movements such as shaking, stiffness, and difficulty in balance and coordination. This disease is closely related to dementia, as many people with Parkinson's disease also suffer from it~\cite{aarsland2004rate}. Although Parkinson's affects the nervous system primarily, it is also associated with a severe decline in cognitive function~\cite{aarsland2017cognitive}. 

\subsubsection{Apathy} Although not a disease, apathy is present in a variety of disorders, including dementia~\cite{van2005apathy}. It refers to a lack of interest or emotion and is associated with cognitive decline. Apathy negatively impacts global cognition, verbal fluency, and visual and verbal memory~\cite{montoya2019impact, konstantakopoulos2011apathy}.

\subsection{Cognitive tests}

Some tools allow us to accurately measure the cognitive decline over time and identify significant decreases. It is crucial to closely monitor the cognitive status of patients with cognitive disorders to track progression. The most relevant measures include the Mini-Mental State Examination (MMSE) and the Montreal Cognitive Assessment (MoCA)~\cite{gluhm2013cognitive}.

The MMSE consists of 11 questions covering five domains: orientation, registration, attention/calculation, recall, and language~\cite{arevalo}. A perfect score on the MMSE is 30, and a score of 25 or higher usually indicates normal cognitive function~\cite{shiroky2007can}. The MMSE is primarily used to assess the risk of dementia.

In contrast, the MoCA is designed to detect Mild Cognitive Impairment and early stages of dementia~\cite{trzepacz2015relationship}. The MoCA test includes seven domains: executive/visuospatial function, naming, attention, language, abstraction, recall, and orientation. Similar to the MMSE, the MoCA also has a maximum score of 30, but it is composed of 30 questions \cite{hoops2009validity}.

\section{Review Methodology}\label{sec:review}

\begin{figure}[htpb]
    \centering
    \includegraphics[width=0.4\textwidth]{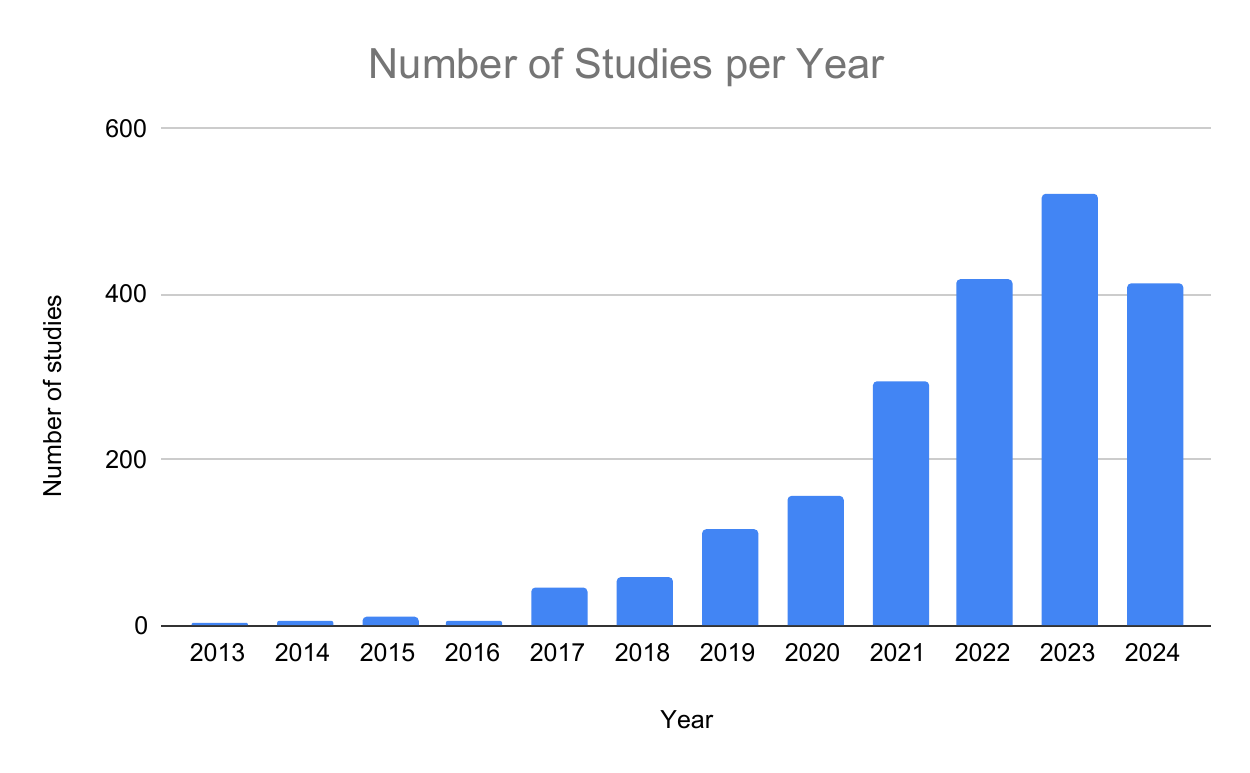}
    \caption{Number of studies identified during the screening process per year.}
    \label{fig:studiesyear}
\end{figure}

\begin{figure}[htpb]
    \centering
    \includegraphics[width=0.4\textwidth]{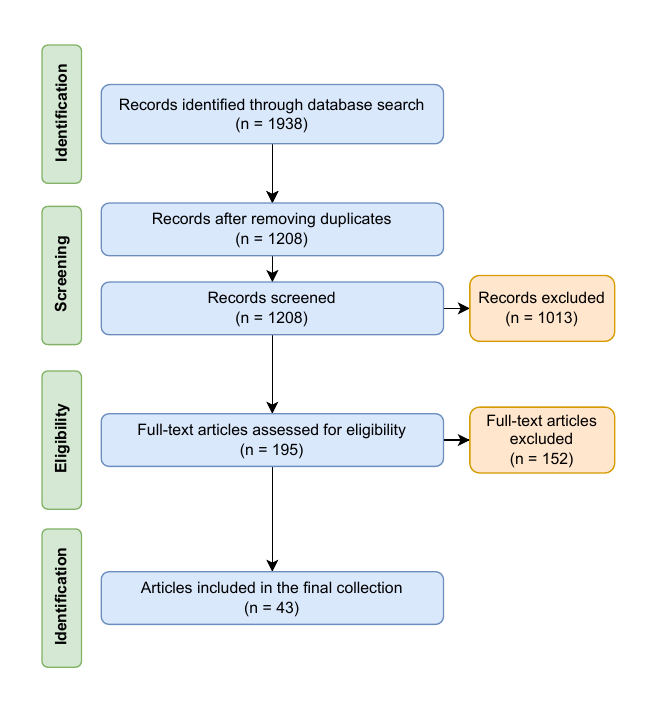}
    \caption{PRISMA flow of the selection process.}
    \label{fig:prisma}
\end{figure}

A systematic review of existing methodologies in the automatic cognitive decline estimation task using Deep Learning techniques was conducted to identify and analyze the most relevant works. This review involved searching for articles in various databases: \emph{PubMed\footnote{\url{https://pubmed.ncbi.nlm.nih.gov/}}}, \emph{Scopus\footnote{\url{http://www.scopus.com/}}}, \emph{Web of Science\footnote{\url{http://www.isiknowledge.com/}}}, and \emph{IEEE Xplore\footnote{\url{http://ieeexplore.ieee.org/}}}. The search was limited to English-language articles published from 2018 to the present. These dates were chosen due to the introduction of the well-known Transformer model in 2017 \cite{vaswani2023attention}, which outperforms many earlier methods. Additionally, there has been a significant increase in publications in our area of interest since that year compared to previous years, as shown in Figure~\ref{fig:studiesyear}. The search was performed on October 1, 2024, and was based on titles, abstracts, and metadata. We followed the Preferred Reporting Items for Systematic Reviews and Meta-Analyses (PRISMA) approach~\cite{moher2010preferred}. An overview of the selection process is depicted in Figure~\ref{fig:prisma}.

The search strategy included several keywords related to cognitive decline, such as ``cognitive decline'', ``aphasia'', ``alzheimer'', ``dementia'', ``MCI'', and ``cognitive impairment''. These keywords have been combined with deep learning terms, such as ``deep learning'', ``CNN'', ``LSTM'', and ``transformer''. Finally, also with the non-intrusive modalities we are targeting, such as ``multimodal'', ``speech'', ``video'', ``RGB'', ``linguistic'', and ``audio''. The final query used over the previously mentioned databases was: \emph{``(``cognitive decline'' OR ``aphasia'' OR ``alzheimer'' OR ``dementia'' OR ``MCI'' OR ``cognitive impairment'') AND ( ``deep learning'' OR ``CNN'' OR ``LSTM'' OR ``transformer'') AND (``multimodal'' OR ``speech'' OR ``video'' OR ``RGB'' OR ``linguistic'' OR ``audio'')''}. 

After screening the different works, the first step was to remove duplicates from the different databases. This was followed by a manual review to assess whether the remaining papers met the inclusion criteria. Specifically, included works were required to propose, train, and test an artificial intelligence model, excluding reviews or surveys that do not train models. Additionally, the selected papers must use the selected non-invasive modalities, excluding any other. The final criterion focused on the detection of cognitive decline, thereby excluding studies that used the specified modalities for assistive purposes rather than for detection. The resulting papers were fully read and scored based on the following quality assessment questions:

\begin{itemize}
    \item Does the work provide implementation details?
    \item Does the work provide detailed information about the models used?
    \item Does the work propose different approaches or modalities?
    \item Does the work provide the source code?
    \item Does the work use standard performance metrics (accuracy, F1-Score, etc.)?
    \item Is the dataset publicly available?
\end{itemize}

By applying these quality assessment questions, we can ensure that the final works included in our study are of high quality and suitable for further analysis. Each study has been evaluated using these questions, with responses scored as follows: Yes (1 point), Partially (0.5 points), and No (0 points). After scoring all candidate studies, an inclusion cutoff threshold of 3.5 out of 6 points was established. Based on this criterion, 43 studies were deemed of sufficient quality and included in the final analysis.

\begin{figure*}
    \centering
    \includegraphics[width=\textwidth]{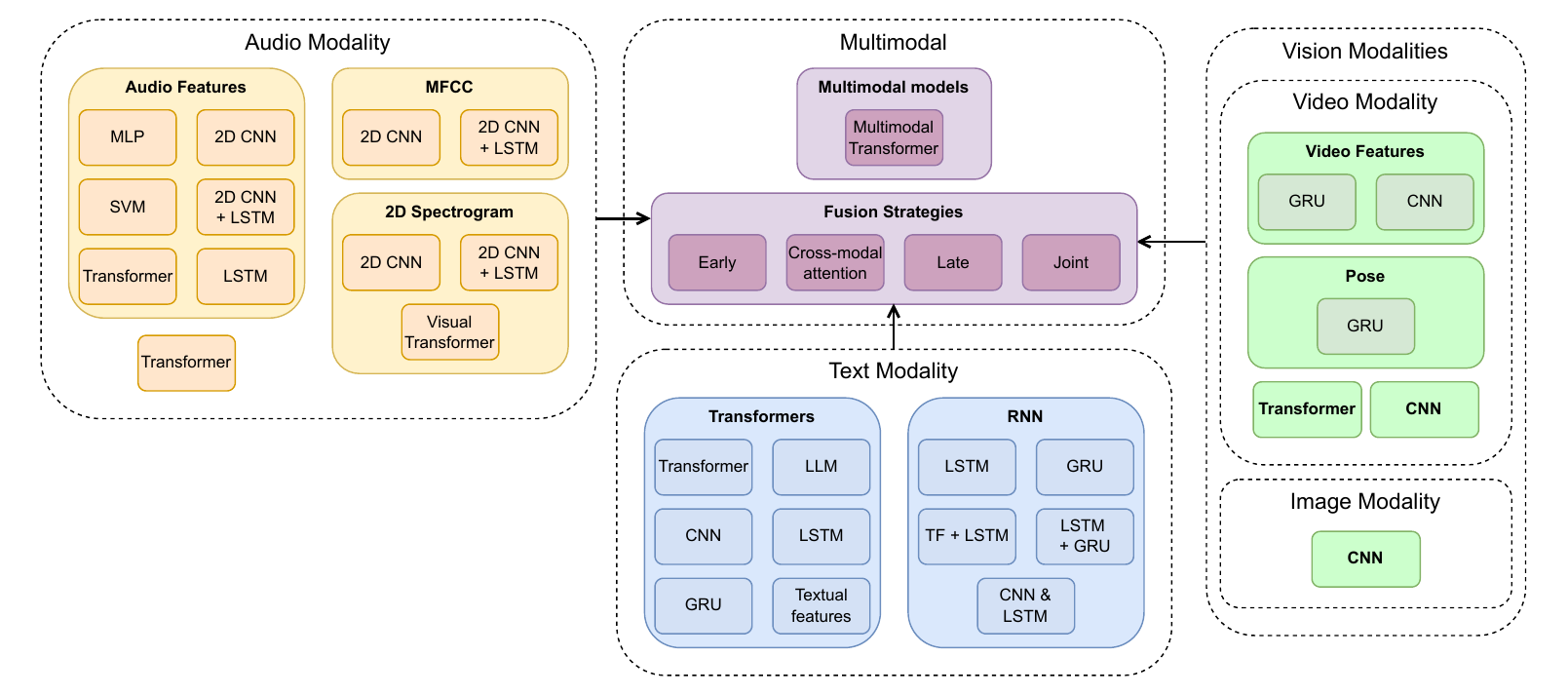}
    \caption{Proposed taxonomy of the reviewed methods, grouped by modality and methodology. Yellow boxes indicate audio methods, blue for text, green for vision (image and video), and purple for multimodal approaches.}
    \label{fig:taxonomy}
\end{figure*}

Finally, the selected studies and their results have been categorized based on the modalities and methodologies employed. We propose a taxonomy that first categorizes methods by the modality used and subsequently by the methodology applied. The modality groups include audio, text, vision (image and video), and multimodal combinations. For the audio modality, several methodologies are employed, including the use of extracted audio features, Transformers, Mel Frequency Cepstral Coefficients (MFCCs), and 2D Spectrograms. These methodologies are combined with deep learning techniques such as Convolutional Neural Networks (CNNs), Long Short-Term Memory (LSTM) networks, Support Vector Machines (SVMs), or Transformer architectures. In the case of the text modality, two primary approaches are distinguished: Transformer-based models and Recurrent Neural Networks (RNNs). These can be further integrated with techniques such as CNNs, Gated Recurrent Units (GRUs), LSTMs, Large Language Models (LLMs), or textual feature extraction. For vision modalities, CNNs are primarily used for image processing, while video analysis often incorporates methods such as video feature extraction, pose estimation, Transformers, or CNNs. Finally, multimodal approaches include the use of multimodal Transformers or fusion strategies, which encompass early fusion, cross-modal attention, late fusion, and joint fusion strategies. Figure~\ref{fig:taxonomy} presents a taxonomy of the reviewed works, illustrating the different approaches.

\section{Datasets}\label{sec:dataset}

The quality and quantity of data utilized in training deep learning models represents a critical factor. Generally, larger and more diverse datasets result in better model performance by improving the model's ability to generalize across a wider range of scenarios. However, in the domain of cognitive impairment detection, the availability of suitable datasets remains a significant challenge. Due to the sensitive nature of data from individuals with cognitive disorders, access to such datasets is often restricted. This section aims to highlight the most relevant and accessible datasets employed in this field. A summary of these datasets can be found in Table~\ref{tab:datasets}:

\begin{itemize}
    \item \textbf{DementiaBank}. The DementiaBank dataset~\cite{lanzi2023dementiabank} is the largest available dataset, comprising 15 corpora with subjects from various regions and languages, including English, German, Mandarin, Spanish, and Taiwanese. The number of samples varies across the different corpora, as do the modalities, which may include video, audio, and text. This dataset consists of subjects with Alzheimer's disease, MCI, and control groups, with over 700 samples.
    
    \item \textbf{DementiaBank Pitt Corpus}. The Pitt Corpus~\cite{becker1994natural} is the most widely used corpus within the DementiaBank project. It consists of 309 English language samples from individuals with Alzheimer’s disease, and 243 samples from cognitively healthy individuals. During the recordings, participants are asked to perform various verbal fluency tasks and to describe a picture, specifically The Cookie Theft Picture~\cite{doi:10.1080/02687039608248419}. These tasks are commonly used in psychological assessments to evaluate cognitive status. Available modalities include both audio and transcribed text.

    \item \textbf{ADReSS}. The Alzheimer’s Dementia Recognition through Spontaneous Speech (ADReSS) dataset~\cite{luz2020alzheimers} is a balanced and higher-quality subset of the Pitt Corpus Cookie Theft Picture description. It is also a challenge dataset used at the Interspeech 2020 conference. This subset has undergone noise removal, voice activity detection, and volume normalization. It includes 78 Alzheimer's patients and 78 control subjects, with data provided in both audio and transcribed text formats.
    
    \item \textbf{ADReSSo}. The Alzheimer’s Dementia Recognition through Spontaneous Speech only (ADReSSo) dataset~\cite{luz2021detecting} is a challenge dataset used at the Interspeech 2021 conference, obtained through fluency and image description tasks. In this case, it is not derived from the Pitt Corpus. The dataset contains audio recordings from 87 patients diagnosed with Alzheimer's disease and 79 control subjects, with no accompanying transcriptions.

    \item \textbf{ADReSS-M}. The Multilingual Alzheimer’s Dementia Recognition through Spontaneous Speech (ADReSS-M) dataset~\cite{luz2023multilingual} is a challenge dataset used at the International Conference on Acoustics, Speech and Signal Processing (ICASSP) 2023 conference. This dataset consists of audio recordings from Greek and English speakers describing pictures and is not derived from the Pitt Corpus. It includes 148 Alzheimer’s patients and 143 controls, with audio being the only modality provided, as it focuses on acoustic features.
    
    \item \textbf{Taukadial}. The Taukadial dataset~\cite{garcia2024connected} is a challenge dataset used at the Interspeech 2024 conference and contains audio recordings from Chinese and English speakers performing three different picture description tasks. It comprises 222 samples from individuals with MCI and 165 control subjects, with only audio data available.

    \item \textbf{AphasiaBank}. The AphasiaBank dataset~\cite{macwhinney2011aphasiabank} comprises various corpora featuring subjects from different regions and languages, including English, Croatian, French, Italian, Mandarin, Romanian, and Spanish. Subjects engage in various conversational tasks with clinicians, with data available in video, audio, and text formats. The dataset size varies across the corpora and includes a total of 180 aphasic patients and 140 control subjects.

    \item \textbf{NCMMSC2021}. The National Conference on Man-Machine Speech Communication (NCMMSC2021) dataset\footnote{\url{https://github.com/lzl32947/NCMMSC2021_AD_Competition}} consists of audio recordings of subjects performing different tasks, including picture descriptions and fluency tasks. The dataset includes 79 subjects with Alzheimer's, 93 with MCI, and 108 control subjects. Data is provided in audio and text modalities, and the dataset is divided into two versions: one containing short conversations and the other containing longer conversations.
    
    \item \textbf{CCC}. The Carolinas Conversation Collection~(CCC)~\cite{PopeDavis2011143161} is a dataset composed of audio conversations. It comprises 400 samples of people with Alzheimer’s and 200 samples for control. Besides the audio modality, textual transcriptions are provided.

    \item \textbf{B-SHARP}. The Brain, Stress, Hypertension, and Aging Research Program (B-SHARP)~\cite{li-etal-2020-analysis} comprises a series of conversations and tasks, including the description of images conducted by a group of subjects. The group is formed by 141 MCI patients and 185 controls. The data modalities are audio and its transcriptions in text.

    \item \textbf{I-CONECT}. The Internet-Based Conversational Engagement Clinical Trial (I-CONECT) dataset~\cite{carr2019successfully,Yu2021,wu2022can} consists of different video recordings from subjects having a conversation. This dataset compromises 100 MCI and 86 control subjects. The data is provided in video and audio modalities.

    \item \textbf{PC-GITA}. The PC-GITA dataset~\cite{orozco-arroyave-etal-2014-new} is composed of audio recordings, where people are asked to perform different fluency and speaking tasks. This dataset is composed of Spanish people, and comprises 50 people with Parkinson's and 50 control people. The available modality is audio.

    \item \textbf{DemCare}. The DemCare dataset~\cite{karakostas2016demcareexperimentsdatasetstechnical} contains video recordings of elderly individuals performing daily life activities. This dataset includes both video and audio modalities. However, since there are no specific speaking tasks, language is not a relevant factor. The subjects in the dataset include both Greek individuals with Alzheimer's and control participants, including a total of 89 subjects.

    \item \textbf{PRAXIS Gesture}. The PRAXIS Gesture dataset~\cite{negin2018praxis} consists of upper-body video recordings of participants performing various gestures. Subjects are asked to repeat each gesture until they complete it correctly. The dataset includes 60 subjects in total, categorized as follows: 29 elderly individuals with normal cognitive function, 2 with amnestic MCI, 7 with unspecified MCI, 2 with vascular dementia, 10 with mixed dementia, 6 with Alzheimer's disease, 1 with posterior cortical atrophy, 1 with corticobasal degeneration, and 2 with severe cognitive impairment (SCI).
    
\end{itemize}

\begin{table*}[htpb]
\centering

\caption{Cognitive disorders estimation datasets. For simplicity, ``V'' denotes Video modality, ``A'' denotes Audio modality, ``T'' denotes Text modality, ``AD'' denotes Alzheimer's disease, ``PAD'' denotes probable Alzheimer's disease, and ``DTP'' denotes different types of dementia.}\label{tab:datasets}
{\renewcommand{\arraystretch}{1.3}

\begin{tabular}{cccccc}
\hline
Dataset & Modalities & Language & Distribution & Task \\
\Xhline{1pt}

\begin{tabular}[c]{@{}c@{}}DementiaBank General \cite{lanzi2023dementiabank}\end{tabular} & \begin{tabular}[c]{@{}c@{}} V, A \& T\end{tabular} & \begin{tabular}[c]{@{}c@{}}Multiple\end{tabular} & \begin{tabular}[c]{@{}c@{}}AD, MCI \& Control: \textgreater 700\end{tabular} & \begin{tabular}[c]{@{}c@{}}Image description\end{tabular} \\

\begin{tabular}[c]{@{}c@{}}DementiaBank Pitt Corpus \cite{becker1994natural}\end{tabular} & \begin{tabular}[c]{@{}c@{}} A \& T \end{tabular} & English & \begin{tabular}[c]{@{}c@{}}AD: 309, Control: 243\end{tabular} & \begin{tabular}[c]{@{}c@{}}Fluency tasks \& Image description\end{tabular} \\

ADReSS \cite{luz2020alzheimers} & \begin{tabular}[c]{@{}c@{}} A \& T\end{tabular} & English & \begin{tabular}[c]{@{}c@{}}AD: 78, Control: 78\end{tabular} & \begin{tabular}[c]{@{}c@{}}Image description\end{tabular} \\

ADReSSo \cite{luz2021detecting} & \begin{tabular}[c]{@{}c@{}} A \end{tabular} & English & \begin{tabular}[c]{@{}c@{}}AD: 87, Control: 79\end{tabular} & \begin{tabular}[c]{@{}c@{}}Fluency tasks \& Image description\end{tabular} \\

ADReSS-M \cite{luz2023multilingual} & \begin{tabular}[c]{@{}c@{}} A \end{tabular} & English \& Greek & \begin{tabular}[c]{@{}c@{}}PAD: 148, Control: 143\end{tabular} & \begin{tabular}[c]{@{}c@{}}Image description\end{tabular} \\

Taukadial \cite{garcia2024connected} & \begin{tabular}[c]{@{}c@{}} A \& T\end{tabular} & English \& Chinese & \begin{tabular}[c]{@{}c@{}}MCI: 222, Control: 165\end{tabular} & \begin{tabular}[c]{@{}c@{}}Image description\end{tabular} \\

\begin{tabular}[c]{@{}c@{}}AphasiaBank \cite{macwhinney2011aphasiabank}\end{tabular} & \begin{tabular}[c]{@{}c@{}} V, A \& T \end{tabular} & \begin{tabular}[c]{@{}c@{}}Multiple\end{tabular} & \begin{tabular}[c]{@{}c@{}}Aphasia: 180, Control: 140\end{tabular} & \begin{tabular}[c]{@{}c@{}}Conversation \& Image description\end{tabular} \\

\begin{tabular}[c]{@{}c@{}}NCMMSC2021 \end{tabular} & \begin{tabular}[c]{@{}c@{}} A \& T \end{tabular} & Mandarin & \begin{tabular}[c]{@{}c@{}}AD: 79, MCI: 93, Control: 108\end{tabular} & \begin{tabular}[c]{@{}c@{}}Image description\end{tabular} \\

\begin{tabular}[c]{@{}c@{}}CCC \cite{PopeDavis2011143161} \end{tabular} & \begin{tabular}[c]{@{}c@{}} A \& T \end{tabular} & English & \begin{tabular}[c]{@{}c@{}}AD: 400, Control: 200\end{tabular} & \begin{tabular}[c]{@{}c@{}}Conversation\end{tabular} \\

\begin{tabular}[c]{@{}c@{}}B-SHARP \cite{li-etal-2020-analysis} \end{tabular} & \begin{tabular}[c]{@{}c@{}} A \& T \end{tabular} & English & \begin{tabular}[c]{@{}c@{}}MCI: 141, Control: 185\end{tabular} & \begin{tabular}[c]{@{}c@{}}Conversation \& Image description\end{tabular} \\

\begin{tabular}[c]{@{}c@{}}I-CONECT \cite{carr2019successfully,Yu2021,wu2022can} \end{tabular} & \begin{tabular}[c]{@{}c@{}} V \& A \end{tabular} & English & \begin{tabular}[c]{@{}c@{}}MCI: 100, Control: 86\end{tabular} & \begin{tabular}[c]{@{}c@{}}Conversation \end{tabular} \\

\begin{tabular}[c]{@{}c@{}}PC-GITA \cite{orozco-arroyave-etal-2014-new} \end{tabular} & \begin{tabular}[c]{@{}c@{}} A \end{tabular} & Spanish & \begin{tabular}[c]{@{}c@{}}Parkinson: 50, Control: 50\end{tabular} & \begin{tabular}[c]{@{}c@{}}Speaking tasks\end{tabular} \\

\begin{tabular}[c]{@{}c@{}}DemCare \cite{karakostas2016demcareexperimentsdatasetstechnical} \end{tabular} & \begin{tabular}[c]{@{}c@{}} V \& A \end{tabular} & - & \begin{tabular}[c]{@{}c@{}}AD, MCI \& Control: 89\end{tabular} & \begin{tabular}[c]{@{}c@{}}Daily life activities\end{tabular} \\

\begin{tabular}[c]{@{}c@{}}PRAXIS Gesture \cite{negin2018praxis} \end{tabular} & \begin{tabular}[c]{@{}c@{}} V \end{tabular} & - & \begin{tabular}[c]{@{}c@{}}DTP: 22, MCI: 9, Control:29\end{tabular} & \begin{tabular}[c]{@{}c@{}}Gestures\end{tabular} \\

\hline
\end{tabular}}
\end{table*}

In this study, we have included basic metrics such as accuracy and F1-Score for the detection of cognitive impairment in the aforementioned datasets. Additionally, we have reported the primary metric used for a regression task. This task involves predicting a subject's cognitive score based on the MMSE, and is measured by the Root Mean Square Error (RMSE). These metrics are applied in the selected works to ensure consistency in performance evaluation.

\section{Single Modality}\label{sec:unimodal}

In the context of cognitive estimation, different modalities provide varied insights into the status of the individual. This section introduces unimodal analysis of audio, text, image, and video data using a variety of approaches depending on the context. It is important to note that many of the presented works focusing on a single modality also propose approaches for other modalities or suggest combining multiple single-modality approaches into a multimodal framework, as will be shown in Section \ref{sec:multimodal}. While a particular work may demonstrate robust performance in a given modality, it may not necessarily exhibit the same level of proficiency in other modalities.

\subsection{Audio Modality}

The audio modality represents how humans perceive and process auditory information from the environment, and is typically captured by microphones. This modality is easy to obtain and, in the case of recording speech, can provide valuable insights into how we speak, as well as facilitate the extraction of other crucial modalities, such as text by a transcription process. Besides text transcription, there is also a wide range of features that can be representative, particularly in how we express our thoughts. The way we express our ideas can be very significant in cognitive estimation, as people with more severe decline tend to struggle to find the correct and appropriate words, leading to long pauses, hesitation, and the use of incorrect words among other issues~\cite{egas2022automatic,themistocleous2020voice}. Furthermore, speech is frequently employed to identify an individual’s emotional state, a process known as Speech Emotion Recognition (SER)~\cite{abbaschian2021deep}. SER typically uses a range of speech features, including prosody, pitch, and rhythm to accurately detect emotions such as happiness or fear~\cite{koolagudi2012emotion,wani2021comprehensive}. Some studies have shown that cognitive disorders negatively affect emotional mood~\cite{ortiz2023deep,code1999emotional}, making it another relevant source of information for this task.

This section presents an overview of the different methods employed for cognitive estimation using the audio modality, summarized in Table \ref{tab:audio}. This table is organized by first distinguishing the disorder being treated, followed by the dataset used for training and testing. Additionally, it provides the publication year, the methodology that achieved the best performance in each study, and the key metrics previously mentioned: accuracy, F1-Score, and RMSE. The best results for each dataset are highlighted in bold. The main techniques employed are detailed in the following subsections, including 2D Spectrograms, MFCCs, Audio Features, and Transformers methods.

\begin{table*}[htpb]
    \centering
    \caption{Performance comparison of audio-based methods over different cognitive disorders and datasets. For simplicity, ``MFCC'' denotes Mel-Frequency Cepstral Coefficients.}
    \begin{tabular}{c|c|cccccc}
        \hline
        \multirow{2}{*}{Disorder} & \multirow{2}{*}{Dataset} & \multirow{2}{*}{Work} & \multirow{2}{*}{Year} & \multirow{2}{*}{Method} & \multicolumn{3}{c}{Score}  \\
        & & & & & Acc & F1 & RMSE \\
        \Xhline{1pt}
        \multirow{20}{*}{Alzheimer} & \multirow{7}{*}{ADReSSo} & \cite{app13074244}  & 2023 & 2D Spectrograms &  \textbf{78.90} & - & - \\
        & & \cite{Cui_2023} & 2023 & Transformers &  - & \textbf{80.5} & - \\
        & & \cite{wang2021modular} & 2021 & Audio Features & 75.30 & 76.00 & - \\
         & & \cite{ying2023multimodal} & 2023 & Transformers & 71.20 & 73.10 & - \\
         & & \cite{bang2024alzheimer} & 2024 & Transformers & 69.01 & 70.39 & - \\
         & & \cite{DBLP:journals/corr/abs-2106-15684} & 2021 & Audio Features & 68.00 & - &  \textbf{6.03} \\
         & & \cite{10307469} & 2023 & 2D Spectrograms &  62.10 & 60.7 & - \\

         \cline{2-8}
         & \multirow{8}{*}{ADReSS} & \cite{hlédiková2022data} & 2022 & 2D Spectrograms & \textbf{74.00} & - & - \\
         & & \cite{interspeech2020} & 2020 & Audio Features & 72.92 & - & \textbf{5.07} \\
         & & \cite{pompili2020inesc} & 2020 & Audio Features & 72.73 & \textbf{72.73} & - \\
         & & \cite{cummins2020comparison} & 2020 & 2D Spectrograms & 70.80 & 70.80 & -\\
         & & \cite{mahajan2021acoustic} & 2021 & Audio Features & 68.75 & - & - \\
         & & \cite{10.3389/fcomp.2021.624683} & 2021 & 2D Spectrograms & 66.67 & - & - \\         
         & & \cite{ILIAS2023101485} & 2023 & 2D Spectrograms & 65.00 & 69.76 & - \\
         & & \cite{meghanani2021exploration}  & 2021 & MFCC & 64.58 & - & 6.24 \\

         \cline{2-8}
         & \multirow{3}{*}{\makecell{DementiaBank \\ Pitt Corpus}} & \cite{9413566} & 2021 &  Audio Features & \textbf{82.59} & \textbf{82.94} & - \\ 
         & & \cite{ORTIZPEREZ2023126413} & 2023 & 2D Spectrograms &  73.49 & - & - \\
         & & \cite{krstev2022multimodal} & 2022 & 2D Spectrograms & 73.00 & 73.00 & - \\

         \cline{2-8}
         & Own dataset & \cite{escobar2023deep}  & 2023 & Transformers & \textbf{88.50} & \textbf{88.30} & - \\

        \hline
         \multirow{2}{*}{MCI} & Taukadial & \cite{ortizperez2024cognitive} & 2024 & Audio Features & \textbf{70.75} & \textbf{76.22} & \textbf{3.11} \\
         \cline{2-8}
         & I-CONECT & \cite{POOR2024109199} & 2024 & Transformers & \textbf{50.42} & \textbf{54.02} & - \\

         \hline
         \multirow{3}{*}{\makecell{Alzheimer \\ \& MCI}} &  NCMMSC2021 L & \cite{ying2023multimodal} & 2023 & Transformers & \textbf{85.70} & \textbf{85.70} & - \\
         \cline{2-8}
         &  NCMMSC2021 S & \cite{ying2023multimodal} & 2023 & Transformers & \textbf{81.70} & \textbf{81.00} & - \\
         \cline{2-8}
         & Private dataset & \cite{9747054} & 2022 & Audio Features & \textbf{82.18} & - & - \\

         \hline
         Parkinson &  PC-GITA speech & \cite{9426437} & 2021 & Audio Features &  \textbf{68.56} & - & - \\

         \hline
         Aphasia & \makecell[c]{Mandarin \\ AphasiaBank} & \cite{10054791} & 2022  & 2D Spectrograms & - & - & \textbf{3.53} \\
         
         \hline
    \end{tabular}
    \label{tab:audio}
\end{table*}

\subsubsection{\textbf{2D Spectrograms Methods}}

The audio recordings are typically represented by waveforms. These waveforms show the amplitude of the signal as a function of time. The audio modality is considered a one-dimensional data type, consisting of a sequence of samples taken at specific and discrete time intervals. However, there are also multi-channel audio representations, such as stereo, which combine two mono (single-channel) signals, thereby creating a two-dimensional data structure.

One of the most significant properties of audio is its sample rate, which indicates the number of samples taken per second of audio. The employment of one-dimensional representations in the field of deep learning can bring many benefits. One such benefit is the simplicity of usage, as the raw data is utilized without further preprocessing. This eliminates the need for additional processing, resulting in low computational cost and training time. Nevertheless, despite the existence of some works that employ raw one-dimensional representations for audio classification, alongside one-dimensional CNNs~\cite{9659979,ABDOLI2019252}, this is not the prevailing approach~\cite{zaman2023survey}. Instead, they are often converted into two-dimensional data, such as spectrograms~\cite{costa2017evaluation}. 

A spectrogram is a visual representation of an audio recording that illustrates the frequency content of the recording over time. This combined time-frequency representation enables deep learning models to capture temporal patterns while also analyzing the frequency spectrum. Consequently, these models can extract more detailed information from audio recordings than from raw waveforms~\cite{satt2017efficient,zeng2019spectrogram}. By showing frequency changes over time, spectrograms make it easier to distinguish sounds with similar time-domain characteristics but different frequency-domain characteristics. Additionally, these representations are closer to how humans perceive sounds based on frequency and temporal changes, which are effectively captured in spectrograms~\cite{howard2013acoustics,moore2012introduction}.

The use of spectrogram representations offers numerous advantages in deep learning, particularly for general tasks. However, when it comes to human speech, other transformations can be even more beneficial. One such transformation is the Mel Spectrogram \cite{10.1121/1.1915893}, a type of spectrogram mapped to the Mel Scale. The Mel Scale was designed to represent frequencies in a way that better reflects how humans perceive sound. Humans do not perceive all frequencies equally; we are more sensitive to differences in lower frequencies than in higher ones. For instance, it's much easier for us to distinguish between sounds at 100 Hz and 200 Hz than between 10,000 Hz and 10,100 Hz. Given this perceptual difference, Mel Spectrograms are especially valuable for processing human speech, as they more accurately capture the nuances of how we hear sounds~\cite{arias2021multi,8678825}.

\begin{figure}[htpb]
    \centering
    \includegraphics[width = 0.48\textwidth]{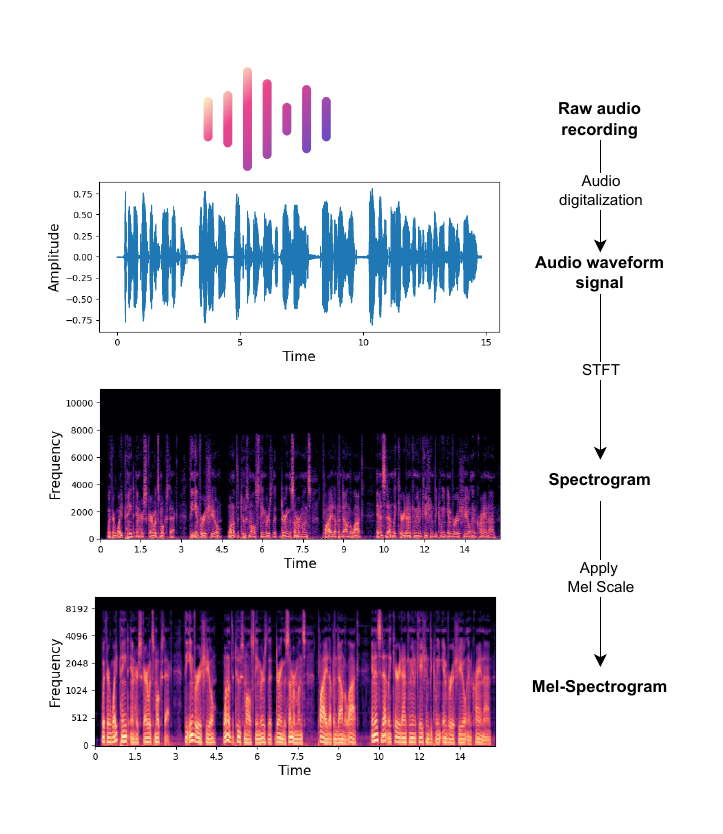}
    \caption{Overview of the process for converting audio signals into a Mel Spectrogram.}
    \label{fig:audioprocessing}
\end{figure}

In order to generate spectrograms and Mel Spectrograms from raw waveforms, it is necessary to apply specific transformations. One crucial step involves the application of the Fourier transform to convert the audio signal into a frequency magnitude representation. The magnitude of each frequency component indicates its influence on the overall audio sample. However, this initial transformation lacks temporal data, which is essential for audio processing as it contains significant information. To ensure the retention of temporal information throughout the audio sample, a windowing technique is employed. This technique segments the audio into overlapping windows over time. Each windowed segment undergoes Fast Fourier Transform (FFT) to convert the signal from the time domain to the frequency domain. The FFT computes the Discrete Fourier Transform (DFT) of a sequence. This process, applied to each segment, is known as Short-Time Fourier Transform (STFT). Combining these processed transformations across all segments yields the final spectrogram, providing a two-dimensional representation of the audio that includes both time and frequency information. To obtain a Mel Spectrogram, another transformation must be performed to scale the frequencies to the previously mentioned Mel scale. Figure~\ref{fig:audioprocessing} summarizes the transformations to obtain the spectrogram and Mel Spectrogram from a raw audio recording.

Once all the features are extracted into a two-dimensional image representation of the audio signal, image-based models can be used for audio classification tasks. Vision models can extract the most relevant features of an image \cite{jiao2019survey,minaee2021image}. In the case of the widely used classical CNNs, a set of different kernels and pooling layers are applied to an image to obtain more complex features and classify the image \cite{gu2018recent,o2015introduction}. These CNNs have been the traditional state-of-the-art method for image processing \cite{li2021survey}. However, the advent of well-known Transformer-based architectures is replacing them.

Transformers \cite{vaswani2023attention}, which were originally designed for text processing, have migrated to other modalities due to their exceptional performance in the text modality field \cite{DBLP:journals/corr/abs-1810-04805,DBLP:journals/corr/abs-1910-01108}. Among the different modalities where Transformers have migrated, we can distinguish video \cite{DBLP:journals/corr/abs-2103-15691}, audio \cite{DBLP:journals/corr/abs-2006-11477}, and image \cite{DBLP:journals/corr/abs-2010-11929}. The combination of both extracting spectrograms, specifically Mel Spectrograms, and using them to feed a CNN has been employed in many fields, for example, in the analysis of music timbre \cite{pons2017timbre}. Additionally, the use of vision Transformer models with the combination of spectrograms has been applied in other fields, such as general audio classification \cite{zhang2022spectrogram} and even Transformers specialized for spectrograms \cite{gong2021ast}. In addition, the Whisper model\cite{radford2022robust}, which is the state-of-the-art model for audio transcription, relies on Transformer architecture. This model also uses spectrograms for processing audio, incorporating convolutional layers to create an encoding for the subsequent Transformer layers.

The cognitive estimation task is not an exception regarding these types of methods. The best results have been obtained through the use of Mel Spectrograms. Despite requiring a preprocessing step to convert the audio signal into a spectrogram, which slightly increases computational time, this process often yields improved performance. For instance, in the work of \cite{escobar2023deep}, using spectrograms and 2D CNNs improved Alzheimer's disease detection accuracy from 72.6\% with raw waveforms and 1D CNNs to 84.4\% using Mel Spectrograms and 2D CNNs, resulting in a relative improvement of 16.3\% In addition, \cite{cummins2020comparison} achieved a performance of 70.8\% in both accuracy and F1-Score in the Alzheimer recognition task using the ADReSS dataset. In this study, the authors first preprocess the audio signal by applying a Mel Spectrogram process with a window size of 25 ms, retaining only segments of eight or sixteen seconds from the full recording. A distinctive feature of this method is the employment of a Siamese network \cite{bertinetto2016fully}, inspired by various studies utilizing this model for health applications \cite{boelders2020detection}, speech emotion recognition \cite{lian2018speech}, and speech impairment detection \cite{wang2019child}. These Siamese networks learn through contrastive learning applied to two separate models, which, in this study, were CNNs. This contrastive learning approach pulls together segments of the same class (Alzheimer's or Control) and separates segments of different classes. Subsequently, the embeddings obtained from each model serve as inputs to another and final CNN for classification. The downside of this method is that the application of contrastive learning restricts the experiment to classification and does not provide information for regression tasks. Additionally, the authors experimented with feeding a CNN the raw audio signal, but the use of Mel Spectrograms and Siamese networks resulted in a 4.1\% improvement. In this work, the authors trained their own CNN and did not use any pre-trained models.

Another notable work using Mel Spectrograms is presented in \cite{ORTIZPEREZ2023126413,10.1007/978-3-031-18050-7_25} and \cite{krstev2022multimodal}, both focusing on the detection of Alzheimer's using the DementiaBank Pitt Corpus. In the case of \cite{ORTIZPEREZ2023126413}, Mel Spectrograms were combined with pre-trained CNNs DenseNet \cite{DBLP:journals/corr/HuangLW16a}, MobileNet \cite{DBLP:journals/corr/HowardZCKWWAA17}, and ResNet \cite{DBLP:journals/corr/HeZRS15}, with the best results achieved using pre-trained DenseNet (accuracy of 73.49\%). In the case of \cite{krstev2022multimodal}, the proposed pre-trained CNNs included ResNet18 \cite{DBLP:journals/corr/HeZRS15}, ResNet34 \cite{DBLP:journals/corr/HeZRS15}, ResNet50 \cite{DBLP:journals/corr/HeZRS15}, SqueezeNet \cite{iandola2016squeezenet}, and VGG16 \cite{simonyan2015deep}, with the best results obtained using VGG16, achieving an accuracy of 66\% and an F1-Score of 62\%. When combined with demographic data from patients, the accuracy and F1-Score both improved to 73\%.

In addition to the use of CNNs to extract features from images, there are also studies where the embeddings obtained from these CNNs are fed into a LSTM model. The LSTM model is a specific type of recurrent neural network that enhances the capabilities of previous recurrent models by integrating long-term dependencies for sequential data \cite{SHERSTINSKY2020132306}. The work of \cite{app13074244} employs this method to detect Alzheimer's using the ADReSSo dataset, achieving an accuracy of 78.9\%. In this study, the employed model is the pre-trained VGG16 model. The same method has been used in the work of \cite{10054791} to detect the severity of Aphasia using the Mandarin AphasiaBank dataset. In this work the pre-trained ResNet model has been used as the CNN. This application to Aphasia achieved an RMSE score of 3.53 on the cognitive score of the subjects. This study also proposes the accuracy in determining each level of severity of Aphasia, but not for the differentiation between aphasic and normal cognitive subjects.

In the work proposed by S. Siddhant et al. \cite{10307469}, they propose the vision Transformer MVITV2 \cite{li2022mvitv2} to analyze the generated Mel Spectrogram. They also provide an ablation study comparing the results with other previously mentioned CNNs, such as VGG19, ResNet-101, and DenseNet-161. The use of this Transformer-based model outperformed the best result obtained from CNNs by 3\%, achieving an accuracy of 62.1 and an F1-Score of 60.7, demonstrating the capabilities of this type of network for this task.

There are also similar works over the ADReSS dataset. In another study by I. Loukas et al. \cite{ILIAS2023101485}, the performance of the vision Transformer ViT \cite{DBLP:journals/corr/abs-2010-11929} is compared with various pre-trained CNNs, including GoogLeNet \cite{szegedy2014going}, ResNet50 \cite{DBLP:journals/corr/HeZRS15}, WideResNet-50, \cite{zagoruyko2017wide} AlexNet \cite{NIPS2012_c399862d}, SqueezeNet \cite{iandola2016squeezenet}, DenseNet-201 \cite{DBLP:journals/corr/HuangLW16a}, MobileNetV2 \cite{DBLP:journals/corr/abs-1801-04381}, MnasNet1 \cite{DBLP:journals/corr/abs-1807-11626}, ResNeXt-50 \cite{DBLP:journals/corr/HeZRS15}, VGG16 \cite{simonyan2015deep}, and EfficientNet-B2 \cite{DBLP:journals/corr/abs-1905-11946}. The performance of this vision transformer outperformed every single CNN model by at least 2\% in accuracy, achieving up to 65\% in accuracy and a 69.76\% F1-Score. In addition, the work in \cite{10.3389/fcomp.2021.624683} includes an ablation study comparing the performance of CNNs and vision Transformers. In this study, the CNNs tested were MobileNet and YamNet, while the vision Transformer model used was Speechbert \cite{DBLP:journals/corr/abs-1910-11559}. Those CNNs had as input MFCCs, a method which will be detailed in the following subsection. The Speechbert model using MFS outperformed the best CNN, MobileNet, by more than 7\% in accuracy, achieving up to 66.67\% in accuracy.

In the work \cite{hlédiková2022data}, authors propose the use of the Audio Spectrogram Transformer model \cite{gong2021ast}, fed with a Mel Spectrogram, which achieved an accuracy of 74\% over the ADReSS dataset.

\subsubsection{\textbf{MFCCs Methods}}

\begin{figure}[htpb]
    \centering
    \includegraphics[width=0.48\textwidth]{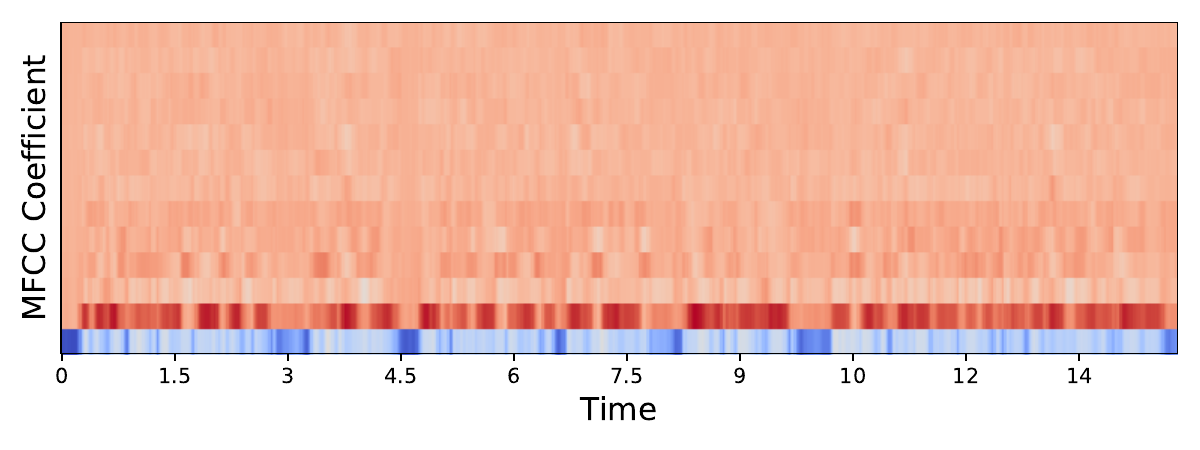}
    \caption{Example of an MFCC representation, obtained from the Mel Spectrogram in Figure \ref{fig:audioprocessing}.}
    \label{fig:mfcc}
\end{figure}

Further preprocessing of audio samples can be performed to obtain MFCCs. This represents a more compact representation of the audio sample, attempting to convey the overall shape of a spectral envelope \cite{gupta2013feature}. In addition to mapping frequencies to the Mel scale, MFCCs transform these features into the cepstral domain, which is useful for a variety of audio and speech processing tasks \cite{muda2010voice,tiwari2010mfcc}. This methodology has also been applied in the health field for tasks such as heart sound classification \cite{deng2020heart} and automatic depression detection \cite{rejaibi2022mfcc}. However, this representation requires additional processing steps in comparison to Mel Spectrograms. This representation is computed by applying a logarithm function to a Mel Spectrogram. After this logarithm procedure, a Discrete Cosine Transform (DCT) is applied to decorrelate these features and reduce their dimensionality. An example of this representation can be seen in Figure \ref{fig:mfcc}.

The primary benefit of using this methodology with Mel Spectrograms is the more compact information and lower dimensionality of data. Additionally, the main idea of this method is to keep the most relevant data regarding human speech tasks. However, the main disadvantage is the need for additional processing steps and the potential loss of information during the discretization of the representation.

In the domain of cognitive estimation tasks, the study that achieved the most favorable results when employing MFCCs was conducted by A. Meghanani et al. \cite{meghanani2021exploration}. In this research, the effectiveness of both Log Mel Spectrograms and MFCCs was investigated using a combination of CNNs and LSTMs for feature classification. The experiments revealed that combining MFCCs with CNNs and LSTMs produced superior accuracy results, outperforming the Log Mel Spectrogram model by 6\%, achieving up to 64.58\% accuracy. For the Log Mel Spectrograms, employing a pre-trained ResNet model improved accuracy to 62.50\%, but it is not sufficient to improve the results obtained using MFCCs. Nevertheless, in the regression task, the Log Mel Spectrograms demonstrated the most optimal performance, achieving an RMSE of 5.9.

\subsubsection{\textbf{Audio Features Methods}}

When listening to speech, it is possible to derive several statistical acoustic features, including frequency, jitter, and shimmer. These values can be obtained using a variety of toolkits that process audio signals and produce different sets of features. Among the most relevant toolkits for this extraction are OpenSMILE \cite{10.1145/1873951.1874246,eyben2013recent}, COVAREP \cite{degottex2014covarep}, Kaldi \cite{povey2011kaldi}, and OpenDBM \footnote{\url{https://github.com/AiCure/open_dbm}}. These toolkits not only provide the previously mentioned features but also other relevant ones, such as MFCCs, offering valuable data for analysis. These features have been applied in various speech-related fields, including emotion recognition \cite{nimitsurachat2024audio}. For instance, the CMU-MOSEI dataset \cite{zadeh2018multimodal} provides audio modality in vectors processed by the COVAREP software. Additionally, these methods have been utilized for tasks such as bird sound classification \cite{qian2015bird} and dysarthric speech classification \cite{narendra2018dysarthric}. The employment of this methodology brings many benefits, as it extracts numerous features, including those previously mentioned. However, it also has disadvantages, such as the significant time required for preprocessing the audio to obtain these features. Some of the set of features available from these toolkits include:

\begin{itemize}
    \item Emobase \cite{10.1145/1873951.1874246}: The representation includes MFCCs, fundamental frequency (F0), F0 envelope, line spectral pairs (LSP), and intensity features. It contains up to 988 features. This feature set can be obtained though the use of OpenSMILE toolkit.
    \item IS10\_linguistic \cite{schuller2010interspeech}: This is an expanded and more recent iteration of the previous Emobase feature set. The feature set includes 16 different types of Low-Level Descriptors (LLDs), Pulse Code Modulation (PCM) loudness, eight log Mel frequency bands, eight line spectral pairs, MFCCs, frequency, F0 envelope, voicing probability, jitter, shimmer, and other features. LLDs include logarithmic harmonicto-noise ratio, voice quality features, Viterbi smoothing for F0, spectral harmonicity, and psycho-acoustic spectral sharpness. This representation yields 1,582 features. This feature set can be obtained though the use of OpenSMILE toolkit.
    \item eGeMAPS \cite{7160715}: This feature set reduces other feature sets to a basic set of 88 features. The feature set comprises seven LLDs, including spectral parameters of MFCC and spectral flux, as well as frequency-related parameters of formant bandwidth. This feature set can be obtained though the use of OpenSMILE toolkit.
    \item ComParE 2016 \cite{schuller16_interspeech}: This set comprises 6,373 features and may be regarded as an extended version of the IS10\_linguistic feature set. It compromises energy, spectral, MFCC, voicing related LLDs, and statistical functionals. This feature set can be obtained though the use of OpenSMILE toolkit.
    \item VGGish \cite{DBLP:journals/corr/HersheyCEGJMPPS16}: This feature set has been obtained from training a CNN with an Audio Set \cite{gemmeke2017audio} for audio classification. The feature set comprises 128 dimensions, which are the embeddings obtained from the pre-trained CNN, with each dimension extracted from audio segments of 960 milliseconds in length.
    \item I-Vector \cite{5545402}: These are statistical speaker representation vectors. These representations have been used in the health domain, including for Parkinson’s disease and dysarthric speech recognition \cite{hauptman2019identifying,laaridh2017automatic}. This feature set are obtained from statistical modeling techniques, compromising 400 features and can be obtained though the use of Kaldi toolkit.
    \item X-Vector \cite{snyder2017deep}: These are discriminative deep neural network-based speaker embeddings that have demonstrated superior performance in speaker and language recognition tasks, outperforming I-Vector \cite{snyder2018spoken,snyder2018x}. This feature set are obtained from the embeddins of pretrained models compromising 512 features can be obtained though the use of Kaldi toolkit.
    \item COVAREP: This set of 79 features encompasses a range of characteristics, including prosodic features, voice quality features, and spectral features (MFCCs, harmonic and phase distortion means and deviations). This feature set can be obtained though the use of COVAREP toolkit.
    \item OpenDBM: The features provided include voice intensity, MFCCs, voice prevalence, shimmer, jitter, pause characteristics, different noises, and frequency information. The feature set compromises 43 features and can be obtained though the use of OpenDBM toolkit.
\end{itemize}

Regarding the cognitive estimation task, numerous studies have achieved promising results through the utilization of this methodology. For instance, in \cite{mahajan2021acoustic} these features were employed to train an MLP and a GRU. In this work, the performance of Emobase, ComParE 2016, and eGeMAPS was evaluated, with Emobase features demonstrating the most favorable outcomes. Furthermore, the combination of these features with the means of the values across the entire recording yielded further performance improvements. The final results demonstrated an accuracy of up to 68.75\% on the ADReSS dataset. Similarly, those features were combined with an MLP in the work proposed by D. Ortiz-Perez et al. \cite{ortizperez2024cognitive}. In this case, the features with the best outcomes come from the OpenDBM toolkit, outperforming the ones obtained from OpenSMILE. This approach yielded an accuracy of up to 70.75\%, an F1-Score of 76.22\%, and an RMSE score of 3.11 on the Taukadial dataset over chinese and english speakers.

These features are also employed in combination with LSTMs, as seen in \cite{DBLP:journals/corr/abs-2106-15684}, where COVAREP features were used on the ADReSSo dataset, achieving up to 68\% accuracy and an RMSE score of 6.03. Additionally, as discussed in the Spectrogram section, features can be processed with CNNs to extract features and an LSTM to establish temporal relations, as demonstrated in \cite{interspeech2020}. In this study, the performance of eGeMAPS, ComParE 2016, and VGGish features was evaluated, with VGGish features yielding significantly better results. The results achieved were 72.92\% accuracy and an RMSE score of 5.07 on the ADReSS test set. The study proposed in \cite{9426437} used their own extracted audio features combined with a CNN for the recognition of Parkinson's disease on the Parkinson PC-GITA speech dataset, obtaining an accuracy of 68.56\%.

There are different works, such as those proposed by L. Zhaoci et al. \cite{9413566} and S. Zhengyan et al. \cite{9747054}, where both use pre-trained models for Automatic Speech Recognition (ASR) and their intermediate hidden states in combination with CNNs and LSTMs to detect Alzheimer's. In \cite{9413566}, this methodology achieved an accuracy of 82.59\% and an F1-Score of 82.94\% on the DementiaBank Pitt Corpus dataset. In \cite{9747054}, the best result was an accuracy of 82.18\% on their own dataset, detecting both Alzheimer's and MCI.

There are other studies, such as the one proposed by N. Wang et al. \cite{wang2021modular}, where the best performance was achieved using X-Vector, slightly improving upon the second-best results and notably improving over features such as Emobase and IS10\_linguistic. In this work, X-Vector features were used in combination with a Transformer encoder, achieving up to 75.3\% accuracy and 76\% F1-Score on the ADReSSo dataset. Similarly, in \cite{pompili2020inesc}, the best results were obtained using X-Vector, which outperformed I-Vector, achieving up to 72.73\% in both accuracy and F1-Score on the ADReSS dataset.

\subsubsection{\textbf{Transformer Methods}}

As previously stated, Transformer models are employed in different modalities, including audio. This architecture typically comprises two principal components: an encoder and a decoder. The encoder processes the given input, resulting in embeddings, which serve as a representation of data. After being processed, embeddings capture the most relevant information. With this representation, the decoder part of the Transformer generates new data based on these representations \cite{lin2022survey,han2022survey}. In the problem presented in this study, we are addressing a classification task for identifying a disorder and a regression task for estimating cognitive state. Therefore, we will focus exclusively on the encoder component, which is responsible for obtaining the data representation and not for generating new data. This representation allows for the detection of either a disorder or the cognitive state.

The main advantage of Transformers over previous methods lies in the attention mechanism, which is a key component of this architecture. This mechanism assigns weights to different parts of a sequence based on their importance relative to other words, enabling the processing of sequences without relying on a strict sequential order \cite{vaswani2023attention}. As a result, the model can focus on the most relevant segments of the input. Attention has been shown to yield excellent results and facilitates parallel computation, improving processing times compared to traditional sequential data processing methods. This attention mechanism is implemented through the Multi-Head Attention layer, which identifies the most relevant relationships among the input elements \cite{khan2022transformers,vaswani2023attention}. In order to feed the Transformer architecture, the input must be in an embedding representation. There are different ways of obtaining those embeddings. In text modality, for example, sentences are encoded into numerical IDs representing words or parts of words, which are then expanded into embeddings for further processing \cite{DBLP:journals/corr/abs-1810-04805}. In the case of audio, there are two main options: using raw waveforms or spectrograms. In both cases, CNNs are employed to extract features and embeddings that feed into the Transformer encoder layers. One example of using raw waveforms is the Wav2Vec model \cite{DBLP:journals/corr/abs-1904-05862}, while an example of using spectrograms for an audio Transformer is the Audio Spectrogram Transformer \cite{gong2021ast}. This architecture has been employed in different audio-related tasks, such as music genre classification \cite{zhuang2020music}, emotion recognition \cite{andayani2022hybrid}, or speaker recognition \cite{vaessen2022fine}.

Regarding the cognitive estimation task, several works utilize the Transformer architecture to process audio signals. One of the most notable is proposed by Z. Cui et al. \cite{Cui_2023}, which not only detects Alzheimer's disease using the ADReSSo dataset but also examines the impact of depression on this disease. Some studies link emotional mood, particularly depression, with cognitive impairment disorders. A significant percentage of people suffering from Alzheimer's exhibit neuropsychiatric symptoms, and half of those with Alzheimer's also suffer from depression \cite{lyketsos2003diagnosis}. Additionally, studies show that individuals with a history of depression have a higher risk of developing Alzheimer's \cite{green2003depression,ownby2006depression}. Therefore, this study also incorporates a depression dataset, specifically the DAIC-WOZ dataset \cite{gratch2014distress}. For this study, various Transformer-based architectures were employed, including WavLM \cite{DBLP:journals/corr/abs-2110-13900}, HuBert \cite{DBLP:journals/corr/abs-2106-07447}, and Wav2Vec2 \cite{DBLP:journals/corr/abs-2006-11477}. Among these, the WavLM model performed best, achieving an F1-Score of up to 80.5, slightly surpassing the others. In addition to these proposed models, a fine-tuned version of WavLM using the depression dataset was also tested, but this approach did not improve upon the basic WavLM model. Another proposed method involved using a hidden state from the transcription model Whisper \cite{radford2022robust}, but this did not yield improvements.

In \cite{bang2024alzheimer} the authors employed the Wav2Vec2 model on the ADReSSo dataset, achieving an accuracy of up to 69.01\% and an F1-Score of 70.39\%. In this case, no other methods or Transformer models were evaluated for performance comparison. However, it included an ablation study to determine whether it was best using Whisper or Wav2Vec2 for the ASR task, with the Whisper model outperforming Wav2Vec2. This model is also employed in the study proposed by D. Escobar-Grisales et al. \cite{escobar2023deep} but uses its own recorded dataset. In this case, it is evaluated in comparison with the use of spectrograms and CNNs. The Transformer-based model slightly outperforms the use of spectrograms and 2-dimensional CNNs, obtaining up to an 88.5\% accuracy and 88.3\% F1-Score. On the other hand, raw waveforms in combination with 1-dimensional CNNs yielded poor results.

Another study that employed Wav2Vec2 is the one proposed by Y. Ying et al. \cite{ying2023multimodal}, which works with three different datasets for Alzheimer's and MCI recognition. In this study, the Transformer-based model outperforms the extraction of various audio features, including eGeMAPS and IS10\_lingusitc. This model achieves an accuracy of up to 71.2\% and an F1-Score of 73.1\% on the ADReSSo dataset for Alzheimer's detection. It also attains an accuracy and F1-Score of 85.7\% on the NCMMSC2021 long speech dataset and 81.7\% and 81\%, respectively, on the NCMMSC2021 short speech dataset, both for the detection of Alzheimer's, MCI, and healthy subjects.

Similarly, the work in \cite{POOR2024109199} utilizes the Wav2Vec2 model in combination with a dense layer to detect MCI on the I-CONECT dataset, achieving up to 50.42\% accuracy and 54.02\% F1-Score. The poor performance of this methodology is because the most relevant features in this dataset come from the video modality.

In general, we can appreciate that regarding the Transformers for audio, the most employed one and the one that achieves the best results is the Wav2Vec2 model.

\subsection{Text Modality}

The text modality represents how humans structure and express their thoughts to others using natural language in written form, as opposed to spoken language. Textual data is highly relevant for cognitive decline estimation because, by analyzing text, we can understand how people structure and express their thoughts. Additionally, it can provide valuable insights into different word choices, as cognitive function can affect expressive tasks. For instance, individuals may struggle with selecting the correct word, leading to incorrect or undesired choices.

Text is typically transcribed from audio modality. There are different approaches for this problem, for example using professional transcribers who manually transcribe audio into text. This method ensures a proper result, but its main disadvantages are its time-consuming nature and the necessity of professionals performing the task. For this purpose, there have been many recent advances in the ASR task, which aims to automatically transcribe audio into text~\cite{KHEDDAR2023110851}.

Recent works have achieved results similar to those of professional transcribers~\cite{radford2022robust}. However, when applied to the domain of health, there are additional challenges that must be addressed. For instance, in the context of cognitive decline, uncertainty or indecision may impede the accurate transcription of audio. Nevertheless, there are some works that address this issue~\cite{jamal2017automatic,torre2021improving,weiner2017manual}. In this context, the work~\cite{zhu21e_interspeech} uses the aforementioned Transformer-based Wav2Vec model to accurately transcribe the corresponding audio, incorporating pauses for further detection of dementia using textual information.

This modality can also be obtained using other approaches, such as text generative models for data augmentation, where we find many recent advantages~\cite{zhang2023survey,li2024pre}. However, as we are dealing with cognitively impaired subjects, the results may not be the most appropriate since the generation may not include the different grammatical errors or ways of expressing themselves. To address this issue, some researchers have proposed fine-tuning a GPT-2 model~\cite{radford2019language} to generate text that is more relevant, and similar to that produced by cognitively impaired individuals~\cite{li-etal-2022-gpt}.

This section presents an overview of the various methods employed for the cognitive estimation task using the text modality, all presented in Table~\ref{tab:text}. This table is organized by disorder and further categorized by dataset, presenting the year of publication, methodology used, and performance metrics such as accuracy, F1-Score, and RMSE. The best results for each dataset are highlighted in bold. The primary methods employed, including the use of RNNs and Transformer architectures, are detailed in the following subsections.

\begin{table*}[htpb]
    \centering
    \caption{Performance comparison of text-based methods over different cognitive disorders and datasets.}
    \begin{tabular}{c|c|cccccc}
        \hline
        \multirow{2}{*}{Disorder} & \multirow{2}{*}{Dataset} & \multirow{2}{*}{Work} & \multirow{2}{*}{Year} & \multirow{2}{*}{Method} & \multicolumn{3}{c}{Score}  \\
        & & & & & Acc & F1 & RMSE \\
        \Xhline{1pt}
         \multirow{29}{*}{Alzheimer} & \multirow{8}{*}{ADReSSo} & \cite{app13074244}  & 2023 & Transformer & \textbf{88.70} & - & - \\
         & & \cite{Cui_2023} & 2023 & Transformer & - & \textbf{88.00} & - \\
         & & \cite{bang2024alzheimer} & 2024 & Transformer & 83.10 & 83.10 & - \\
         & & \cite{zhu21e_interspeech} & 2021 & Transformer & 83.10 & 83.02 & 4.45 \\
         & & \cite{DBLP:journals/corr/abs-2106-15684} & 2021 & RNN & 81.00 &  - & \textbf{4.43} \\
         & & \cite{ying2023multimodal} & 2023 & Transformer & 78.90 & 79.00 & - \\
         & & \cite{10307469} & 2023 & Transformer &  77.50 & 77.40 & -  \\
         & & \cite{wang2021modular} & 2021 & Transformer & 73.50 & 73.50 & - \\

         \cline{2-8}
         & \multirow{12}{*}{ADReSS} & \cite{yuan2020disfluencies}  & 2020 & Transformer & \textbf{89.60} & - & - \\
         & & \cite{ilias2022explainable}  & 2022 & Transformer & 87.50 & \textbf{93.33} & - \\
         & & \cite{hlédiková2022data} & 2022 & Transformer & 85.00 & - & - \\
         & & \cite{Balagopalan2020ToBO}  & 2020 & Transformer & 83.30 & - & 4.56 \\
         
         & & \cite{cummins2020comparison} & 2020 & RNN & 81.30 & 81.20 & 4.66 \\
         & & \cite{interspeech2020} & 2020 & Transformer & 81.25 & - & \textbf{4.01} \\
         & & \cite{10.3389/fcomp.2021.624683} & 2021 & Transformer & 82.08 & - & - \\
         & & \cite{nambiar2022comparative} & 2022 & Transformer & 81.00 & 79.00 & - \\
         & & \cite{pompili2020inesc} & 2020 & Transformer & 72.92 & - & - \\

         \cline{2-8}
         & \multirow{8}{*}{\makecell{DementiaBank \\ Pitt Corpus}} & \cite{ORTIZPEREZ2023126413} & 2023 & Transformer & \textbf{90.36} & - & - \\
         & & \cite{liu2023approach} & 2023  & RNN & 89.50 & \textbf{89.30} & - \\
         & & \cite{roshanzamir2021transformer} & 2021  & Transformer & 88.08 & 87.23 & - \\
         & & \cite{info15010002}  & 2024 &  RNN & 83.60 & - & - \\
         & & \cite{krstev2022multimodal} & 2022 & Transformer &  82.00 & 81.00 & - \\
         & & \cite{9870792}  & 2022 &  RNN & 81.54 & 85.19 & - \\
         & & \cite{pan2019automatic} & 2019 & RNN & - & 74.37 & - \\

         \cline{2-8}
         & Private dataset & \cite{escobar2023deep}  & 2023 & Transformer & \textbf{77.90} & \textbf{76.90} & - \\

         \hline
          \multirow{2}{*}{MCI} & B-SHARP & \cite{li-etal-2020-analysis} & 2020 & Transformer & \textbf{74.10} & - & -  \\
          \cline{2-8}
          & I-CONECT & \cite{POOR2024109199} & 2024 & Transformer & \textbf{60.75} & \textbf{59.67} & - \\

         \cline{2-8}
          & Taukadial & \cite{ortizperez2024cognitive} & 2024 & Transformer & \textbf{74.82} & \textbf{79.43} & \textbf{3.14} \\

         \hline
         \multirow{2}{*}{\makecell{Alzheimer \\ \& MCI}} &  NCMMSC2021 L & \cite{ying2023multimodal} & 2023 & Transformer & \textbf{70.70} & \textbf{70.00} & - \\
         \cline{2-8}
         &  NCMMSC2021 S & \cite{ying2023multimodal} & 2023 & Transformer & \textbf{54.30} & \textbf{52.70} & - \\
         
        \hline
        
    \end{tabular}
    \label{tab:text}
\end{table*}

\subsubsection{\textbf{Recurrent Neural Networks}}

Before the arrival of the Transformer architecture, which revolutionized the entire Natural Language Processing field, the standard was to use RNNs~\cite{rumelhart1986learning}. These networks process sequential and temporal data, making them suitable for the text modality. However, RNNs have some limitations, such as struggling with long sentences and losing information from the beginning of the sequence. For this reason, different approaches have been proposed to improve these limitations. For instance, LSTM models were designed to maintain long-term dependencies better than the original RNNs \cite{hochreiter1997long,van2020review}. Similarly, GRUs were introduced to handle long-term dependencies more effectively, offering a more lightweight and faster architecture than LSTMs while achieving similar performance \cite{cho2014learning,chung2014empirical}.

The application of this methodology offers promising benefits, including the ability to capture the dependencies among different words and their relationships. Additionally, these types of networks are simple and easy to train. However, since they process data sequentially, they cannot be parallelized, which represents a significant disadvantage. This methodology has been widely employed in many tasks related to the Natural Language Processing (NLP) field, such as general text classification or sentiment analysis \cite{liu2019bidirectional,zhou2015c,basiri2021abcdm}.

In the context of cognitive estimation, in \cite{DBLP:journals/corr/abs-2106-15684} the authors propose using LSTMs along with the BERT Transformer \cite{DBLP:journals/corr/abs-1810-04805} for Alzheimer’s detection over the ADReSSo dataset. In this study, the Transformer architecture outperformed the LSTM model. However, using the LSTM in combination with text features, such as disfluencies and pauses obtained from the recordings, achieved the best results with an accuracy of up to 81\% and an RMSE score of 4.43. The work in \cite{cummins2020comparison}, also exploiting the ADReSS dataset, proposes using a bidirectional LSTM in combination with an attention module achieving up to 81.3\% and 81.2\% accuracy and F1-Score, respectively, and an RMSE score of 4.66.

In the work in \cite{9870792} over the DementiaBank Pitt Corpus, the text undergoes an enrichment preprocess, including the addition of POS-Tagging features. This, in combination with LSTMs, achieves up to 81.54\% accuracy and 85.19\% F1-Score, surpassing other proposed approaches, such as the use of Transformer encoder blocks. In the work in \cite{liu2023approach} over the same dataset, the authors combine a Text CNN \cite{DBLP:journals/corr/Kim14f} with an LSTM. The application of CNNs to sentence classification has also been demonstrated in several studies \cite{zhang2015sensitivity,liu2019bidirectional}. This combination of TextCNN and LSTM achieved an accuracy of 89.5\% and an F1-Score of 89.3\%, outperforming other proposed methods, including both approaches separately and Transformers such as DistilBERT\cite{DBLP:journals/corr/abs-1910-01108}, BERT \cite{DBLP:journals/corr/abs-1810-04805}, GPT2 \cite{radford2019language}, and RoBERTa \cite{liu2019robertarobustlyoptimizedbert}.

\begin{figure}[htpb]
    \centering
    \includegraphics[width=0.48\textwidth]{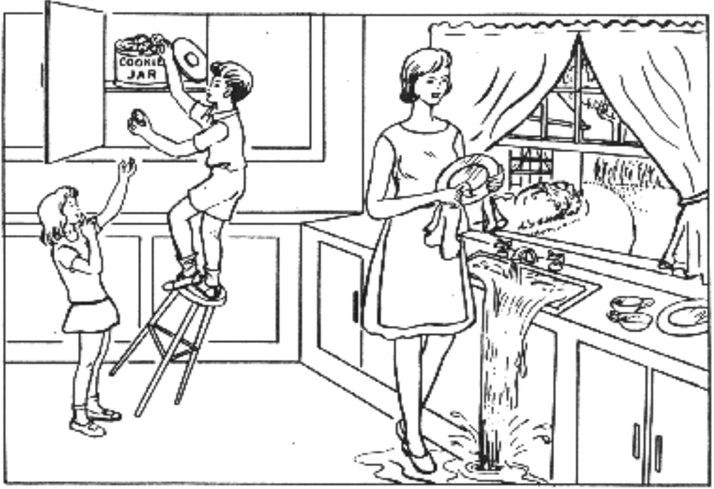}
    \caption{\textit{The Cookie Theft Picture}. The image used for the picture description task, from \cite{fritsch2019automatic}.}
    \label{fig:cookie}
\end{figure}

As previously stated, the ADDReSS dataset consists of audio recordings in which subjects describe \textit{The Cookie Theft Picture} (see Figure \ref{fig:cookie}). The work in \cite{info15010002} utilizes this image in their approach, where the picture is segmented into distinct areas. For instance, the mother and the child represent separate areas. Additionally, a cluster of synonymous words is proposed for each region (e.g., synonyms for ``mother'' include ``lady'' or ``woman''). These sets of synonyms, combined with descriptions of the image, are used to construct a matrix that counts co-occurrences in the descriptions. This matrix serves as the input embedding for a subsequent LSTM model, achieving an accuracy of up to 83.6\%. The work proposed by Y. Pan et al. \cite{pan2019automatic} combines various methodologies, including LSTMs, an attention layer, and GRU. This method, which surpasses the performance of a single LSTM used for the detection of Alzheimer's disease, achieves an F1-Score of 74.37 on the DementiaBank Pitt Corpus.

\subsubsection{\textbf{Transformers}}

As previously stated, the Transformer architecture has become state-of-the-art in many NLP tasks. A pertinent advantage of this approach is that it eliminates the need for sequential training, allowing for parallelized learning. Additionally, large pre-trained Transformer-based models, which have been trained on extensive datasets, bring significant benefits by incorporating vast textual knowledge \cite{DBLP:journals/corr/abs-2005-14165,achiam2023gpt}. This extensive preexisting knowledge has proven to attain impressive results in various related tasks.

However, these large models also have drawbacks, such as higher demands on computational power and memory, and longer training times compared to earlier methods like RNNs. This methodology is employed extensively in the field of NLP, including applications such as text classification \cite{DBLP:journals/corr/abs-1810-04805}, text summarization \cite{liu2019text}, text generation \cite{DBLP:journals/corr/abs-1910-13461,zhang2023survey}, and text translation \cite{DBLP:journals/corr/abs-1910-10683,DBLP:journals/corr/abs-2010-11934}.

Regarding the cognitive estimation tasks, a notable number of works propose the employment of this Transformer architecture. One example is the work in \cite{10307469}, in which the authors propose the use of different pre-trained Transformer models, including DistilBERT \cite{DBLP:journals/corr/abs-1910-01108}, BERT \cite{DBLP:journals/corr/abs-1810-04805}, RoBERTa \cite{liu2019robertarobustlyoptimizedbert}, DistilRoBERTa \cite{DBLP:journals/corr/abs-1910-01108}, and ALBERT \cite{DBLP:journals/corr/abs-1909-11942}. Experiments were carried out over the ADReSSo dataset, obtaining the most promising results with the use of the DistilBERT model, with 77.5\% accuracy and 77.4\% F1-Score. Similarly, in the work \cite{app13074244}, authors compare the use over the same dataset of different Transformer models, such as RoBERTa, BERT, DistilBERT, and XLNET \cite{DBLP:journals/corr/abs-1906-08237}. In this study, the most significant findings were obtained using the RoBERTa model, with an accuracy of 88.7\%. Within the same dataset, in \cite{zhu21e_interspeech} the authors experiment with the use of BERT in combination with the addition of encoded pauses during the speech to the transcriptions. The addition of such pauses results in a notable enhancement in performance, with an approximately 10\% increase in accuracy and F1-Score, and a slight improvement in the RMSE score. The final scores with these pauses have shown results of up to 83.1 and 83.02 in accuracy and F1-Score and 4.45 in RMSE. Another work using Transformers over this dataset is the one proposed in \cite{Cui_2023}, where the authors employ a BERT model obtaining up to 88\% accuracy. BERT has shown to be the most used Transformer model in this task, and in \cite{ying2023multimodal} this model is tested not only on the ADReSSo dataset but also on the NCMMSC2021 dataset, with both long and short sentences. In this study, BERT obtains 78.9\% and 74\% accuracy and F1-Score over the ADReSSo dataset, 70.7\% and 70\% in the NCMMSC2021 long sentences, and 54.3\% and 52.7\% over the short sentences of this dataset.

Regarding the works that use the textual modality on the ADReSS dataset, all but one employ Transformer architectures. The work in \cite{10.3389/fcomp.2021.624683} compares the performance of LongFormer \cite{DBLP:journals/corr/abs-2004-05150} and the base and large versions of BERT. In this case, the best performance was achieved by the LongFormer, with an accuracy of up to 82.02\% on the dataset. Moreover, the employment of the large version of BERT also resulted in an improvement over the base version. In the work proposed by I. Loukas et al. \cite{ilias2022explainable}, the authors experiment with several pre-trained Transformers, including BERT, BioBERT \cite{DBLP:journals/corr/abs-1901-08746}, BioClinicalBERT \cite{DBLP:journals/corr/abs-1904-03323}, RoBERTa, ConvBERT \cite{DBLP:journals/corr/abs-2008-02496}, ALBERT, and XLNet. The most relevant results were obtained using BERT, achieving up to 87.5\% accuracy and a 93.33\% F1-Score. This work proposed a multiclass model to detect the severity of dementia based on the MMSE Score, but it did not improve the classification approach. The work proposed in \cite{hlédiková2022data} employs a BERT model in combination with a SVM, obtaining an accuracy of 85\% on ADReSS dataset.

The work in \cite{Balagopalan2020ToBO} evaluates the BERT model against classical machine learning algorithms and textual features. The tested machine learning algorithms include Support Vector Machines (SVM), Random Forest, and Naive Bayes. The textual features included, which are very relevant to dementia detection \cite{Yancheva2015UsingLF,Zhu2018DetectingCI}, were speech-graph-based \cite{Mota2012SpeechGP}, lexical \cite{Warriner2013NormsOV}, syntactic \cite{ai2010web}, and proportions of various POS-Tagging features based on the picture-described features \cite{Croisile1996ComparativeSO}. The experimentation shows that the best results were obtained with BERT, achieving 83.3\% accuracy and a 4.56 RMSE score.

In another work \cite{interspeech2020}, the authors propose different methods, such as GLOVE embeddings \cite{pennington-etal-2014-glove} with attention modules and Transformers, such as RoBERTa, Transformer-XL \cite{DBLP:journals/corr/abs-1901-02860}, and GPT \cite{Radford2018ImprovingLU}. In addition to these models, extensive research on the addition of different textual features has also been carried out, in particular, the addition of hand-crafted features, such as psycholinguistic, repetitiveness, and lexical complexity features. These psycholinguistic features have proven to be relevant in Alzheimer’s recognition \cite{fraser2016linguistic}. Additionally, POS-Tagging features have been added and compared to the transcriptions, as they have also proven to be relevant \cite{di-palo-parde-2019-enriching}. The results showed that the best outcomes were obtained using Transformer-XL and the addition of these textual features resulted in a slight improvement over simple Transformers. The final results were 81.25\% accuracy and 4.01 RMSE.

Similarly, the work in \cite{yuan2020disfluencies} compares two Transformer models, BERT and ERNIE \cite{DBLP:journals/corr/abs-1907-12412}, in combination with the enhancement of textual features by the addition of different tokens representing pauses, depending on how long those pauses are in the conversation. This study showed better performance with the use of ERNIE and also that the inclusion of these pauses reflects an improvement, obtaining an accuracy of up to 89.6\%. The work in \cite{wang2021modular} combines Transformer architecture with various linguistic features extracted using the NLTK toolkit \cite{bird2009natural} tested on the ADReSSo dataset. Those linguistic features included POS-Tagging, as well as lexical diversity. Studies have shown that this type of feature can be relevant for Alzheimer's detection \cite{garrard2005effects,berisha2015tracking,bucks2000analysis}, and results achieved 73.5\% in both accuracy and F1-Score. 

The work \cite{roshanzamir2021transformer} proposes experimenting with different Transformers, such as BERT, XLNet, and XLM on the DementiaBank Pitt Corpus. The authors present a pipeline composed of a data augmenter, which includes synonym and contextual substitution, splitting the description into different sentences for majority voting, Transformer embeddings, and classifiers. The best results were obtained using the Large version of BERT, outperforming the base version and the other Transformers. In addition, this approach also outperformed the combination of the Transformer with an LSTM, achieving accuracy and F1-Score of 88.08 and 87.23, respectively.

In \cite{li-etal-2020-analysis}, three different Transformers, RoBERTa, BERT, and AlBERT, were tested on the B-SHARP dataset for the detection of MCI. The RoBERTa model yielded the best individual results. However, combining the embeddings from all three Transformers resulted in even better performance, achieving an accuracy of up to 74.1\%. The work proposed in \cite{ortizperez2024cognitive} employs the combination of different Transformer embeddings for the text modality, specifically for the detection of MCI in the Taukadial challenge. In this study, the embeddings obtained from a BERT model were integrated with similarity embeddings derived from a comparison between the subjects' descriptions and the actual descriptions of the images in question. This approach achieves an accuracy of 74.82, an F1-Score of 79.43, and an RMSE score of 3.14. The study by \cite{POOR2024109199} employs BERT embeddings combined with a dense layer to detect MCI using the I-CONECT dataset, achieving an accuracy of 60.75\% and an F1-Score of 59.67\%. This approach demonstrates an improvement over the results obtained using the audio modality.

In \cite{escobar2023deep}, three different approaches were proposed for the detection of Alzheimer's disease using the researchers' private dataset. These approaches, which have been combined with CNNs, included the use of Word2Vec embeddings \cite{mikolov2013efficient}, BERT, and BETO \cite{cañete2023spanishpretrainedbertmodel} Transformers. The BETO model, a Spanish version of BERT, achieved the best results with accuracy and F1-Score reaching 77.9\% and 76.9\%, respectively. The main reason for the superior performance of the BETO model is that their dataset contains recordings in Spanish.

In the work in \cite{bang2024alzheimer}, Transformers and the capabilities of large language models (LLMs) are utilized. These models have demonstrated exceptional performance in various language tasks and can be applied across many fields \cite{thirunavukarasu2023large,zhao2023survey,chang2024survey}. This study employs embeddings obtained from a BERT model alongside the ChatGPT model \cite{openai2024chatgpt}. ChatGPT has been tasked with evaluating the fluency of the speech. This evaluation, combined with the BERT embeddings yielded the best results, achieving up to 83.1\% in both accuracy and F1-Score over the ADReSSo dataset.

In addition, Transformer architectures can be integrated with more simple RNNs. An example is the work in \cite{nambiar2022comparative} on the ADReSS dataset. In this study, various embeddings, including Glove, Word2Vec, Doc2Vec \cite{DBLP:journals/corr/LeM14}, BERT, AlBERT, and RoBERTa were combined with Bi-directional LSTMs. The best results were achieved using BERT embeddings, with an accuracy of up to 81\% and an F1-Score of 79\%. It is noteworthy that most of the Transformer embeddings outperformed the other approaches. Similarly, in \cite{pompili2020inesc} BERT embeddings were employed along with LSTMs to achieve an accuracy of up to 72.92\% on the same dataset. The work proposed in \cite{ORTIZPEREZ2023126413} employs a similar approach by combining BERT embeddings with Bi-directional LSTMs, but applied to the Dementia Bank Pitt Corpus dataset, achieving an accuracy of up to 90.36\%. The work proposed in \cite{krstev2022multimodal} explores various approaches, including Sentence-BERT \cite{reimers2019sentence}, GloVe, FastText \cite{joulin2016bag}, and FLAIR \cite{akbik2018contextual}. The best performance was achieved using the Sentence-BERT model in combination with GRU, attaining an accuracy of up to 82\% and an F1-Score of 81\% on the DementiaBank Pitt Corpus dataset.

\subsection{Vision modalities}

While most research on non-intrusive modalities focuses on audio and text, other relevant modalities can also offer valuable insights for detecting these disorders. Among these are image and video modalities, which together form our proposed visual modalities. The most relevant works within these modalities are shown in Table \ref{tab:others}. This table is structured by modality and further categorized by disease and dataset. It presents the year of publication, the methodologies used, and performance metrics such as accuracy and F1-Score. The best results for each dataset are highlighted in bold. The methodologies are detailed in the following subsections according to the modality applied. These modalities are image and video. In this case, the RMSE is not reported, as none of the included works provide this information. This absence is partly due to the nature of the datasets and tasks addressed in these studies. Cognitive decline is not a binary state but rather a continuous process that varies in severity. As such, regression tasks based on cognitive assessment scores, such as those from the MMSE or MoCA, are highly relevant, as they enable the estimation of both the degree of decline and the overall cognitive functioning.

Public datasets provided by TalkBank, such as DementiaBank or ADReSS, provide these continuous metrics, making them suitable for such analyses. In contrast, most vision-based studies rely on private datasets due to privacy concerns and limited availability. Additionally, the I-Connect dataset does not include cognitive assessment scores, further limiting its applicability to regression tasks. For these reasons, the works presented in this section do not report RMSE or any regression-based performance metrics.

\begin{table*}[htpb]
    \centering
    \caption{Performance comparison of vision-based methods over different cognitive disorders and datasets.}
    \begin{tabular}{c|c|c|cccccc}
        \hline
        \multirow{2}{*}{Modality} & \multirow{2}{*}{Disorder} & \multirow{2}{*}{Dataset} & \multirow{2}{*}{Work} & \multirow{2}{*}{Year} & \multirow{2}{*}{Method} & \multicolumn{2}{c}{Score}  \\
        & & & & & & Acc & F1 \\
        \Xhline{1pt}

        Image & Alzheimer & Private dataset & \cite{9507270} &  2021  & CNN & \textbf{81.03} & - \\
        \hline

        \multirow{6}{*}{Video} & \multirow{3}{*}{MCI} & \multirow{2}{*}{I-CONECT} & \cite{SUN2024121929} & 2024 & Transformer & \textbf{90.63} & \textbf{93.03} \\
        & & & \cite{POOR2024109199} & 2024 & Transformer & 69.75 & 66.45 \\
        \cline{3-8}
        & & Private dataset & \cite{10.1007/978-3-030-66096-3_20} & 2020  & Video Features & \textbf{87.88} & - \\
        \cline{2-8}
        & Parkinson & Private dataset & \cite{hu2019graph} & 2019 & Pose & \textbf{82.50} & - \\
        \cline{2-8}
        & Apathy & Private dataset & \cite{9438658} & 2022 & Video Features & \textbf{95.34} & \textbf{94.5} \\
        \cline{2-8}
         & \makecell{Alzheimer \& MCI} & Private dataset & \cite{9747054} & 2022 & CNN & \textbf{72.21} & - \\
        
        \hline
    \end{tabular}
    \label{tab:others}
\end{table*}

\subsubsection{\textbf{Image}}

In the image modality, we observe that the proposed work by Nicole D. Cilia  et al. \cite{9507270} uses images of handwriting from healthy subjects and people with Alzheimer's recorded in their dataset. Research has shown that patients with cognitive disorders exhibit alterations in spatial organization and poor control of movements \cite{vessio2019dynamic}. Handwriting, which results from a complex network of cognitive, kinesthetic, and perceptive motor skills, can be significantly altered. Additionally, handwriting features are relevant in medicine and psychology for diagnostic purposes \cite{onofri2013dysgraphia,muller2017diagnostic}. In this work, the authors ask subjects to perform handwriting tasks on a tablet, later processing these images. In addition to RGB images, the dataset includes features such as writing speed and straightness error. This data was used to fine-tune several pre-trained CNNs, including VGG19, ResNet, and InceptionV3 \cite{szegedy2016rethinking}. The most relevant results came from using the ResNet model, yielding up to 81.03\% accuracy.

Building on that initial image-based study, several subsequent works by Cilia et al. investigate handwriting in greater depth. First, in~\cite{CILIA2022104822}, authors introduce the DARWIN dataset, the largest public online handwriting corpus for Alzheimer’s detection, and rigorously benchmark task-specific features across multiple classifiers to provide solid baseline results for future research. Second, authors introduce a machine-learning framework that extracts 26 static and dynamic features from individual handwriting strokes, including in-air movements. Using the DARWIN dataset across nine handwriting tasks, they demonstrate that stroke-based analysis performs as well as, or better than, traditional task-level features for predicting Alzheimer’s disease. This insight highlights individual strokes as an informative unit for early cognitive screening~\cite{10.1007/978-3-031-37660-3_44}. Third, authors extend stroke-level handwriting analysis to 34 tasks and show that combining discriminative feature selection with new ranking-based and stacking ensembles plus SHAP interpretability outperforms aggregated methods for Alzheimer’s detection~\cite{CILIA2024107891}. Collectively, these studies confirm that handwriting, captured as static images or high-resolution digital traces, offers a non-intrusive, information-rich modality that effectively complements audio and text-based assessments for early cognitive decline. In particular, handwriting analysis can reveal subtle motor planning and visuospatial deficits that may not surface in purely linguistic evaluations.

\subsubsection{\textbf{Video}}

Another relevant modality for these cognitive estimation tasks is video. Videos provide valuable information from subjects, such as their facial expressions and body poses during different actions, which offer insights into their movements. In contrast to image-based approaches that rely on static frames, video-based methods can leverage temporal information to provide richer insights. This temporal dimension allows models to extract information not only from individual visual representations but also from the changes and dynamics across consecutive frames.

For example, video-based analysis is particularly valuable when studying facial cues, as it captures how facial expressions evolve over time. This includes observing how expressions shift in response to spoken discourse, revealing whether certain words or emotional content elicit noticeable facial reactions. In the context of cognitive impairment, such analysis can be especially informative, for instance, by detecting changes in facial expressions as individuals struggle to find the right words. It also extends to microexpressions, subtle facial movements that often occur unconsciously during speech. Furthermore, in cognitive decline estimation tasks, many of which involve picture description, the ability to track whether a person's gaze or expressions correspond to specific elements of the image being described adds another layer of relevant information. This alignment between visual focus and verbal output can offer valuable cues about cognitive processing.

Regarding cognitive decline estimation, the study proposed by J. Sun et al. \cite{SUN2024121929} employs video modality to detect MCI, using the I-CONNECT dataset. In this work, the authors analyzed the facial expressions and movements of subjects while interacting with others on video recordings. Additionally, features such as lip and eye movements, which play a critical role in both verbal and non-verbal communication, are essential for detecting cognitive impairment \cite{nam2020analyzing,dourado2019facial,jin2020diagnosing,tanaka2019detecting}. The videos were processed using a video Transformer called MC-ViVit, based on the ViVit transformer for video \cite{DBLP:journals/corr/abs-2103-15691}. With this implementation, the authors achieved an accuracy of 90.63\% and an F1-Score of 93.03\%. In the same dataset, the approach proposed in \cite{POOR2024109199} utilizes the video modality, specifically raw frames, which are processed using a video Transformer, namely the Video Swin Transformer \cite{DBLP:journals/corr/abs-2106-13230}. This architecture extracts spatio-temporal information from the frames throughout the video. The extracted features are then combined with a dense layer to perform the final detection of the subject's cognitive state, achieving an accuracy of up to 69.75\% and an F1-Score of 66.45\%. This modality proves to be the most relevant, as it delivers the best performance when compared to other modalities.

In the work by L. Xing et al. \cite{10.1007/978-3-030-66096-3_20}, authors propose using video features with their dataset, which contains recordings of individuals using sign language, both those with MCI and healthy individuals. The videos were preprocessed to extract various video features, including facial landmarks and poses. This information was used to train different CNN models, specifically VGG16 and ResNet50. Experiments demonstrated that the VGG16 model achieved the best performance, with an accuracy of 87.88\%. 
Similarly, in the work proposed by D. Abhijit et al. \cite{9438658} various video features were obtained, including facial action units, gaze, pose, and emotional features. This study focused on their dataset to detect apathy. Although apathy is not a disease, it is common in other conditions and typically affects people's cognitive abilities. Using the extracted information and a recurrent GRU model, the authors achieved an accuracy of 95.34\% and an F1-Score of 94.5\%.

Parkinson's disease has various symptoms, including Freezing of Gait (FoG). FoG is a symptom where patients feel stuck while walking and experience a cessation of movement, even when they intend to keep walking \cite{hely2008sydney,macht2007predictors}. FoG becomes more frequent over time as the disease progresses, significantly affecting daily life and quality of life \cite{bloem2004falls,lewis2009pathophysiological}. Therefore, detecting this symptom is crucial for identifying Parkinson's disease and can be achieved through deep learning video analysis \cite{khan2013motion,nieto2016vision}. In the work in \cite{hu2019graph}, the authors suggest using pose information from videos of people walking to detect symptoms and the disease in their dataset. This pose information is used to train a recurrent GRU model, which achieves an accuracy of 82.5\%.

Eye movement tracking is a technique used to monitor where individuals focus their gaze, providing valuable insights into their visual attention. This data can be obtained through video analysis, similar to how textual data can be derived from the transcription of spoken conversations. In the study proposed in \cite{9747054}, this method has been applied to their dataset to detect Alzheimer’s and MCI. Research has demonstrated a connection between impaired eye movements and cognitive disorders, as these conditions can affect subjects' attention \cite{molitor2015eye,oyama2019novel,lagun2011detecting,mengoudi2020augmenting}. In this study, subjects were asked to describe an image that had been segmented into different regions of interest. By analyzing eye movement data, the researchers could determine which areas the subjects focused on while describing the image. This information has been used to train a CNN, specifically the VGG13 model, achieving an accuracy of 72.21\%.

\section{Multimodality}\label{sec:multimodal}

Humans engage with the environment through multiple sensory channels to gather diverse information for decision making. We use our senses to listen, observe, feel, taste, and interpret facial expressions, among other cues~\cite{lazarus1976multimodal}. The combination of different sources of information is known as a multimodal approach, and it can be very beneficial because it provides a better global understanding of our environment, enabling us to make better decisions.

As mentioned in the previous section, different sources of information offer additional insights into the problem we are addressing \cite{xu2023multimodal,guo2019deep,baltruvsaitis2018multimodal}. Combining these sources can be highly beneficial, as each modality complements the others. For instance, text alone does not convey how people speak, such as their tone of hesitation or speech jitter. However, combining text (how people structure and choose thoughts) with audio (how they utter them) provides a better understanding than using either modality alone. The works presented in Table \ref{tab:multimodal} illustrate this point. This table is organized by disorder and further categorized by dataset. It presents the year of publication, the methodology used for combining modalities, the modalities themselves, whether the multimodal approach outperforms unimodal methods, and performance metrics such as accuracy, F1-Score, and RMSE. The best results for each dataset are highlighted in bold.

\begin{table*}[htpb]
    \centering
    \caption{Performance comparison of multimodal methods over different cognitive disorders and datasets. For simplicity, ``CMA'' denotes Cross-Modal Attention, ``Imp.'' denotes improvement, ``A'' denotes Audio Modality, ``T'' denotes Text Modality, ``V'' denotes Video Modality, and ``?'' denotes whether the performance of unimodal models has not been reported, and we cannot confirm if the multimodal approach achieved better or worse results.}
    \begin{tabular}{c|c|cccccccc}
        \hline
        \multirow{2}{*}{Disorder} & \multirow{2}{*}{Dataset} & \multirow{2}{*}{Work} & \multirow{2}{*}{Year} & \multirow{2}{*}{Method} & \multirow{2}{*}{Modalities} & \multirow{2}{*}{Imp.} & \multicolumn{3}{c}{Score}  \\
        & & & & & & & Acc & F1 & RMSE \\
        \Xhline{1pt}

        \multirow{19}{*}{Alzheimer} & \multirow{7}{*}{ADReSSo} & \cite{altinok2024explainable} & 2024 & CMA & A \& T & ? & \textbf{90.00} & 90.00 & - \\ 
         & & \cite{Cui_2023} & 2023 & Joint & A \& T & \ding{51} & - & \textbf{92.80} & - \\
         & & \cite{bang2024alzheimer} & 2024 & Joint & A \& T & \ding{51} & 87.32 & 87.25 & - \\
         & & \cite{DBLP:journals/corr/abs-2106-15684} & 2021 & Joint & A \& T & \ding{51} & 84.00 &  - & \textbf{4.26} \\
         & & \cite{ying2023multimodal} & 2023 & Joint & A \& T & \ding{51} & 83.70 & 83.80 & - \\
         & & \cite{10307469} & 2023 & Transformers & A \& T & \ding{51} & 83.70 & 83.70 & -  \\
         & & \cite{wang2021modular} & 2021 & \makecell{CMA} & A \& T & \ding{51} & 77.2 & 77.6 & - \\

         \cline{2-10}
         & \multirow{9}{*}{ADReSS} & \cite{10.3389/fnagi.2022.830943} & 2022  &  \makecell{Joint} & A \& T & ? & \textbf{90.00} & \textbf{89.94} & - \\
         & & \cite{ILIAS2023101485} & 2023 & \makecell{CMA} & A \& T & \ding{51} & 88.33 & 88.69 & - \\
         & & \cite{9926818} & 2022 & \makecell{Joint} & A \& T & ? & 86.25 & 85.48 & - \\
         & & \cite{hlédiková2022data} & 2022 & \makecell{Late} & A \& T & \ding{51} & 86.00 & - & - \\         
         & & \cite{cummins2020comparison} & 2020 & \makecell{Late} & A \& T & \ding{51} & 85.20 & 85.40 & 4.65 \\
         & & \cite{10.3389/fcomp.2021.624683} & 2021 & \makecell{Joint} & A \& T & \ding{51} & 82.92 & - & - \\
         & & \cite{interspeech2020} & 2020 & Joint & A \& T & \ding{51} & 81.25 & - & \textbf{3.77} \\         
         & & \cite{pompili2020inesc} & 2020 & Early &  A \& T & \ding{51} & 81.25 & - & - \\
         & & \cite{mahajan2021acoustic} & 2021 & Joint & A \& T & \ding{51} & 72.92 & - & - \\
         
         \cline{2-10}
         & \multirow{2}{*}{\makecell{DementiaBank \\ Pitt Corpus}} & \cite{ORTIZPEREZ2023126413}  & 2023 &  Joint & A \& T & \ding{55} & \textbf{86.65} & - & - \\
         & & \cite{krstev2022multimodal} & 2022 & Joint & A \& T & \ding{55} &  81.00 & \textbf{80.00} & - \\

         \cline{2-10}
         & Own dataset & \cite{escobar2023deep}  & 2023 & Joint & A \& T & \ding{55} & \textbf{77.20} & \textbf{76.30} & - \\

         \hline
          \multirow{2}{*}{MCI} & Taukadial & \cite{ortizperez2024cognitive} & 2024 & Joint & A \& T & \ding{51} & \textbf{75.09} & \textbf{78.49}  & \textbf{2.93} \\
          \cline{2-10}
          & I-CONECT & \cite{POOR2024109199} & 2024 & CMA & T \& V & \ding{51} & \textbf{81.57} & \textbf{76.80} & - \\

         \hline
         \multirow{3}{*}{\makecell{Alzheimer \\ \& MCI}} & Own dataset & \cite{9747054} & 2022 & \makecell{CMA} & A \& V & \ding{51} & \textbf{84.26} & - & - \\
         \cline{2-10}
         & NCMMSC2021 L & \cite{ying2023multimodal} & 2023 & Joint & A \& T & \ding{51} & \textbf{89.10} & \textbf{88.60} & - \\
         \cline{2-10}
         & NCMMSC2021 S & \cite{ying2023multimodal} & 2023 & Joint & A \& T & \ding{51} & \textbf{84.00} & \textbf{83.50} & - \\

        \hline
         
    \end{tabular}
    \label{tab:multimodal}
\end{table*}

Recent advances in technology have focused on integrating multiple modalities into a single model capable of processing them simultaneously \cite{DBLP:journals/corr/abs-2103-00020,touvron2023llamaopenefficientfoundation,li2023blip2bootstrappinglanguageimagepretraining}. Various approaches exist for combining modalities, such as employing pre-existing multimodal models or using fusion strategies, which can be implemented in several ways. In this survey, the majority of methodologies for combining modalities rely on the fusion of unimodal models.

Regarding the pre-existing multimodal models, there are many examples relying on the Transformer architecture, including VideoBERT \cite{sun2019videobertjointmodelvideo} or VisualBERT \cite{li2019visualbert}, among others. The work proposed in \cite{10307469} experimented with multimodal CLIP \cite{DBLP:journals/corr/abs-2103-00020} and ViLBERT \cite{DBLP:journals/corr/abs-1908-02265}. Additionally, experiments have been carried out on the fusion of ResNet and BERT features obtained from unimodal approaches. These experiments were performed on the ADReSSo dataset to detect Alzheimer’s disease. ViLBERT achieved the best results, with an accuracy and F1-Score of up to 83.7\%, outperforming the fusion of single modality approaches.

\begin{figure*}[htbp]
    \centering
    \begin{subfigure}[b]{0.2\textwidth}
        \includegraphics[width=\linewidth]{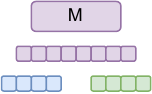}
        \caption*{(a)}
    \end{subfigure}
    \hspace{0.01\textwidth}
    \begin{subfigure}[b]{0.2\textwidth}
        \includegraphics[width=\linewidth]{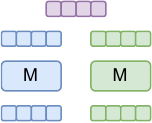}
        \caption*{(b)}
    \end{subfigure}
    \hspace{0.01\textwidth}
    \begin{subfigure}[b]{0.2\textwidth}
        \includegraphics[width=\linewidth]{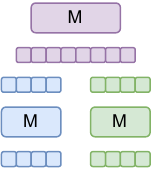}
        \caption*{(c)}
    \end{subfigure}
    \hspace{0.01\textwidth}
    \begin{subfigure}[b]{0.2\textwidth}
        \includegraphics[width=\linewidth]{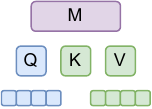}
        \caption*{(d)}
    \end{subfigure}
    \caption{
Illustration of modality fusion architectures evaluated in this study: 
(a) Early Fusion, 
(b) Late Fusion, 
(c) Joint Fusion, 
(d) Cross-Modal Attention Fusion. For simplicity, ``M'' denotes a Deep Learning model, ``Q'', ``K'', and ``V'' denote the query, key, and value, respectively, in the transformer attention mechanism. Blue blocks represent one input modality, green blocks represent the other modality, and purple blocks correspond to fused multimodal representations. Better viewed in color.
}
\label{fig:fusions}
\end{figure*}

The other methodology to combine different modalities is the use of fusion strategies. These fusions can be implemented in diverse ways and for various purposes, primarily including early, cross-modal, late, and joint fusion strategies. A visual representation of these fusions is presented in Figure~\ref{fig:fusions}. The goal of these fusions is to integrate different modalities into a single representation of the data.

The early fusion strategy involves combining different sources of information before they are processed by the model \cite{zheng2021fused,shi2022learning}. Early fusion enables the model to capture cross-modal relationships from the outset because the modalities are fused before processing. Several implementations are typical. One is concatenation, where the individual modality representations are concatenated and fed to a shared multimodal module, thereby preserving the entire information from each modality~\cite{guo2020graphcodebert}. Another option is to apply simple arithmetic operations, such as element-wise summation or averaging, to obtain a single composite vector. While computationally efficient, these operations blend the modalities and can obscure or distort modality-specific information~\cite{9156959}. This fusion strategy is also applied in pre-existing multimodal Transformers. One example is the previously mentioned VideoBERT, a Transformer model that utilizes early concatenation to encode the full multimodal context \cite{lin2021interbertvisionandlanguageinteractionmultimodal}. In \cite{pompili2020inesc}, an early fusion strategy was used to combine audio and text representations, achieving 81.25\% accuracy on the ADReSS dataset and outperforming single-modality approaches.

Cross-modal attention allows interactions across different modalities within a single model \cite{tsai2019multimodaltransformerunalignedmultimodal,xu2020cross}. Through attention mechanisms, such as those employed in transformers, the model learns to assign varying levels of importance to elements from one modality when interpreting another. In essence, it allows the representation of one modality (e.g., text) to inform and influence the interpretation of another (e.g., audio) based on their contextual relevance. This approach is particularly powerful because it supports fine-grained, temporally aligned interactions, capturing intricate dependencies across modalities that may evolve. These interactions make cross-modal attention especially well-suited for speech tasks, where audio and textual data are inherently synchronized and highly correlated. For instance, subtle prosodic cues in audio can enhance the interpretation of ambiguous textual content, while textual context can guide the model in focusing on relevant acoustic features. One typical implementation involves a cross-modal Transformer encoder that combines the key and value vectors of one modality with the query of another \cite{xu2023multimodal,9383618}. This method, known as cross-attention \cite{murahari2020largescalepretrainingvisualdialog}, was first used in the VilBERT model \cite{DBLP:journals/corr/abs-1908-02265}. In \cite{9747054}, the authors introduce a cross-modal attention Transformer that combines eye movement from video and audio features. This approach achieved an accuracy of 84.26\% on their dataset, outperforming single modalitiy approaches.

In the study \cite{altinok2024explainable}, authors propose the employment of Mel Spectrograms in combination with Visual Transformers, specifically the ViT model, for the audio modality. While this approach is integrated into their multimodal framework, no results are presented for the unimodal method. This work introduces a cross-modal attention Transformer to integrate audio and text representations. An additional fusion strategy is also considered, which involves the joint fusion of these representations through concatenation. Nevertheless, the cross-modal attention Transformer demonstrates superior performance, attaining up to 90\% accuracy and F1-Score on the ADReSSo dataset. The work proposed by W. Ning et al. \cite{wang2021modular} presents an attention layer to fuse cross-modal attention in the audio and text modalities, achieving up to 77.2\% accuracy and 77.6\% F1-Score on the same dataset. The study in \cite{ILIAS2023101485} proposes integrating BERT for the text modality as part of their multimodal framework. However, no results are reported for this unimodal approach. In this study, authors explores different fusion strategies, including cross-modal attention, joint fusion, and a Gated Multimodal Unit \cite{arevalo2020gated}. In their experiments on the ADReSS dataset, the best performance was achieved using the cross-modal attention mechanism, resulting in 88.33\% accuracy and an 88.69\% F1-Score, outperforming unimodal approaches.

The work in \cite{POOR2024109199} proposes a cross-modal attention module to fuse different modalities into bimodal approaches, comparing the outcomes obtained from each combination of modalities. Furthermore, the performance of unimodal models is also presented, offering valuable insights into the most promising information that can be extracted from this dataset. Among the unimodal models, the best performance is obtained from the video modality, followed by the text modality, and then the audio modality. This dataset consists in video recordings of elderly individuals having conversations, having the visual features relevant information. Consistent with this, the best performance is achieved by combining video and text modalities, outperforming all other bimodal and unimodal models, with up to 81.57\% accuracy and 76.80\% F1-Score. The second most effective combination comes from fusing audio and video modalities, improving the remaining unimodal models. Finally, the combination of audio and text also improves the unimodal models but ranks as the weakest bimodal combination. This study demonstrates the relevance of combining different data sources to achieve better understanding and performance. However, the authors do not present a combination of all three modalities into a trimodal model, which could potentially outperform the proposed combinations.

The late fusion strategy involves combining the results of each unimodal model after processing them separately \cite{ZHANG2021104042}. This approach requires the development of separate models for each modality, allowing each to extract the most relevant information from its respective input. As a result, it provides a clearer understanding of each modality before their outputs are fused. The final fusion can be performed in various ways, for instance, by applying a weighted combination of modality-specific predictions. This combination is particularly beneficial in scenarios where one modality consistently outperforms the others. Alternatively, the fusion can be guided by the confidence scores of the models, assigning greater weight to predictions with higher confidence.

In \cite{cummins2020comparison}, the authors propose using majority voting for classification tasks and a weighted average of the results for regression tasks on the ADReSS dataset. This methodology achieved an accuracy of 85.2\%, 85.4\% F1-Score, and a RMSE score of 4.65. Similarly, the work proposed in \cite{hlédiková2022data} also employs majority voting for their unimodal approaches. In this study, this fusion strategy yielded 86\% accuracy on the same dataset, outperforming the unimodal approaches.

A joint or hybrid fusion strategy involves combining features obtained from each modality for later use in a final model, but only after being processed individually by each modality model \cite{10.1093/bib/bbab569}. These intermediate representations are then combined and passed to a final multimodal model for joint learning. This approach leverages the strengths of unimodal processing, allowing each modality-specific model to extract the most salient and task-relevant features from its input before any fusion occurs. 

By combining these rich, modality-specific features at an intermediate level, the joint fusion strategy enables the final multimodal model to learn complex cross-modal relationships and interactions more effectively than early or late fusion alone. However, this strategy comes with increased computational cost and complexity. Since it relies on multiple components, two or more unimodal encoders followed by a multimodal model, it demands more memory and training resources than other fusion strategies. Despite this overhead, this combination is the most common approach in the selected works for this study.

The work proposed in \cite{Cui_2023} employs a concatenation of the features obtained from each modality model, followed by training a Transformer encoder. This approach achieved a maximum F1-Score of 92.8\% on the ADReSSo dataset. Similarly, the work proposed by B. Jeong-Uk et al. \cite{bang2024alzheimer} and M. Rohania\cite{DBLP:journals/corr/abs-2106-15684} uses a concatenation of features for subsequent training of a dense layer. The former yielded an accuracy of 87.32\% and F1-Score of 87.25\% on the ADReSSo dataset, while the latter resulted in an accuracy of 84\% and RMSE score of 4.26 on the same dataset.
In \cite{ying2023multimodal}, after concatenation, the trained model was a SVM, which achieved an accuracy of 83.7\% and an F1-Score of 83.8\% on the ADReSSo dataset. This methodology was also tested on the NCMMSC2021 dataset, achieving up to 89.1\% accuracy and F1-Score of 88.6\% for large sentences. In the case of short sentences, the results yielded 84\% accuracy and F1-Score of 83.5\%.

In \cite{mahajan2021acoustic}, the authors introduce a LSTM model for the text modality, which is implemented in their multimodal approach. However, no results are provided using only this modality. For the combination of features, authors propose concatenating the features followed by a dense layer for final multimodal classification. This approach, applied to the ADReSS dataset, yielded 72.92\% accuracy. Similarly, the work in \cite{10.3389/fcomp.2021.624683} employs this methodology and compares it with a late fusion strategy. In this study, the joint approach outperforms the late one, achieving 82.92\% accuracy on the same dataset.
In the case of \cite{9926818}, instead of concatenation, joint fusion is performed by taking the product of the features, followed by a dense layer for final multimodal classification. This method yielded accuracy of 86.25\% and 85.48\% F1-Score on the ADReSS dataset. However, it is uncertain whether this approach could improve on unimodal approaches, as unimodal results are not reported.

In the work in \cite{escobar2023deep}, the authors propose implementing a joint concatenation followed by a dense layer, comparing the results with early and late fusion strategies. The best performance was achieved by the joint function, with an accuracy of 77.2\% and F1-Score of 77.6\% on their dataset. However, this multimodal approach did not outperform the audio-proposed modality.

In \cite{interspeech2020}, the authors experiment with joint concatenation followed by a MLP on the ADReSS dataset. The results achieved were 81.25\% accuracy and an RMSE score of 3.77. Similarly, the work in \cite{ORTIZPEREZ2023126413} employs concatenation followed by an MLP, yielding 86.65\% accuracy on the DementiaBank Pitt Corpus, though it does not improve on the text modality results.
In the work in \cite{krstev2022multimodal}, the authors propose a joint concatenation followed by an MLP, contrasting the results with early and late fusion strategies. The best results were achieved by the joint fusion, with 81\% accuracy and an F1-Score of 80\%. The authors in \cite{ortizperez2024cognitive} also propose a joint concatenation followed by an MLP, achieving up to 75.09\% accuracy, an F1-Score of 78.49\%, and an RMSE score of 2.93 on the Taukadial dataset. Finally, the work in \cite{10.3389/fnagi.2022.830943} proposes a joint concatenation followed by an attention module and a dense layer, outperforming other proposed fusion strategies, such as cross-attention. This methodology achieved 90\% accuracy and an F1-Score of 89.94\% on the ADReSS dataset.

The application of multimodal approaches, which use all available information for a task, has been shown to enhance the outcomes of unimodal models in most situations, with few exceptions (see \cite{ORTIZPEREZ2023126413,krstev2022multimodal,escobar2023deep}). Therefore, employing these methodologies is crucial for improving not only the performance of the models but also their robustness. Moreover, the joint fusion approach is the most prominent and usually achieves the best results.

\section{Discussion}\label{sec:discussion}

In previous sections, the most promising publicly available datasets and methodologies for cognitive state estimation using non-invasive modalities have been reviewed. This section presents the main findings of our analysis, together with suggestions for future research directions. These findings are related to cognitive disorders, datasets, and both unimodal and multimodal approaches.

\subsection{Cognitive disorders}

In addition to the natural decline associated with aging, several neurological and neurodegenerative conditions can significantly impair cognitive abilities. The severity and nature of these impairments vary depending on the disorder, and in many cases, they can lead to progressive functional decline and increased dependency. Early detection is crucial for timely intervention that can significantly improve a patient's quality of life.

Cognitive disorders include a range of conditions such as dementia, MCI, aphasia, Parkinson’s disease, and apathy. As mentioned earlier, the global population is aging, which has led to a growing prevalence of age-related conditions, most notably dementia. As a result, dementia remains the primary focus of research in this area, in part due to its public health impact and the relatively higher availability of relevant datasets. However, there is a growing body of work exploring other disorders, such as MCI, and, in some cases, examining their combined or differential diagnosis.

Notably, the selection of appropriate data modalities for cognitive assessment tasks is inherently task and disorder-specific. This selection should be determined by the nature of the impairments associated with each condition. For instance, Alzheimer’s disease and MCI are primarily characterized by cognitive and linguistic deficits. Patients frequently experience challenges with word retrieval, semantic coherence, and syntactic complexity. These impairments are often detectable in the structure and content of their spoken or written language, making text-based analysis a particularly effective modality for early detection~\cite{luz2020alzheimers,app13074244}.

In contrast, conditions such as Parkinson’s disease and aphasia more directly affect motor control and speech production. These disorders often lead to disruptions in articulation, prosody, phonation, and overall speech fluency. Consequently, audio-based approaches, which can capture features such as jitter, shimmer, intonation, and speech rate, are more suitable for identifying relevant patterns in these populations~\cite{tsanas2012novel,ALMEIDA201955}.

This alignment between disorder characteristics and modality relevance is reflected in the literature. Alzheimer’s and MCI are frequently studied using language-based models, whereas Parkinson’s disease and aphasia are more commonly addressed using speech-based models that emphasize vocal features. Understanding these distinctions is essential not only for designing effective predictive models but also for ensuring that research is clinically grounded and tailored to the specific diagnostic needs of each condition.

\subsection{Datasets}

Due to the sensitivity of this domain, large-scale datasets for training deep learning models remain scarce, posing a significant barrier to advancement. Deep learning techniques typically require extensive and diverse datasets to achieve robust and generalizable performance. This scarcity not only affects model accuracy on benchmark test sets but, more critically, limits the ability of models to generalize beyond their training environments.

Among the limited datasets that are available, there is considerable variation in both structure and content. Some assess performance across multiple cognitive tasks, such as image description or verbal fluency exercises, providing a multifaceted perspective on cognitive function. While most datasets focus on Alzheimer’s disease, some also include data related to conditions such as MCI, thereby expanding their clinical relevance. In terms of data modalities, the majority of resources rely on audio and text. This preference is primarily driven by privacy concerns surrounding video data, mainly when it involves identifiable facial recordings.

From a demographic standpoint, these datasets primarily feature older adults, reflecting the age-related nature of dementia. While this focus aligns with the target population, it reduces the generalizability of models to younger individuals or atypical cases. Gender distribution tends to be relatively balanced.
Many existing approaches have been trained on relatively small and homogeneous datasets collected under tightly controlled conditions. While such standardization can enhance internal validity, it often compromises external validity. As a result, models trained under narrow parameters may underperform in real-world environments that differ from the training setup. These models are particularly vulnerable to overfitting and may struggle with variability in speaker demographics, contextual differences, or environmental noise. These imbalances can be addressed through more inclusive data collection practices, as well as through domain adaptation techniques. These methods enable models trained on specific populations to adapt to new target groups, such as individuals from different age brackets, genders, or cultural backgrounds. In addition, the inclusion of relevant metadata, when ethically and legally permissible, allows for more granular performance evaluation and targeted optimization.

Furthermore, the dominance of English in publicly available datasets aggravates this issue. Resources in other languages, such as Spanish or Mandarin, are limited and often lack sufficient diversity and size for reasonable training. This linguistic imbalance can lead to models that underperform on underrepresented populations, reinforcing existing disparities. In clinical domains where fairness, accuracy, and equity are paramount, such biases raise significant ethical concerns. If left unaddressed, these issues could result in misdiagnoses or unequal access to care. This dominance of English in existing datasets can be partially addressed through the use of multilingual and cross-lingual learning strategies. Models pretrained on large-scale multilingual corpora can be fine-tuned using limited data from underrepresented languages, enabling effective knowledge transfer and reducing performance disparities.

Some initiatives have sought to mitigate these biases. For instance, the Taukadial dataset includes both English and Chinese speakers. However, inconsistencies in task design, such as using different pictures for each language group, introduce confounding factors that complicate cross-linguistic comparisons. These protocol variations make it difficult to determine whether differences in model behavior are due to linguistic characteristics or task-related discrepancies.

Fairness-aware machine learning techniques can also be employed during training or post-processing to reduce disparities in model performance across demographic or linguistic subgroups. Approaches such as loss function reweighting, adversarial debiasing, and subgroup-specific threshold calibration can promote more equitable outcomes without significantly compromising overall accuracy. These considerations are particularly critical in healthcare settings, where fairness and transparency are essential to avoid potential harm, misdiagnosis, or unequal access to care.

Despite meaningful technical progress, the clinical adoption of these models remains limited. Key obstacles include the need for large-scale validation across diverse populations, integration with existing clinical workflows, and compliance with regulatory and ethical standards. Moreover, computational limitations are common in clinical settings, where available infrastructure is often optimized for administrative tasks rather than resource-intensive machine learning operations. Effective deployment will require not only more powerful computing systems but also standardized data collection protocols and recording environments aligned with those used during model development.

These limitations underscore the importance of model interpretability. Explainable models can help identify when decisions are due to speech patterns or biased linguistic features. Transparent and accountable decision-making is crucial to ensure fairness, trust, and efficacy, particularly when working with underrepresented or demographically uneven populations.

\subsection{Explainability}

An essential yet often underemphasized aspect in the development of machine learning models for cognitive assessment is explainability. In the context of sensitive medical data, it is essential to understand why and how predictions are obtained. Explainability not only enhances trust and transparency but also facilitates the extraction of clinically meaningful insights from model behavior. Additionally, this information can bring valuable insights about the processed data. Although recent studies have increasingly acknowledged the importance of explainability, this area remains underdeveloped, with many models still operating as opaque ``black boxes''~\cite{GORRIZ2023101945,holzinger2019causability,vilone2021notions}.

Among the studies included in this review, explainability methods were notably limited. Nevertheless, some efforts have attempted to address this gap. For instance, authors in~\cite{ORTIZPEREZ2023126413} examined attention mechanisms within their models to assess the contribution of individual words to final classification outcomes. Their analysis revealed that certain discourse markers, such as ``Okay'' and ``Well'' were consistently emphasized by the model while predicting dementia. These markers are normally employed by individuals experiencing word-finding difficulties or requiring additional time to organize their thoughts, providing both interpretability and clinical relevance.

Similarly, Ilias et al.~\cite{ilias2022explainable} analyzed linguistic patterns using the LIME framework~\cite{ribeiro2016should} to identify influential tokens, including interjections, personal pronouns, and past-tense verbs. Their findings indicate that individuals with dementia often rely on vague or generalized language, consistent with cognitive impairments. Frequent use of fillers such as “yeah” or “oh” emerged as strong indicators of dementia, reflecting hesitation or lexical retrieval difficulties. These insights not only support the model's predictions but also align with established clinical observations~\cite{almor1999alzheimer}.

Beyond performance metrics, these examples underscore the growing recognition of explainability as a foundational component in the development of trustworthy AI systems for cognitive assessment. While tools such as attention analysis and LIME offer promising avenues for interpretability, their adoption remains limited across the field. Future research should prioritize transparency as a core design principle, embedding explainability throughout the model development, evaluation, and validation pipeline. Establishing standardized protocols to assess and benchmark model explanations will be critical to ensuring clinical reliability, ethical alignment, and equitable deployment across diverse populations.

\subsection{Unimodal Approaches}
As previously discussed, the clinical profile of each cognitive disorder plays a key role in determining the most informative data modality for assessment. This section focuses on how individual modalities, specifically audio, text, and, to a lesser extent, visual data, have been employed in unimodal machine learning models to estimate cognitive states.

Audio-based approaches rely on speech recordings, which can reveal subtle markers of cognitive decline through articulation, prosody, fluency, and hesitation patterns. Deep learning techniques, particularly Transformer-based architectures, have enabled automatic feature extraction and improved predictive performance in this domain. One of the most widely used models is Wav2Vec2, which learns meaningful acoustic representations from raw audio. However, working with audio data often requires extensive preprocessing, including segmentation, normalization, and noise reduction, all of which add complexity to the modeling pipeline. Moreover, variability in recording conditions and speaker characteristics can impact model generalizability. While Wav2Vec2 remains a strong baseline, emerging models such as HuBERT and Whisper offer promising alternatives that warrant further exploration in cognitive assessment contexts.

Text-based approaches typically analyze transcriptions derived from spoken language. These offer access to syntactic structure, semantic content, and discourse coherence, all of which are often disrupted in individuals with cognitive impairments. Transformer models such as BERT and its derivatives (e.g., DistilBERT, RoBERTa) are widely used in this domain. When combined with handcrafted linguistic features, these models have often outperformed audio-based approaches, highlighting the strength of language structure as a signal of cognitive decline. Notably, most current work employs relatively compact pretrained models (e.g., BERT-base) rather than larger foundation models like LLaMA or Mistral. This trend likely reflects the small size of available datasets and concerns about overfitting with high-capacity models. In some cases, large language models are used not as primary encoders but as feature extractors to enrich downstream representations. Exploring scalable and efficient integration of these larger models remains an open and promising research direction.

Visual modalities, such as image and video data, are underrepresented in the reviewed studies, primarily due to the limited availability of suitable datasets. Nonetheless, these modalities hold substantial potential for cognitive assessment, particularly through the analysis of facial expressions, gaze behavior, and motor features during speech. Prior work in related areas, such as emotion recognition or gait analysis for Parkinson’s disease, demonstrates that visual cues can offer valuable insight into neurological function. However, the limited availability of visual datasets is largely due to the privacy risks associated with video recordings, which frequently contain identifiable personal information. These privacy concerns introduce ethical and legal barriers to data collection and sharing. Potential solutions include data anonymization techniques, such as facial blurring or skeletal pose extraction, which can preserve behavioral features while protecting individual identities. As privacy-preserving technologies continue to evolve, the integration of visual data into cognitive assessment research may become more viable and impactful.

While unimodal approaches have demonstrated promising results, especially when aligned with disorder-specific impairments, they also face limitations in terms of robustness and generalizability. Future work may benefit from combining complementary modalities to capture a richer and more holistic picture of cognitive function.

\subsection{Multimodal Approaches}

The integration of multiple modalities in cognitive assessment tasks has shown considerable promise, often leading to improved performance and enhanced model robustness. By combining complementary sources of information, such as speech, language, and visual cues, multimodal approaches enable a more comprehensive representation of an individual’s cognitive state. This strategy is particularly beneficial in real-world scenarios where specific modalities may be missing, noisy, or less informative, allowing the model to rely on alternative sources.

Despite these advantages, several challenges persist. Multimodal processing requires more complex model architectures capable of aligning and integrating heterogeneous data types. The training and inference stages are often more computationally intensive, and ensuring meaningful interactions between modalities requires careful design choices. Empirical findings from the reviewed literature reflect both the potential and the limitations of this approach. Specifically, multimodal integration resulted in improved outcomes in 75\% of the analyzed studies. However, in 12.5\% of cases, performance decreased when combining modalities. While the other 12.5\% of studies did not report unimodal baselines, making it difficult to draw definitive conclusions about the universal benefit of multimodality.

Multimodal integration is typically achieved through two main strategies: the use of pretrained multimodal models or the application of fusion techniques. While fusion-based approaches dominate current research, the use of pretrained multimodal models remains limited. Only one notable study was identified in this survey. This gap highlights a promising direction for future work, particularly in exploring models trained on large-scale datasets that encode structured knowledge across modalities. These models have the potential to enhance cognitive assessment by capturing subtle cross-modal interactions and context-dependent behavioral patterns.

In parallel, LLMs, pretrained on vast textual corpora, have demonstrated exceptional performance across a variety of tasks. Many state-of-the-art text-based approaches in this field rely on features derived from these models, underscoring the effectiveness of transfer learning. Extending this paradigm to multimodal settings, pretrained models capable of jointly encoding text, audio, and video inputs represent a compelling opportunity. These architectures can learn integrated multimodal representations that capture complex relationships between language, speech, and behavior, potentially improving accuracy in detecting cognitive decline and related conditions.

Among the fusion strategies examined, joint fusion consistently emerged as the most effective. Early fusion, which combines modalities at the input level, often fails to preserve the distinct characteristics of each data type. Late fusion, which merges predictions at the decision level, may overlook meaningful intermodal relationships. In contrast, joint fusion integrates intermediate feature representations, enabling both modality-specific learning and intermodal interaction modeling. This balanced integration typically leads to superior performance.

Attention-based mechanisms, particularly those implemented within Transformer architectures, have also been explored for cross-modal integration. However, they have not consistently outperformed joint fusion strategies. One possible explanation lies in the preprocessing differences between modalities. Audio data is often segmented into temporal windows, while textual data is tokenized into subword units. These divergent representations may hinder effective alignment in attention layers, limiting the model’s ability to capture fine-grained cross-modal dependencies~\cite{ortiz2025cognialign}.

Emerging techniques, such as dynamic fusion and hierarchical fusion, offer promising alternatives. These methods enable models to adaptively weigh the contribution of each modality based on task-specific or context-dependent relevance. Such flexibility may further enhance performance and applicability, particularly in clinical settings where input quality and availability can vary significantly.

\subsection{Current Limitations}

Despite the promising advancements outlined in previous sections, particularly in the use of multimodal architectures and pretrained models, several critical limitations continue to constrain progress in real-world applications of the cognitive state assessment.

One of the most persistent challenges is the scarcity of labeled data, which limits the model's robustness and generalizability. Multiple reviewed studies have conclusively demonstrated the effectiveness of data augmentation techniques tailored to specific modalities, reporting improved performance compared to models trained without them. These strategies offer a promising avenue for mitigating dataset limitations. Alternatively, the use of large pretrained models, particularly in few-shot and zero-shot learning contexts, can reduce the dependency on densely annotated data. Notably, Vision-Language Models (VLMs) have shown potential for video understanding tasks~\cite{wang2021actionclipnewparadigmvideo,wu2023bidirectionalcrossmodalknowledgeexploration,DBLP:journals/corr/abs-2103-00020}.

Another key limitation lies in the gap between academic benchmarks and real-world clinical implementation. Many studies focus primarily on maximizing performance on curated datasets while overlooking generalizability, usability, and integration into healthcare workflows. Moreover, the computational demands of training and deploying state-of-the-art models pose practical constraints. Video and speech modalities, in particular, require substantial preprocessing and processing power. This is especially problematic in clinical settings where hardware is often optimized for administrative use rather than high-throughput machine learning.

Finally, several overarching issues continue to limit the adoption and trustworthiness of these models. These include overfitting due to small sample sizes, poor generalization across diverse populations and languages, and the lack of interpretability in many high-performing models. Additionally, ethical and regulatory concerns, including transparency, data privacy, and algorithmic fairness, present non-trivial obstacles to clinical deployment. Collectively, these limitations underscore the need for more robust, interpretable, and resource-efficient models, as well as rigorous validation in real-world, heterogeneous environments.

\subsection{Future directions}

One of the most important directions for future research is the development of large-scale, diverse, and multimodal datasets for training deep learning models in cognitive assessment. These datasets should incorporate detailed demographic information and richer data sources, such as longitudinal medical records and symptom progression data, to enable more personalized and temporally-aware modeling. Including nonverbal cues, such as facial expressions and gestures, may further enhance multimodal approaches. To improve generalizability and fairness, future datasets should also prioritize multilingual and multicultural representation through standardized data collection protocols.

An additional promising source of data comes from the use of Human Digital Twins (HDTs). These digital representations of individuals can include comprehensive medical records, offering crucial insights into a patient's health status. Furthermore, HDTs can simulate and analyze daily human activities, generating synthetic data that could be beneficial for training models aimed at cognitive decline estimation \cite{tao2018digital,bruynseels2018digital}.

\section{Conclusions}\label{sec:conclusions}

This survey reviewed key non-intrusive methodologies for analyzing and detecting this decline, focusing on non-intrusive methods. These approaches do not disrupt daily activities, preserving quality of life without relying on invasive techniques like medical imaging. By bridging technical methods with real-world medical needs, we aim to foster interdisciplinary collaboration on the early detection of cognitive decline.

To the best of our knowledge, this study represents the first survey of estimation of general cognitive decline based on non-invasive data. The methodologies covered include audio, textual, and visual analyses, leveraging deep learning techniques. We also assess the key features and benefits of each approach, their integration into multimodal frameworks, and their performance on standard datasets. We present the most significant datasets relevant to this task, providing detailed information about their content, including the activities performed by subjects, the disorders studied, and the modalities provided.

We conclude that there is a significant lack of datasets in this area, with limited options that are not optimal in terms of sample size. Existing datasets typically focus on audio and text modalities, primarily due to the sensitivity of the data involved. For example, video data can raise additional concerns regarding privacy and ethical considerations, which further limits its availability and use. Consequently, most proposed studies rely on audio and text modalities, while other non-intrusive modalities, including video, have very few representations and often use their own recorded datasets.

From our discussion comparing these studies, we also determine that the text modality is the most relevant feature, yielding the most significant results across the datasets. Furthermore, the combination of these modalities generally leads to improved performance, surpassing any unimodal approach. This is because different data types complement one another, providing a more comprehensive understanding of the problem.

\section*{Acknowledgments}
\sloppy
This work was supported by the Spanish State Research Agency (AEI) under grant: GEMELIA PID2024-161711OB-I00; and ERDF/EU. Additionally, it received support from the CIAICO/2022/132 Consolidated group project ``AI4Health'',funded by the Valencian government and International Center for Aging Research ICAR funded project ``IASISTEM''. This work is also a part of the ENIA Chair of Artificial Intelligence from the University of Alicante (TSI-100927-2023-6) funded by the Recovery, Transformation and Resilience Plan from the European Union Next Generation through the Ministry for Digital Transformation and the Civil Service. This work has also been supported by a Spanish national and a regional grant for PhD studies, FPU21/00414, and CIACIF/2022/175.

\bibliographystyle{IEEEtran}
\bibliography{refs}

\begin{thebibliography}{100}
\providecommand{\url}[1]{#1}
\csname url@samestyle\endcsname
\providecommand{\newblock}{\relax}
\providecommand{\bibinfo}[2]{#2}
\providecommand{\BIBentrySTDinterwordspacing}{\spaceskip=0pt\relax}
\providecommand{\BIBentryALTinterwordstretchfactor}{4}
\providecommand{\BIBentryALTinterwordspacing}{\spaceskip=\fontdimen2\font plus
\BIBentryALTinterwordstretchfactor\fontdimen3\font minus \fontdimen4\font\relax}
\providecommand{\BIBforeignlanguage}[2]{{%
\expandafter\ifx\csname l@#1\endcsname\relax
\typeout{** WARNING: IEEEtran.bst: No hyphenation pattern has been}%
\typeout{** loaded for the language `#1'. Using the pattern for}%
\typeout{** the default language instead.}%
\else
\language=\csname l@#1\endcsname
\fi
#2}}
\providecommand{\BIBdecl}{\relax}
\BIBdecl

\bibitem{cognitive}
D.~Kiely, ``Cognitive function in encyclopaedia of quality of life and well-being research (ed. michalos, ac) 974--978,'' 2014.

\bibitem{age}
\BIBentryALTinterwordspacing
I.~J. Deary, J.~Corley, A.~J. Gow, S.~E. Harris, L.~M. Houlihan, R.~E. Marioni, L.~Penke, S.~B. Rafnsson, and J.~M. Starr, ``{Age-associated cognitive decline},'' \emph{British Medical Bulletin}, vol.~92, no.~1, pp. 135--152, 09 2009. [Online]. Available: \url{https://doi.org/10.1093/bmb/ldp033}
\BIBentrySTDinterwordspacing

\bibitem{dementia}
\BIBentryALTinterwordspacing
S.~Duong, T.~Patel, and F.~Chang, ``Dementia: What pharmacists need to know,'' \emph{Canadian Pharmacists Journal / Revue des Pharmaciens du Canada}, vol. 150, no.~2, pp. 118--129, 2017. [Online]. Available: \url{https://doi.org/10.1177/1715163517690745}
\BIBentrySTDinterwordspacing

\bibitem{mci}
\BIBentryALTinterwordspacing
F.~Portet, P.~J. Ousset, P.~J. Visser, G.~B. Frisoni, F.~Nobili, P.~Scheltens, B.~Vellas, J.~Touchon, and the MCI Working Group of the European Consortium~on Alzheimer{\textquoteright}s Disease~(EADC), ``Mild cognitive impairment (mci) in medical practice: a critical review of the concept and new diagnostic procedure. report of the mci working group of the european consortium on alzheimer{\textquoteright}s disease,'' \emph{Journal of Neurology, Neurosurgery \& Psychiatry}, vol.~77, no.~6, pp. 714--718, 2006. [Online]. Available: \url{https://jnnp.bmj.com/content/77/6/714}
\BIBentrySTDinterwordspacing

\bibitem{mci2}
\BIBentryALTinterwordspacing
P.~Celsis, ``Age-related cognitive decline, mild cognitive impairment or preclinical alzheimer's disease?'' \emph{Annals of Medicine}, vol.~32, no.~1, pp. 6--14, 2000, pMID: 10711572. [Online]. Available: \url{https://doi.org/10.3109/07853890008995904}
\BIBentrySTDinterwordspacing

\bibitem{ageing}
{World Health Organization}, ``Mental health action plan 2013-2020,'' \emph{WHO Library Cataloguing-in-Publication DataLibrary Cataloguing-in-Publication Data}, pp. 1--44, 2023.

\bibitem{damasio1992aphasia}
A.~R. Damasio, ``Aphasia,'' \emph{New England Journal of Medicine}, vol. 326, no.~8, pp. 531--539, 1992.

\bibitem{s24092751}
\BIBentryALTinterwordspacing
C.~Cabello-Collado, J.~Rodriguez-Juan, D.~Ortiz-Perez, J.~Garcia-Rodriguez, D.~Tomás, and M.~F. Vizcaya-Moreno, ``Automated generation of clinical reports using sensing technologies with deep learning techniques,'' \emph{Sensors}, vol.~24, no.~9, 2024. [Online]. Available: \url{https://www.mdpi.com/1424-8220/24/9/2751}
\BIBentrySTDinterwordspacing

\bibitem{zhao2019deep}
R.~Zhao, R.~Yan, Z.~Chen, K.~Mao, P.~Wang, and R.~X. Gao, ``Deep learning and its applications to machine health monitoring,'' \emph{Mechanical Systems and Signal Processing}, vol. 115, pp. 213--237, 2019.

\bibitem{venugopalan2021multimodal}
J.~Venugopalan, L.~Tong, H.~R. Hassanzadeh, and M.~D. Wang, ``Multimodal deep learning models for early detection of alzheimer’s disease stage,'' \emph{Scientific reports}, vol.~11, no.~1, p. 3254, 2021.

\bibitem{liu2014early}
S.~Liu, S.~Liu, W.~Cai, S.~Pujol, R.~Kikinis, and D.~Feng, ``Early diagnosis of alzheimer's disease with deep learning,'' in \emph{2014 IEEE 11th international symposium on biomedical imaging (ISBI)}.\hskip 1em plus 0.5em minus 0.4em\relax IEEE, 2014, pp. 1015--1018.

\bibitem{early}
M.~Di~Luca, D.~Nutt, W.~Oertel, P.~Boyer, J.~Jaarsma, F.~Destrebecq, G.~Esposito, and V.~Quoidbach, ``Towards earlier diagnosis and treatment of disorders of the brain,'' \emph{Bulletin of the World Health Organization}, vol.~96, no.~5, p. 298, 2018.

\bibitem{mri1}
D.~Wen, Z.~Wei, Y.~Zhou, G.~Li, X.~Zhang, and W.~Han, ``Deep learning methods to process fmri data and their application in the diagnosis of cognitive impairment: a brief overview and our opinion,'' \emph{Frontiers in neuroinformatics}, vol.~12, p.~23, 2018.

\bibitem{mri2}
H.~Taheri~Gorji and N.~Kaabouch, ``A deep learning approach for diagnosis of mild cognitive impairment based on mri images,'' \emph{Brain sciences}, vol.~9, no.~9, p. 217, 2019.

\bibitem{mri3}
L.~Kang, J.~Jiang, J.~Huang, and T.~Zhang, ``Identifying early mild cognitive impairment by multi-modality mri-based deep learning,'' \emph{Frontiers in aging neuroscience}, vol.~12, p. 206, 2020.

\bibitem{mri4}
X.~Feng, Z.~C. Lipton, J.~Yang, S.~A. Small, F.~A. Provenzano, A.~D.~N. Initiative, F.~L. D.~N. Initiative \emph{et~al.}, ``Estimating brain age based on a uniform healthy population with deep learning and structural magnetic resonance imaging,'' \emph{Neurobiology of aging}, vol.~91, pp. 15--25, 2020.

\bibitem{mri5}
S.~Chauhan, L.~Vig, M.~De~Filippo De~Grazia, M.~Corbetta, S.~Ahmad, and M.~Zorzi, ``A comparison of shallow and deep learning methods for predicting cognitive performance of stroke patients from mri lesion images,'' \emph{Frontiers in neuroinformatics}, vol.~13, p.~53, 2019.

\bibitem{mri6}
G.~Zhu, B.~Jiang, L.~Tong, Y.~Xie, G.~Zaharchuk, and M.~Wintermark, ``Applications of deep learning to neuro-imaging techniques,'' \emph{Frontiers in neurology}, vol.~10, p. 869, 2019.

\bibitem{mri7}
T.~Jo, K.~Nho, and A.~J. Saykin, ``Deep learning in alzheimer's disease: diagnostic classification and prognostic prediction using neuroimaging data,'' \emph{Frontiers in aging neuroscience}, vol.~11, p. 220, 2019.

\bibitem{pellegrini2018machine}
E.~Pellegrini, L.~Ballerini, M.~d. C.~V. Hernandez, F.~M. Chappell, V.~Gonz{\'a}lez-Castro, D.~Anblagan, S.~Danso, S.~Mu{\~n}oz-Maniega, D.~Job, C.~Pernet \emph{et~al.}, ``Machine learning of neuroimaging for assisted diagnosis of cognitive impairment and dementia: a systematic review,'' \emph{Alzheimer's \& Dementia: Diagnosis, Assessment \& Disease Monitoring}, vol.~10, pp. 519--535, 2018.

\bibitem{graham2020artificial}
S.~A. Graham, E.~E. Lee, D.~V. Jeste, R.~Van~Patten, E.~W. Twamley, C.~Nebeker, Y.~Yamada, H.-C. Kim, and C.~A. Depp, ``Artificial intelligence approaches to predicting and detecting cognitive decline in older adults: A conceptual review,'' \emph{Psychiatry research}, vol. 284, p. 112732, 2020.

\bibitem{choi2018predicting}
H.~Choi, K.~H. Jin, A.~D.~N. Initiative \emph{et~al.}, ``Predicting cognitive decline with deep learning of brain metabolism and amyloid imaging,'' \emph{Behavioural brain research}, vol. 344, pp. 103--109, 2018.

\bibitem{grueso2021machine}
S.~Grueso and R.~Viejo-Sobera, ``Machine learning methods for predicting progression from mild cognitive impairment to alzheimer’s disease dementia: a systematic review,'' \emph{Alzheimer's research \& therapy}, vol.~13, pp. 1--29, 2021.

\bibitem{warren2023functional}
S.~L. Warren and A.~A. Moustafa, ``Functional magnetic resonance imaging, deep learning, and alzheimer's disease: A systematic review,'' \emph{Journal of Neuroimaging}, vol.~33, no.~1, pp. 5--18, 2023.

\bibitem{ansart2021predicting}
M.~Ansart, S.~Epelbaum, G.~Bassignana, A.~B{\^o}ne, S.~Bottani, T.~Cattai, R.~Couronn{\'e}, J.~Faouzi, I.~Koval, M.~Louis \emph{et~al.}, ``Predicting the progression of mild cognitive impairment using machine learning: a systematic, quantitative and critical review,'' \emph{Medical Image Analysis}, vol.~67, p. 101848, 2021.

\bibitem{ALBERDI20161}
\BIBentryALTinterwordspacing
A.~Alberdi, A.~Aztiria, and A.~Basarab, ``On the early diagnosis of alzheimer's disease from multimodal signals: A survey,'' \emph{Artificial Intelligence in Medicine}, vol.~71, pp. 1--29, 2016. [Online]. Available: \url{https://www.sciencedirect.com/science/article/pii/S0933365716300732}
\BIBentrySTDinterwordspacing

\bibitem{10129131}
V.~Skaramagkas, A.~Pentari, Z.~Kefalopoulou, and M.~Tsiknakis, ``Multi-modal deep learning diagnosis of parkinson’s disease—a systematic review,'' \emph{IEEE Transactions on Neural Systems and Rehabilitation Engineering}, vol.~31, pp. 2399--2423, 2023.

\bibitem{yang2022deep}
Q.~Yang, X.~Li, X.~Ding, F.~Xu, and Z.~Ling, ``Deep learning-based speech analysis for alzheimer’s disease detection: a literature review,'' \emph{Alzheimer's Research \& Therapy}, vol.~14, no.~1, p. 186, 2022.

\bibitem{10.3389/fnagi.2023.1224723}
\BIBentryALTinterwordspacing
X.~Qi, Q.~Zhou, J.~Dong, and W.~Bao, ``Noninvasive automatic detection of alzheimer's disease from spontaneous speech: a review,'' \emph{Frontiers in Aging Neuroscience}, vol.~15, 2023. [Online]. Available: \url{https://www.frontiersin.org/journals/aging-neuroscience/articles/10.3389/fnagi.2023.1224723}
\BIBentrySTDinterwordspacing

\bibitem{gauthier2006mild}
S.~Gauthier, B.~Reisberg, M.~Zaudig, R.~C. Petersen, K.~Ritchie, K.~Broich, S.~Belleville, H.~Brodaty, D.~Bennett, H.~Chertkow \emph{et~al.}, ``Mild cognitive impairment,'' \emph{The lancet}, vol. 367, no. 9518, pp. 1262--1270, 2006.

\bibitem{petersen2016mild}
R.~C. Petersen, ``Mild cognitive impairment,'' \emph{CONTINUUM: lifelong Learning in Neurology}, vol.~22, no.~2, pp. 404--418, 2016.

\bibitem{rosenberg2011neuropsychiatric}
P.~B. Rosenberg, M.~M. Mielke, B.~Appleby, E.~Oh, J.-M. Leoutsakos, and C.~G. Lyketsos, ``Neuropsychiatric symptoms in mci subtypes: the importance of executive dysfunction,'' \emph{International journal of geriatric psychiatry}, vol.~26, no.~4, pp. 364--372, 2011.

\bibitem{arvanitakis2019diagnosis}
Z.~Arvanitakis, R.~C. Shah, and D.~A. Bennett, ``Diagnosis and management of dementia,'' \emph{Jama}, vol. 322, no.~16, pp. 1589--1599, 2019.

\bibitem{geldmacher1996evaluation}
D.~S. Geldmacher and P.~J. Whitehouse, ``Evaluation of dementia,'' \emph{New England Journal of Medicine}, vol. 335, no.~5, pp. 330--336, 1996.

\bibitem{ferris2013language}
S.~H. Ferris and M.~Farlow, ``Language impairment in alzheimer’s disease and benefits of acetylcholinesterase inhibitors,'' \emph{Clinical interventions in aging}, pp. 1007--1014, 2013.

\bibitem{kirshner2021aphasia}
H.~S. Kirshner and S.~M. Wilson, ``Aphasia and aphasic syndromes,'' \emph{Bradley’s Neurology in Clinical Practice E-Book}, vol. 133, 2021.

\bibitem{alexander2008aphasia}
M.~P. Alexander and A.~E. Hillis, ``Aphasia,'' \emph{Handbook of clinical neurology}, vol.~88, pp. 287--309, 2008.

\bibitem{kalia2015parkinson}
L.~V. Kalia and A.~E. Lang, ``Parkinson's disease,'' \emph{The Lancet}, vol. 386, no. 9996, pp. 896--912, 2015.

\bibitem{aarsland2004rate}
D.~Aarsland, K.~Andersen, J.~P. Larsen, R.~Perry, T.~Wentzel-Larsen, A.~Lolk, and P.~Kragh-S{\o}rensen, ``The rate of cognitive decline in parkinson disease,'' \emph{Archives of neurology}, vol.~61, no.~12, pp. 1906--1911, 2004.

\bibitem{aarsland2017cognitive}
D.~Aarsland, B.~Creese, M.~Politis, K.~R. Chaudhuri, D.~H. Ffytche, D.~Weintraub, and C.~Ballard, ``Cognitive decline in parkinson disease,'' \emph{Nature Reviews Neurology}, vol.~13, no.~4, pp. 217--231, 2017.

\bibitem{van2005apathy}
R.~Van~Reekum, D.~T. Stuss, and L.~Ostrander, ``Apathy: why care?'' \emph{The Journal of neuropsychiatry and clinical neurosciences}, vol.~17, no.~1, pp. 7--19, 2005.

\bibitem{montoya2019impact}
G.~Montoya-Murillo, N.~Ibarretxe-Bilbao, J.~Pe{\~n}a, and N.~Ojeda, ``The impact of apathy on cognitive performance in the elderly,'' \emph{International Journal of Geriatric Psychiatry}, vol.~34, no.~5, pp. 657--665, 2019.

\bibitem{konstantakopoulos2011apathy}
G.~Konstantakopoulos, D.~Ploumpidis, P.~Oulis, P.~Patrikelis, A.~Soumani, G.~N. Papadimitriou, and A.~M. Politis, ``Apathy, cognitive deficits and functional impairment in schizophrenia,'' \emph{Schizophrenia research}, vol. 133, no. 1-3, pp. 193--198, 2011.

\bibitem{gluhm2013cognitive}
S.~Gluhm, J.~Goldstein, K.~Loc, A.~Colt, C.~Van~Liew, and J.~Corey-Bloom, ``Cognitive performance on the mini-mental state examination and the montreal cognitive assessment across the healthy adult lifespan,'' \emph{Cognitive and Behavioral Neurology}, vol.~26, no.~1, pp. 1--5, 2013.

\bibitem{arevalo}
\BIBentryALTinterwordspacing
I.~Arevalo-Rodriguez, N.~Smailagic, M.~R. i~Figuls, A.~Ciapponi, E.~Sanchez-Perez, A.~Giannakou, O.~L. Pedraza, X.~B. Cosp, and S.~Cullum, ``Mini‐mental state examination (mmse) for the detection of alzheimer's disease and other dementias in people with mild cognitive impairment (mci),'' \emph{Cochrane Database of Systematic Reviews}, vol.~3, 2015. [Online]. Available: \url{https://doi.org//10.1002/14651858.CD010783.pub2}
\BIBentrySTDinterwordspacing

\bibitem{shiroky2007can}
J.~S. Shiroky, H.~M. Schipper, H.~Bergman, and H.~Chertkow, ``Can you have dementia with an mmse score of 30?'' \emph{American Journal of Alzheimer's Disease \& Other Dementias{\textregistered}}, vol.~22, no.~5, pp. 406--415, 2007.

\bibitem{trzepacz2015relationship}
P.~T. Trzepacz, H.~Hochstetler, S.~Wang, B.~Walker, A.~J. Saykin, and A.~D.~N. Initiative, ``Relationship between the montreal cognitive assessment and mini-mental state examination for assessment of mild cognitive impairment in older adults,'' \emph{BMC geriatrics}, vol.~15, pp. 1--9, 2015.

\bibitem{hoops2009validity}
S.~Hoops, S.~Nazem, A.~Siderowf, J.~Duda, S.~Xie, M.~Stern, and D.~Weintraub, ``Validity of the moca and mmse in the detection of mci and dementia in parkinson disease,'' \emph{Neurology}, vol.~73, no.~21, pp. 1738--1745, 2009.

\bibitem{vaswani2023attention}
A.~Vaswani, N.~Shazeer, N.~Parmar, J.~Uszkoreit, L.~Jones, A.~N. Gomez, L.~Kaiser, and I.~Polosukhin, ``Attention is all you need,'' 2023.

\bibitem{moher2010preferred}
D.~Moher, A.~Liberati, J.~Tetzlaff, D.~G. Altman, P.~Group \emph{et~al.}, ``Preferred reporting items for systematic reviews and meta-analyses: the prisma statement,'' \emph{International journal of surgery}, vol.~8, no.~5, pp. 336--341, 2010.

\bibitem{lanzi2023dementiabank}
A.~M. Lanzi, A.~K. Saylor, D.~Fromm, H.~Liu, B.~MacWhinney, and M.~L. Cohen, ``Dementiabank: Theoretical rationale, protocol, and illustrative analyses,'' \emph{American Journal of Speech-Language Pathology}, vol.~32, no.~2, pp. 426--438, 2023.

\bibitem{becker1994natural}
J.~T. Becker, F.~Boiler, O.~L. Lopez, J.~Saxton, and K.~L. McGonigle, ``The natural history of alzheimer's disease: description of study cohort and accuracy of diagnosis,'' \emph{Archives of neurology}, vol.~51, no.~6, pp. 585--594, 1994.

\bibitem{doi:10.1080/02687039608248419}
\BIBentryALTinterwordspacing
K.~P. Elaine~Giles and J.~R. Hodges, ``Performance on the boston cookie theft picture description task in patients with early dementia of the alzheimer's type: Missing information,'' \emph{Aphasiology}, vol.~10, no.~4, pp. 395--408, 1996. [Online]. Available: \url{https://doi.org/10.1080/02687039608248419}
\BIBentrySTDinterwordspacing

\bibitem{luz2020alzheimers}
S.~Luz, F.~Haider, S.~de~la Fuente, D.~Fromm, and B.~MacWhinney, ``Alzheimer's dementia recognition through spontaneous speech: The adress challenge,'' 2020.

\bibitem{luz2021detecting}
------, ``Detecting cognitive decline using speech only: The adresso challenge,'' 2021.

\bibitem{luz2023multilingual}
S.~Luz, F.~Haider, D.~Fromm, I.~Lazarou, I.~Kompatsiaris, and B.~MacWhinney, ``Multilingual alzheimer's dementia recognition through spontaneous speech: a signal processing grand challenge,'' 2023.

\bibitem{garcia2024connected}
S.~D. L.~F. Garcia, F.~Haider, D.~Fromm, B.~MacWhinney, A.~Lanzi, Y.-N. Chang, C.-J. Chou, Y.-C. Liu \emph{et~al.}, ``Connected speech-based cognitive assessment in chinese and english,'' \emph{arXiv preprint arXiv:2406.10272}, 2024.

\bibitem{macwhinney2011aphasiabank}
B.~MacWhinney, D.~Fromm, M.~Forbes, and A.~Holland, ``Aphasiabank: Methods for studying discourse,'' \emph{Aphasiology}, vol.~25, no.~11, pp. 1286--1307, 2011.

\bibitem{PopeDavis2011143161}
\BIBentryALTinterwordspacing
C.~Pope and B.~H. Davis, ``Finding a balance: The carolinas conversation collection,'' \emph{Corpus Linguistics and Linguistic Theory}, vol.~7, no.~1, pp. 143--161, 2011. [Online]. Available: \url{https://doi.org/10.1515/cllt.2011.007}
\BIBentrySTDinterwordspacing

\bibitem{li-etal-2020-analysis}
\BIBentryALTinterwordspacing
R.~A. Li, I.~Hajjar, F.~Goldstein, and J.~D. Choi, ``Analysis of hierarchical multi-content text classification model on {B}-{SHARP} dataset for early detection of {A}lzheimer{'}s disease,'' in \emph{Proceedings of the 1st Conference of the Asia-Pacific Chapter of the Association for Computational Linguistics and the 10th International Joint Conference on Natural Language Processing}, K.-F. Wong, K.~Knight, and H.~Wu, Eds.\hskip 1em plus 0.5em minus 0.4em\relax Suzhou, China: Association for Computational Linguistics, Dec. 2020, pp. 358--365. [Online]. Available: \url{https://aclanthology.org/2020.aacl-main.38}
\BIBentrySTDinterwordspacing

\bibitem{carr2019successfully}
D.~Carr, ``How to successfully navigate a revise-and-resubmit decision and handle rejections,'' \emph{Innovation in Aging}, vol.~3, no. Suppl 1, p. S224, 2019.

\bibitem{Yu2021}
K.~Yu, K.~Wild, K.~Potempa, B.~M. Hampstead, P.~A. Lichtenberg, L.~M. Struble, P.~Pruitt, E.~L. Alfaro, J.~Lindsley, M.~MacDonald, J.~A. Kaye, L.~C. Silbert, and H.~H. Dodge, ``The internet-based conversational engagement clinical trial (i-conect) in socially isolated adults 75+ years old: Randomized controlled trial protocol and covid-19 related study modifications,'' \emph{Frontiers in Digital Health}, vol.~3, 2021.

\bibitem{wu2022can}
C.-Y. Wu, N.~Mattek, K.~Wild, L.~M. Miller, J.~A. Kaye, L.~C. Silbert, and H.~H. Dodge, ``Can changes in social contact (frequency and mode) mitigate low mood before and during the covid-19 pandemic? the i-conect project,'' \emph{Journal of the American Geriatrics Society}, vol.~70, no.~3, pp. 669--676, 2022.

\bibitem{orozco-arroyave-etal-2014-new}
\BIBentryALTinterwordspacing
J.~R. Orozco-Arroyave, J.~D. Arias-Londo{\~n}o, J.~F. Vargas-Bonilla, M.~C. Gonz{\'a}lez-R{\'a}tiva, and E.~N{\"o}th, ``New {S}panish speech corpus database for the analysis of people suffering from {P}arkinson{'}s disease,'' in \emph{Proceedings of the Ninth International Conference on Language Resources and Evaluation ({LREC}'14)}, N.~Calzolari, K.~Choukri, T.~Declerck, H.~Loftsson, B.~Maegaard, J.~Mariani, A.~Moreno, J.~Odijk, and S.~Piperidis, Eds.\hskip 1em plus 0.5em minus 0.4em\relax Reykjavik, Iceland: European Language Resources Association (ELRA), May 2014, pp. 342--347. [Online]. Available: \url{http://www.lrec-conf.org/proceedings/lrec2014/pdf/7_Paper.pdf}
\BIBentrySTDinterwordspacing

\bibitem{karakostas2016demcareexperimentsdatasetstechnical}
\BIBentryALTinterwordspacing
A.~Karakostas, A.~Briassouli, K.~Avgerinakis, I.~Kompatsiaris, and M.~Tsolaki, ``The dem@care experiments and datasets: a technical report,'' 2016. [Online]. Available: \url{https://arxiv.org/abs/1701.01142}
\BIBentrySTDinterwordspacing

\bibitem{negin2018praxis}
F.~Negin, P.~Rodriguez, M.~Koperski, A.~Kerboua, J.~Gonz{\`a}lez, J.~Bourgeois, E.~Chapoulie, P.~Robert, and F.~Bremond, ``Praxis: Towards automatic cognitive assessment using gesture recognition,'' \emph{Expert Systems with Applications}, 2018.

\bibitem{egas2022automatic}
J.~V. Egas-L{\'o}pez, R.~Balogh, N.~Imre, I.~Hoffmann, M.~K. Szab{\'o}, L.~T{\'o}th, M.~P{\'a}k{\'a}ski, J.~K{\'a}lm{\'a}n, and G.~Gosztolya, ``Automatic screening of mild cognitive impairment and alzheimer’s disease by means of posterior-thresholding hesitation representation,'' \emph{Computer Speech \& Language}, vol.~75, p. 101377, 2022.

\bibitem{themistocleous2020voice}
C.~Themistocleous, M.~Eckerstr{\"o}m, and D.~Kokkinakis, ``Voice quality and speech fluency distinguish individuals with mild cognitive impairment from healthy controls,'' \emph{Plos one}, vol.~15, no.~7, p. e0236009, 2020.

\bibitem{abbaschian2021deep}
B.~J. Abbaschian, D.~Sierra-Sosa, and A.~Elmaghraby, ``Deep learning techniques for speech emotion recognition, from databases to models,'' \emph{Sensors}, vol.~21, no.~4, p. 1249, 2021.

\bibitem{koolagudi2012emotion}
S.~G. Koolagudi and K.~S. Rao, ``Emotion recognition from speech: a review,'' \emph{International journal of speech technology}, vol.~15, pp. 99--117, 2012.

\bibitem{wani2021comprehensive}
T.~M. Wani, T.~S. Gunawan, S.~A.~A. Qadri, M.~Kartiwi, and E.~Ambikairajah, ``A comprehensive review of speech emotion recognition systems,'' \emph{IEEE access}, vol.~9, pp. 47\,795--47\,814, 2021.

\bibitem{ortiz2023deep}
D.~Ortiz-Perez, P.~Ruiz-Ponce, J.~Rodr{\'\i}guez-Juan, D.~Tom{\'a}s, J.~Garcia-Rodriguez, and G.~J. Nalepa, ``Deep learning-based emotion detection in aphasia patients,'' in \emph{International Conference on Soft Computing Models in Industrial and Environmental Applications}.\hskip 1em plus 0.5em minus 0.4em\relax Springer, 2023, pp. 195--204.

\bibitem{code1999emotional}
C.~Code, G.~Hemsley, and M.~Herrmann, ``The emotional impact of aphasia,'' in \emph{Seminars in speech and language}, vol.~20.\hskip 1em plus 0.5em minus 0.4em\relax {\copyright} 1999 by Thieme Medical Publishers, Inc., 1999, pp. 19--31.

\bibitem{app13074244}
\BIBentryALTinterwordspacing
P.~Priyadarshinee, C.~J. Clarke, J.~Melechovsky, C.~M.~Y. Lin, B.~B.~T., and J.-M. Chen, ``Alzheimer’s dementia speech (audio vs. text): Multi-modal machine learning at high vs. low resolution,'' \emph{Applied Sciences}, vol.~13, no.~7, 2023. [Online]. Available: \url{https://www.mdpi.com/2076-3417/13/7/4244}
\BIBentrySTDinterwordspacing

\bibitem{Cui_2023}
Z.~Cui, W.~Wu, W.-Q. Zhang, J.~Wu, and C.~Zhang, ``Transferring speech-generic and depression-specific knowledge for alzheimer’s disease detection,'' in \emph{2023 IEEE Automatic Speech Recognition and Understanding Workshop (ASRU)}.\hskip 1em plus 0.5em minus 0.4em\relax IEEE, 2023, pp. 1--8.

\bibitem{wang2021modular}
N.~Wang, Y.~Cao, S.~Hao, Z.~Shao, and K.~Subbalakshmi, ``Modular multi-modal attention network for alzheimer's disease detection using patient audio and language data.'' in \emph{Interspeech}, 2021, pp. 3835--3839.

\bibitem{ying2023multimodal}
Y.~Ying, T.~Yang, and H.~Zhou, ``Multimodal fusion for alzheimer’s disease recognition,'' \emph{Applied Intelligence}, vol.~53, no.~12, pp. 16\,029--16\,040, 2023.

\bibitem{bang2024alzheimer}
J.-U. Bang, S.-H. Han, and B.-O. Kang, ``Alzheimer's disease recognition from spontaneous speech using large language models,'' \emph{ETRI Journal}, 2024.

\bibitem{DBLP:journals/corr/abs-2106-15684}
\BIBentryALTinterwordspacing
M.~Rohanian, J.~Hough, and M.~Purver, ``Alzheimer's dementia recognition using acoustic, lexical, disfluency and speech pause features robust to noisy inputs,'' \emph{CoRR}, vol. abs/2106.15684, 2021. [Online]. Available: \url{https://arxiv.org/abs/2106.15684}
\BIBentrySTDinterwordspacing

\bibitem{10307469}
S.~B. Shah, A.~Bhandari, and P.~G. Shambharkar, ``Leveraging multimodal information in speech data for the non-invasive detection of alzheimer’s disease,'' in \emph{2023 14th International Conference on Computing Communication and Networking Technologies (ICCCNT)}, 2023, pp. 1--6.

\bibitem{hlédiková2022data}
A.~Hlédiková, D.~Woszczyk, A.~Akman, S.~Demetriou, and B.~Schuller, ``Data augmentation for dementia detection in spoken language,'' 2022.

\bibitem{interspeech2020}
J.~Koo, J.~Lee, J.~Pyo, Y.~Jo, and K.~Lee, ``Exploiting multi-modal features from pre-trained networks for alzheimer’s dementia recognition,'' in \emph{Interspeech}, 10 2020, pp. 2217--2221.

\bibitem{pompili2020inesc}
A.~Pompili, T.~Rolland, and A.~Abad, ``The inesc-id multi-modal system for the adress 2020 challenge,'' \emph{arXiv preprint arXiv:2005.14646}, 2020.

\bibitem{cummins2020comparison}
N.~Cummins, Y.~Pan, Z.~Ren, J.~Fritsch, V.~S. Nallanthighal, H.~Christensen, D.~Blackburn, B.~W. Schuller, M.~Magimai-Doss, H.~Strik \emph{et~al.}, ``A comparison of acoustic and linguistics methodologies for alzheimer’s dementia recognition,'' in \emph{Interspeech 2020}.\hskip 1em plus 0.5em minus 0.4em\relax ISCA-International Speech Communication Association, 2020, pp. 2182--2186.

\bibitem{mahajan2021acoustic}
P.~Mahajan and V.~Baths, ``Acoustic and language based deep learning approaches for alzheimer's dementia detection from spontaneous speech,'' \emph{Frontiers in Aging Neuroscience}, vol.~13, p. 623607, 2021.

\bibitem{10.3389/fcomp.2021.624683}
\BIBentryALTinterwordspacing
Y.~Zhu, X.~Liang, J.~A. Batsis, and R.~M. Roth, ``Exploring deep transfer learning techniques for alzheimer's dementia detection,'' \emph{Frontiers in Computer Science}, vol.~3, 2021. [Online]. Available: \url{https://www.frontiersin.org/articles/10.3389/fcomp.2021.624683}
\BIBentrySTDinterwordspacing

\bibitem{ILIAS2023101485}
\BIBentryALTinterwordspacing
L.~Ilias, D.~Askounis, and J.~Psarras, ``Detecting dementia from speech and transcripts using transformers,'' \emph{Computer Speech \& Language}, vol.~79, p. 101485, 2023. [Online]. Available: \url{https://www.sciencedirect.com/science/article/pii/S0885230823000049}
\BIBentrySTDinterwordspacing

\bibitem{meghanani2021exploration}
A.~Meghanani, C.~Anoop, and A.~Ramakrishnan, ``An exploration of log-mel spectrogram and mfcc features for alzheimer’s dementia recognition from spontaneous speech,'' in \emph{2021 IEEE spoken language technology workshop (SLT)}.\hskip 1em plus 0.5em minus 0.4em\relax IEEE, 2021, pp. 670--677.

\bibitem{9413566}
Z.~Liu, Z.~Guo, Z.~Ling, and Y.~Li, ``Detecting alzheimer’s disease from speech using neural networks with bottleneck features and data augmentation,'' in \emph{ICASSP 2021 - 2021 IEEE International Conference on Acoustics, Speech and Signal Processing (ICASSP)}, 2021, pp. 7323--7327.

\bibitem{ORTIZPEREZ2023126413}
\BIBentryALTinterwordspacing
D.~Ortiz-Perez, P.~Ruiz-Ponce, D.~Tomás, J.~Garcia-Rodriguez, M.~F. Vizcaya-Moreno, and M.~Leo, ``A deep learning-based multimodal architecture to predict signs of dementia,'' \emph{Neurocomputing}, vol. 548, p. 126413, 2023. [Online]. Available: \url{https://www.sciencedirect.com/science/article/pii/S0925231223005362}
\BIBentrySTDinterwordspacing

\bibitem{krstev2022multimodal}
I.~Krstev, M.~Pavikjevikj, M.~Toshevska, and S.~Gievska, ``Multimodal data fusion for automatic detection of alzheimer’s disease,'' in \emph{International Conference on Human-Computer Interaction}.\hskip 1em plus 0.5em minus 0.4em\relax Springer, 2022, pp. 79--94.

\bibitem{escobar2023deep}
D.~Escobar-Grisales, C.~D. R{\'\i}os-Urrego, and J.~R. Orozco-Arroyave, ``Deep learning and artificial intelligence applied to model speech and language in parkinson’s disease,'' \emph{Diagnostics}, vol.~13, no.~13, p. 2163, 2023.

\bibitem{ortizperez2024cognitive}
D.~Ortiz-Perez, J.~Garcia-Rodriguez, and D.~Tomás, ``Cognitive insights across languages: Enhancing multimodal interview analysis,'' in \emph{Interspeech 2024}, 2024, pp. 952--956.

\bibitem{POOR2024109199}
\BIBentryALTinterwordspacing
F.~F. Poor, H.~H. Dodge, and M.~H. Mahoor, ``A multimodal cross-transformer-based model to predict mild cognitive impairment using speech, language and vision,'' \emph{Computers in Biology and Medicine}, vol. 182, p. 109199, 2024. [Online]. Available: \url{https://www.sciencedirect.com/science/article/pii/S0010482524012848}
\BIBentrySTDinterwordspacing

\bibitem{9747054}
Z.~Sheng, Z.~Guo, X.~Li, Y.~Li, and Z.~Ling, ``Dementia detection by fusing speech and eye-tracking representation,'' in \emph{ICASSP 2022 - 2022 IEEE International Conference on Acoustics, Speech and Signal Processing (ICASSP)}, 2022, pp. 6457--6461.

\bibitem{9426437}
N.~Narendra, B.~Schuller, and P.~Alku, ``The detection of parkinson's disease from speech using voice source information,'' \emph{IEEE/ACM Transactions on Audio, Speech, and Language Processing}, vol.~29, pp. 1925--1936, 2021.

\bibitem{10054791}
Y.~Ge, T.~Wang, J.~Cao, and S.~Xu, ``A novel multi-task learning based automatic speech impairment assessment algorithm,'' in \emph{2022 China Automation Congress (CAC)}, 2022, pp. 887--892.

\bibitem{9659979}
S.~Allamy and A.~L. Koerich, ``1d cnn architectures for music genre classification,'' in \emph{2021 IEEE Symposium Series on Computational Intelligence (SSCI)}, 2021, pp. 01--07.

\bibitem{ABDOLI2019252}
\BIBentryALTinterwordspacing
S.~Abdoli, P.~Cardinal, and A.~{Lameiras Koerich}, ``End-to-end environmental sound classification using a 1d convolutional neural network,'' \emph{Expert Systems with Applications}, vol. 136, pp. 252--263, 2019. [Online]. Available: \url{https://www.sciencedirect.com/science/article/pii/S0957417419304403}
\BIBentrySTDinterwordspacing

\bibitem{zaman2023survey}
K.~Zaman, M.~Sah, C.~Direkoglu, and M.~Unoki, ``A survey of audio classification using deep learning,'' \emph{IEEE Access}, 2023.

\bibitem{costa2017evaluation}
Y.~M. Costa, L.~S. Oliveira, and C.~N. Silla~Jr, ``An evaluation of convolutional neural networks for music classification using spectrograms,'' \emph{Applied soft computing}, vol.~52, pp. 28--38, 2017.

\bibitem{satt2017efficient}
A.~Satt, S.~Rozenberg, R.~Hoory \emph{et~al.}, ``Efficient emotion recognition from speech using deep learning on spectrograms.'' in \emph{Interspeech}, 2017, pp. 1089--1093.

\bibitem{zeng2019spectrogram}
Y.~Zeng, H.~Mao, D.~Peng, and Z.~Yi, ``Spectrogram based multi-task audio classification,'' \emph{Multimedia Tools and Applications}, vol.~78, pp. 3705--3722, 2019.

\bibitem{howard2013acoustics}
D.~Howard and J.~Angus, \emph{Acoustics and psychoacoustics}.\hskip 1em plus 0.5em minus 0.4em\relax Routledge, 2013.

\bibitem{moore2012introduction}
B.~C. Moore, \emph{An introduction to the psychology of hearing}.\hskip 1em plus 0.5em minus 0.4em\relax Brill, 2012.

\bibitem{10.1121/1.1915893}
\BIBentryALTinterwordspacing
S.~S. Stevens, J.~Volkmann, and E.~B. Newman, ``{A Scale for the Measurement of the Psychological Magnitude Pitch},'' \emph{The Journal of the Acoustical Society of America}, vol.~8, no.~3, pp. 185--190, 01 1937. [Online]. Available: \url{https://doi.org/10.1121/1.1915893}
\BIBentrySTDinterwordspacing

\bibitem{arias2021multi}
T.~Arias-Vergara, P.~Klumpp, J.~C. Vasquez-Correa, E.~N{\"o}th, J.~R. Orozco-Arroyave, and M.~Schuster, ``Multi-channel spectrograms for speech processing applications using deep learning methods,'' \emph{Pattern Analysis and Applications}, vol.~24, pp. 423--431, 2021.

\bibitem{8678825}
H.~Purwins, B.~Li, T.~Virtanen, J.~Schlüter, S.-Y. Chang, and T.~Sainath, ``Deep learning for audio signal processing,'' \emph{IEEE Journal of Selected Topics in Signal Processing}, vol.~13, no.~2, pp. 206--219, 2019.

\bibitem{jiao2019survey}
L.~Jiao and J.~Zhao, ``A survey on the new generation of deep learning in image processing,'' \emph{Ieee Access}, vol.~7, pp. 172\,231--172\,263, 2019.

\bibitem{minaee2021image}
S.~Minaee, Y.~Boykov, F.~Porikli, A.~Plaza, N.~Kehtarnavaz, and D.~Terzopoulos, ``Image segmentation using deep learning: A survey,'' \emph{IEEE transactions on pattern analysis and machine intelligence}, vol.~44, no.~7, pp. 3523--3542, 2021.

\bibitem{gu2018recent}
J.~Gu, Z.~Wang, J.~Kuen, L.~Ma, A.~Shahroudy, B.~Shuai, T.~Liu, X.~Wang, G.~Wang, J.~Cai \emph{et~al.}, ``Recent advances in convolutional neural networks,'' \emph{Pattern recognition}, vol.~77, pp. 354--377, 2018.

\bibitem{o2015introduction}
K.~O'shea and R.~Nash, ``An introduction to convolutional neural networks,'' \emph{arXiv preprint arXiv:1511.08458}, 2015.

\bibitem{li2021survey}
Z.~Li, F.~Liu, W.~Yang, S.~Peng, and J.~Zhou, ``A survey of convolutional neural networks: analysis, applications, and prospects,'' \emph{IEEE transactions on neural networks and learning systems}, vol.~33, no.~12, pp. 6999--7019, 2021.

\bibitem{DBLP:journals/corr/abs-1810-04805}
\BIBentryALTinterwordspacing
J.~Devlin, M.~Chang, K.~Lee, and K.~Toutanova, ``{BERT:} pre-training of deep bidirectional transformers for language understanding,'' \emph{CoRR}, vol. abs/1810.04805, 2018. [Online]. Available: \url{http://arxiv.org/abs/1810.04805}
\BIBentrySTDinterwordspacing

\bibitem{DBLP:journals/corr/abs-1910-01108}
\BIBentryALTinterwordspacing
V.~Sanh, L.~Debut, J.~Chaumond, and T.~Wolf, ``Distilbert, a distilled version of {BERT:} smaller, faster, cheaper and lighter,'' \emph{CoRR}, vol. abs/1910.01108, 2019. [Online]. Available: \url{http://arxiv.org/abs/1910.01108}
\BIBentrySTDinterwordspacing

\bibitem{DBLP:journals/corr/abs-2103-15691}
\BIBentryALTinterwordspacing
A.~Arnab, M.~Dehghani, G.~Heigold, C.~Sun, M.~Lucic, and C.~Schmid, ``Vivit: {A} video vision transformer,'' \emph{CoRR}, vol. abs/2103.15691, 2021. [Online]. Available: \url{https://arxiv.org/abs/2103.15691}
\BIBentrySTDinterwordspacing

\bibitem{DBLP:journals/corr/abs-2006-11477}
\BIBentryALTinterwordspacing
A.~Baevski, H.~Zhou, A.~Mohamed, and M.~Auli, ``wav2vec 2.0: {A} framework for self-supervised learning of speech representations,'' \emph{CoRR}, vol. abs/2006.11477, 2020. [Online]. Available: \url{https://arxiv.org/abs/2006.11477}
\BIBentrySTDinterwordspacing

\bibitem{DBLP:journals/corr/abs-2010-11929}
\BIBentryALTinterwordspacing
A.~Dosovitskiy, L.~Beyer, A.~Kolesnikov, D.~Weissenborn, X.~Zhai, T.~Unterthiner, M.~Dehghani, M.~Minderer, G.~Heigold, S.~Gelly, J.~Uszkoreit, and N.~Houlsby, ``An image is worth 16x16 words: Transformers for image recognition at scale,'' \emph{CoRR}, vol. abs/2010.11929, 2020. [Online]. Available: \url{https://arxiv.org/abs/2010.11929}
\BIBentrySTDinterwordspacing

\bibitem{pons2017timbre}
J.~Pons, O.~Slizovskaia, R.~Gong, E.~G{\'o}mez, and X.~Serra, ``Timbre analysis of music audio signals with convolutional neural networks,'' in \emph{2017 25th European Signal Processing Conference (EUSIPCO)}.\hskip 1em plus 0.5em minus 0.4em\relax IEEE, 2017, pp. 2744--2748.

\bibitem{zhang2022spectrogram}
Y.~Zhang, B.~Li, H.~Fang, and Q.~Meng, ``Spectrogram transformers for audio classification,'' in \emph{2022 IEEE International Conference on Imaging Systems and Techniques (IST)}.\hskip 1em plus 0.5em minus 0.4em\relax IEEE, 2022, pp. 1--6.

\bibitem{gong2021ast}
Y.~Gong, Y.-A. Chung, and J.~Glass, ``Ast: Audio spectrogram transformer,'' \emph{arXiv preprint arXiv:2104.01778}, 2021.

\bibitem{radford2022robust}
A.~Radford, J.~W. Kim, T.~Xu, G.~Brockman, C.~McLeavey, and I.~Sutskever, ``Robust speech recognition via large-scale weak supervision,'' in \emph{International conference on machine learning}.\hskip 1em plus 0.5em minus 0.4em\relax PMLR, 2023, pp. 28\,492--28\,518.

\bibitem{bertinetto2016fully}
L.~Bertinetto, J.~Valmadre, J.~F. Henriques, A.~Vedaldi, and P.~H. Torr, ``Fully-convolutional siamese networks for object tracking,'' in \emph{Computer Vision--ECCV 2016 Workshops: Amsterdam, The Netherlands, October 8-10 and 15-16, 2016, Proceedings, Part II 14}.\hskip 1em plus 0.5em minus 0.4em\relax Springer, 2016, pp. 850--865.

\bibitem{boelders2020detection}
S.~Boelders, V.~S. Nallanthighal, V.~Menkovski, and A.~H{\"a}rm{\"a}, ``Detection of mild dyspnea from pairs of speech recordings,'' in \emph{ICASSP 2020-2020 IEEE International Conference on Acoustics, Speech and Signal Processing (ICASSP)}.\hskip 1em plus 0.5em minus 0.4em\relax IEEE, 2020, pp. 4102--4106.

\bibitem{lian2018speech}
Z.~Lian, Y.~Li, J.~Tao, and J.~Huang, ``Speech emotion recognition via contrastive loss under siamese networks,'' in \emph{Proceedings of the Joint Workshop of the 4th Workshop on Affective Social Multimedia Computing and First Multi-Modal Affective Computing of Large-Scale Multimedia Data}, 2018, pp. 21--26.

\bibitem{wang2019child}
J.~Wang, Y.~Qin, Z.~Peng, and T.~Lee, ``Child speech disorder detection with siamese recurrent network using speech attribute features.'' in \emph{INTERSPEECH}, vol.~2, 2019, pp. 3885--3889.

\bibitem{10.1007/978-3-031-18050-7_25}
D.~Ortiz-Perez, P.~Ruiz-Ponce, D.~Tom{\'a}s, and J.~Garcia-Rodriguez, ``Deep learning-based dementia prediction using multimodal data,'' in \emph{17th International Conference on Soft Computing Models in Industrial and Environmental Applications (SOCO 2022)}, P.~Garc{\'i}a~Bringas, H.~P{\'e}rez~Garc{\'i}a, F.~J. Martinez-de Pison, J.~R. Villar~Flecha, A.~Troncoso~Lora, E.~A. de~la Cal, {\'A}.~Herrero, F.~Mart{\'i}nez~{\'A}lvarez, G.~Psaila, H.~Quinti{\'a}n, and E.~S. Corchado~Rodriguez, Eds.\hskip 1em plus 0.5em minus 0.4em\relax Cham: Springer Nature Switzerland, 2023, pp. 260--269.

\bibitem{DBLP:journals/corr/HuangLW16a}
\BIBentryALTinterwordspacing
G.~Huang, Z.~Liu, and K.~Q. Weinberger, ``Densely connected convolutional networks,'' \emph{CoRR}, vol. abs/1608.06993, 2016. [Online]. Available: \url{http://arxiv.org/abs/1608.06993}
\BIBentrySTDinterwordspacing

\bibitem{DBLP:journals/corr/HowardZCKWWAA17}
\BIBentryALTinterwordspacing
A.~G. Howard, M.~Zhu, B.~Chen, D.~Kalenichenko, W.~Wang, T.~Weyand, M.~Andreetto, and H.~Adam, ``Mobilenets: Efficient convolutional neural networks for mobile vision applications,'' \emph{CoRR}, vol. abs/1704.04861, 2017. [Online]. Available: \url{http://arxiv.org/abs/1704.04861}
\BIBentrySTDinterwordspacing

\bibitem{DBLP:journals/corr/HeZRS15}
\BIBentryALTinterwordspacing
K.~He, X.~Zhang, S.~Ren, and J.~Sun, ``Deep residual learning for image recognition,'' \emph{CoRR}, vol. abs/1512.03385, 2015. [Online]. Available: \url{http://arxiv.org/abs/1512.03385}
\BIBentrySTDinterwordspacing

\bibitem{iandola2016squeezenet}
F.~N. Iandola, S.~Han, M.~W. Moskewicz, K.~Ashraf, W.~J. Dally, and K.~Keutzer, ``Squeezenet: Alexnet-level accuracy with 50x fewer parameters and< 0.5 mb model size,'' \emph{arXiv preprint arXiv:1602.07360}, 2016.

\bibitem{simonyan2015deep}
K.~Simonyan and A.~Zisserman, ``Very deep convolutional networks for large-scale image recognition,'' 2015.

\bibitem{SHERSTINSKY2020132306}
\BIBentryALTinterwordspacing
A.~Sherstinsky, ``Fundamentals of recurrent neural network (rnn) and long short-term memory (lstm) network,'' \emph{Physica D: Nonlinear Phenomena}, vol. 404, p. 132306, 2020. [Online]. Available: \url{https://www.sciencedirect.com/science/article/pii/S0167278919305974}
\BIBentrySTDinterwordspacing

\bibitem{li2022mvitv2}
Y.~Li, C.-Y. Wu, H.~Fan, K.~Mangalam, B.~Xiong, J.~Malik, and C.~Feichtenhofer, ``Mvitv2: Improved multiscale vision transformers for classification and detection,'' 2022.

\bibitem{szegedy2014going}
C.~Szegedy, W.~Liu, Y.~Jia, P.~Sermanet, S.~Reed, D.~Anguelov, D.~Erhan, V.~Vanhoucke, and A.~Rabinovich, ``Going deeper with convolutions,'' 2014.

\bibitem{zagoruyko2017wide}
S.~Zagoruyko and N.~Komodakis, ``Wide residual networks,'' 2017.

\bibitem{NIPS2012_c399862d}
A.~Krizhevsky, I.~Sutskever, and G.~E. Hinton, ``Imagenet classification with deep convolutional neural networks,'' \emph{Advances in neural information processing systems}, vol.~25, 2012.

\bibitem{DBLP:journals/corr/abs-1801-04381}
\BIBentryALTinterwordspacing
M.~Sandler, A.~G. Howard, M.~Zhu, A.~Zhmoginov, and L.~Chen, ``Inverted residuals and linear bottlenecks: Mobile networks for classification, detection and segmentation,'' \emph{CoRR}, vol. abs/1801.04381, 2018. [Online]. Available: \url{http://arxiv.org/abs/1801.04381}
\BIBentrySTDinterwordspacing

\bibitem{DBLP:journals/corr/abs-1807-11626}
\BIBentryALTinterwordspacing
M.~Tan, B.~Chen, R.~Pang, V.~Vasudevan, and Q.~V. Le, ``Mnasnet: Platform-aware neural architecture search for mobile,'' \emph{CoRR}, vol. abs/1807.11626, 2018. [Online]. Available: \url{http://arxiv.org/abs/1807.11626}
\BIBentrySTDinterwordspacing

\bibitem{DBLP:journals/corr/abs-1905-11946}
\BIBentryALTinterwordspacing
M.~Tan and Q.~V. Le, ``Efficientnet: Rethinking model scaling for convolutional neural networks,'' \emph{CoRR}, vol. abs/1905.11946, 2019. [Online]. Available: \url{http://arxiv.org/abs/1905.11946}
\BIBentrySTDinterwordspacing

\bibitem{DBLP:journals/corr/abs-1910-11559}
\BIBentryALTinterwordspacing
Y.~Chuang, C.~Liu, and H.~Lee, ``Speechbert: Cross-modal pre-trained language model for end-to-end spoken question answering,'' \emph{CoRR}, vol. abs/1910.11559, 2019. [Online]. Available: \url{http://arxiv.org/abs/1910.11559}
\BIBentrySTDinterwordspacing

\bibitem{gupta2013feature}
S.~Gupta, J.~Jaafar, W.~W. Ahmad, and A.~Bansal, ``Feature extraction using mfcc,'' \emph{Signal \& Image Processing: An International Journal}, vol.~4, no.~4, pp. 101--108, 2013.

\bibitem{muda2010voice}
L.~Muda, M.~Begam, and I.~Elamvazuthi, ``Voice recognition algorithms using mel frequency cepstral coefficient (mfcc) and dynamic time warping (dtw) techniques,'' \emph{arXiv preprint arXiv:1003.4083}, 2010.

\bibitem{tiwari2010mfcc}
V.~Tiwari, ``Mfcc and its applications in speaker recognition,'' \emph{International journal on emerging technologies}, vol.~1, no.~1, pp. 19--22, 2010.

\bibitem{deng2020heart}
M.~Deng, T.~Meng, J.~Cao, S.~Wang, J.~Zhang, and H.~Fan, ``Heart sound classification based on improved mfcc features and convolutional recurrent neural networks,'' \emph{Neural Networks}, vol. 130, pp. 22--32, 2020.

\bibitem{rejaibi2022mfcc}
E.~Rejaibi, A.~Komaty, F.~Meriaudeau, S.~Agrebi, and A.~Othmani, ``Mfcc-based recurrent neural network for automatic clinical depression recognition and assessment from speech,'' \emph{Biomedical Signal Processing and Control}, vol.~71, p. 103107, 2022.

\bibitem{10.1145/1873951.1874246}
\BIBentryALTinterwordspacing
F.~Eyben, M.~W\"{o}llmer, and B.~Schuller, ``Opensmile: the munich versatile and fast open-source audio feature extractor,'' in \emph{Proceedings of the 18th ACM International Conference on Multimedia}, ser. MM '10.\hskip 1em plus 0.5em minus 0.4em\relax New York, NY, USA: Association for Computing Machinery, 2010, p. 1459–1462. [Online]. Available: \url{https://doi.org/10.1145/1873951.1874246}
\BIBentrySTDinterwordspacing

\bibitem{eyben2013recent}
F.~Eyben, F.~Weninger, F.~Gross, and B.~Schuller, ``Recent developments in opensmile, the munich open-source multimedia feature extractor,'' in \emph{Proceedings of the 21st ACM international conference on Multimedia}, 2013, pp. 835--838.

\bibitem{degottex2014covarep}
G.~Degottex, J.~Kane, T.~Drugman, T.~Raitio, and S.~Scherer, ``Covarep—a collaborative voice analysis repository for speech technologies,'' in \emph{2014 ieee international conference on acoustics, speech and signal processing (icassp)}.\hskip 1em plus 0.5em minus 0.4em\relax IEEE, 2014, pp. 960--964.

\bibitem{povey2011kaldi}
D.~Povey, A.~Ghoshal, G.~Boulianne, L.~Burget, O.~Glembek, N.~Goel, M.~Hannemann, P.~Motlicek, Y.~Qian, P.~Schwarz, J.~Silovsky, G.~Stemmer, and K.~Vesely, ``The kaldi speech recognition toolkit,'' in \emph{IEEE 2011 Workshop on Automatic Speech Recognition and Understanding}.\hskip 1em plus 0.5em minus 0.4em\relax IEEE Signal Processing Society, Dec. 2011, iEEE Catalog No.: CFP11SRW-USB.

\bibitem{nimitsurachat2024audio}
P.~Nimitsurachat and P.~Washington, ``Audio-based emotion recognition using self-supervised learning on an engineered feature space,'' \emph{AI}, vol.~5, no.~1, pp. 195--207, 2024.

\bibitem{zadeh2018multimodal}
A.~B. Zadeh, P.~P. Liang, S.~Poria, E.~Cambria, and L.-P. Morency, ``Multimodal language analysis in the wild: Cmu-mosei dataset and interpretable dynamic fusion graph,'' in \emph{Proceedings of the 56th Annual Meeting of the Association for Computational Linguistics (Volume 1: Long Papers)}, 2018, pp. 2236--2246.

\bibitem{qian2015bird}
K.~Qian, Z.~Zhang, F.~Ringeval, and B.~Schuller, ``Bird sounds classification by large scale acoustic features and extreme learning machine,'' in \emph{2015 IEEE Global Conference on Signal and Information Processing (GlobalSIP)}.\hskip 1em plus 0.5em minus 0.4em\relax IEEE, 2015, pp. 1317--1321.

\bibitem{narendra2018dysarthric}
N.~Narendra and P.~Alku, ``Dysarthric speech classification using glottal features computed from non-words, words and sentences.'' in \emph{Interspeech}, 2018, pp. 3403--3407.

\bibitem{schuller2010interspeech}
B.~Schuller, S.~Steidl, A.~Batliner, F.~Burkhardt, L.~Devillers, C.~M{\"u}ller, and S.~Narayanan, ``The interspeech 2010 paralinguistic challenge,'' in \emph{Proc. INTERSPEECH 2010, Makuhari, Japan}, 2010, pp. 2794--2797.

\bibitem{7160715}
F.~Eyben, K.~R. Scherer, B.~W. Schuller, J.~Sundberg, E.~André, C.~Busso, L.~Y. Devillers, J.~Epps, P.~Laukka, S.~S. Narayanan, and K.~P. Truong, ``The geneva minimalistic acoustic parameter set (gemaps) for voice research and affective computing,'' \emph{IEEE Transactions on Affective Computing}, vol.~7, no.~2, pp. 190--202, 2016.

\bibitem{schuller16_interspeech}
B.~Schuller, S.~Steidl, A.~Batliner, J.~Hirschberg, J.~K. Burgoon, A.~Baird, A.~Elkins, Y.~Zhang, E.~Coutinho, and K.~Evanini, ``{The INTERSPEECH 2016 Computational Paralinguistics Challenge: Deception, Sincerity \& Native Language},'' in \emph{Proc. Interspeech 2016}, 2016, pp. 2001--2005.

\bibitem{DBLP:journals/corr/HersheyCEGJMPPS16}
\BIBentryALTinterwordspacing
S.~Hershey, S.~Chaudhuri, D.~P.~W. Ellis, J.~F. Gemmeke, A.~Jansen, R.~C. Moore, M.~Plakal, D.~Platt, R.~A. Saurous, B.~Seybold, M.~Slaney, R.~J. Weiss, and K.~W. Wilson, ``{CNN} architectures for large-scale audio classification,'' \emph{CoRR}, vol. abs/1609.09430, 2016. [Online]. Available: \url{http://arxiv.org/abs/1609.09430}
\BIBentrySTDinterwordspacing

\bibitem{gemmeke2017audio}
J.~F. Gemmeke, D.~P. Ellis, D.~Freedman, A.~Jansen, W.~Lawrence, R.~C. Moore, M.~Plakal, and M.~Ritter, ``Audio set: An ontology and human-labeled dataset for audio events,'' in \emph{2017 IEEE international conference on acoustics, speech and signal processing (ICASSP)}.\hskip 1em plus 0.5em minus 0.4em\relax IEEE, 2017, pp. 776--780.

\bibitem{5545402}
N.~Dehak, P.~J. Kenny, R.~Dehak, P.~Dumouchel, and P.~Ouellet, ``Front-end factor analysis for speaker verification,'' \emph{IEEE Transactions on Audio, Speech, and Language Processing}, vol.~19, no.~4, pp. 788--798, 2011.

\bibitem{hauptman2019identifying}
Y.~Hauptman, R.~Aloni-Lavi, I.~Lapidot, T.~Gurevich, Y.~Manor, S.~Naor, N.~Diamant, and I.~Opher, ``Identifying distinctive acoustic and spectral features in parkinson's disease.'' in \emph{Interspeech}, 2019, pp. 2498--2502.

\bibitem{laaridh2017automatic}
I.~Laaridh, W.~B. Kheder, C.~Fredouille, and C.~Meunier, ``Automatic prediction of speech evaluation metrics for dysarthric speech,'' in \emph{Interspeech 2017}, 2017, pp. 1834--1838.

\bibitem{snyder2017deep}
D.~Snyder, D.~Garcia-Romero, D.~Povey, and S.~Khudanpur, ``Deep neural network embeddings for text-independent speaker verification.'' in \emph{Interspeech}, vol. 2017, 2017, pp. 999--1003.

\bibitem{snyder2018spoken}
D.~Snyder, D.~Garcia-Romero, A.~McCree, G.~Sell, D.~Povey, and S.~Khudanpur, ``Spoken language recognition using x-vectors.'' in \emph{Odyssey}, vol. 2018, 2018, pp. 105--111.

\bibitem{snyder2018x}
D.~Snyder, D.~Garcia-Romero, G.~Sell, D.~Povey, and S.~Khudanpur, ``X-vectors: Robust dnn embeddings for speaker recognition,'' in \emph{2018 IEEE international conference on acoustics, speech and signal processing (ICASSP)}.\hskip 1em plus 0.5em minus 0.4em\relax IEEE, 2018, pp. 5329--5333.

\bibitem{lin2022survey}
T.~Lin, Y.~Wang, X.~Liu, and X.~Qiu, ``A survey of transformers,'' \emph{AI open}, vol.~3, pp. 111--132, 2022.

\bibitem{han2022survey}
K.~Han, Y.~Wang, H.~Chen, X.~Chen, J.~Guo, Z.~Liu, Y.~Tang, A.~Xiao, C.~Xu, Y.~Xu \emph{et~al.}, ``A survey on vision transformer,'' \emph{IEEE transactions on pattern analysis and machine intelligence}, vol.~45, no.~1, pp. 87--110, 2022.

\bibitem{khan2022transformers}
S.~Khan, M.~Naseer, M.~Hayat, S.~W. Zamir, F.~S. Khan, and M.~Shah, ``Transformers in vision: A survey,'' \emph{ACM computing surveys (CSUR)}, vol.~54, no. 10s, pp. 1--41, 2022.

\bibitem{DBLP:journals/corr/abs-1904-05862}
\BIBentryALTinterwordspacing
S.~Schneider, A.~Baevski, R.~Collobert, and M.~Auli, ``wav2vec: Unsupervised pre-training for speech recognition,'' \emph{CoRR}, vol. abs/1904.05862, 2019. [Online]. Available: \url{http://arxiv.org/abs/1904.05862}
\BIBentrySTDinterwordspacing

\bibitem{zhuang2020music}
Y.~Zhuang, Y.~Chen, and J.~Zheng, ``Music genre classification with transformer classifier,'' in \emph{Proceedings of the 2020 4th international conference on digital signal processing}, 2020, pp. 155--159.

\bibitem{andayani2022hybrid}
F.~Andayani, L.~B. Theng, M.~T. Tsun, and C.~Chua, ``Hybrid lstm-transformer model for emotion recognition from speech audio files,'' \emph{IEEE Access}, vol.~10, pp. 36\,018--36\,027, 2022.

\bibitem{vaessen2022fine}
N.~Vaessen and D.~A. Van~Leeuwen, ``Fine-tuning wav2vec2 for speaker recognition,'' in \emph{ICASSP 2022-2022 IEEE International Conference on Acoustics, Speech and Signal Processing (ICASSP)}.\hskip 1em plus 0.5em minus 0.4em\relax IEEE, 2022, pp. 7967--7971.

\bibitem{lyketsos2003diagnosis}
C.~G. Lyketsos and H.~B. Lee, ``Diagnosis and treatment of depression in alzheimer’s disease: a practical update for the clinician,'' \emph{Dementia and geriatric cognitive disorders}, vol.~17, no. 1-2, pp. 55--64, 2003.

\bibitem{green2003depression}
R.~C. Green, L.~A. Cupples, A.~Kurz, S.~Auerbach, R.~Go, D.~Sadovnick, R.~Duara, W.~A. Kukull, H.~Chui, T.~Edeki \emph{et~al.}, ``Depression as a risk factor for alzheimer disease: the mirage study,'' \emph{Archives of neurology}, vol.~60, no.~5, pp. 753--759, 2003.

\bibitem{ownby2006depression}
R.~L. Ownby, E.~Crocco, A.~Acevedo, V.~John, and D.~Loewenstein, ``Depression and risk for alzheimer disease: systematic review, meta-analysis, and metaregression analysis,'' \emph{Archives of general psychiatry}, vol.~63, no.~5, pp. 530--538, 2006.

\bibitem{gratch2014distress}
J.~Gratch, R.~Artstein, G.~M. Lucas, G.~Stratou, S.~Scherer, A.~Nazarian, R.~Wood, J.~Boberg, D.~DeVault, S.~Marsella \emph{et~al.}, ``The distress analysis interview corpus of human and computer interviews.'' in \emph{LREC}.\hskip 1em plus 0.5em minus 0.4em\relax Reykjavik, 2014, pp. 3123--3128.

\bibitem{DBLP:journals/corr/abs-2110-13900}
\BIBentryALTinterwordspacing
S.~Chen, C.~Wang, Z.~Chen, Y.~Wu, S.~Liu, Z.~Chen, J.~Li, N.~Kanda, T.~Yoshioka, X.~Xiao, J.~Wu, L.~Zhou, S.~Ren, Y.~Qian, Y.~Qian, J.~Wu, M.~Zeng, and F.~Wei, ``Wavlm: Large-scale self-supervised pre-training for full stack speech processing,'' \emph{CoRR}, vol. abs/2110.13900, 2021. [Online]. Available: \url{https://arxiv.org/abs/2110.13900}
\BIBentrySTDinterwordspacing

\bibitem{DBLP:journals/corr/abs-2106-07447}
\BIBentryALTinterwordspacing
W.~Hsu, B.~Bolte, Y.~H. Tsai, K.~Lakhotia, R.~Salakhutdinov, and A.~Mohamed, ``Hubert: Self-supervised speech representation learning by masked prediction of hidden units,'' \emph{CoRR}, vol. abs/2106.07447, 2021. [Online]. Available: \url{https://arxiv.org/abs/2106.07447}
\BIBentrySTDinterwordspacing

\bibitem{KHEDDAR2023110851}
\BIBentryALTinterwordspacing
H.~Kheddar, Y.~Himeur, S.~Al-Maadeed, A.~Amira, and F.~Bensaali, ``Deep transfer learning for automatic speech recognition: Towards better generalization,'' \emph{Knowledge-Based Systems}, vol. 277, p. 110851, 2023. [Online]. Available: \url{https://www.sciencedirect.com/science/article/pii/S0950705123006019}
\BIBentrySTDinterwordspacing

\bibitem{jamal2017automatic}
\BIBentryALTinterwordspacing
N.~Jamal, S.~Shanta, F.~Mahmud, and M.~Sha’abani, ``{Automatic speech recognition (ASR) based approach for speech therapy of aphasic patients: A review},'' \emph{AIP Conference Proceedings}, vol. 1883, no.~1, p. 020028, 09 2017. [Online]. Available: \url{https://doi.org/10.1063/1.5002046}
\BIBentrySTDinterwordspacing

\bibitem{torre2021improving}
I.~G. Torre, M.~Romero, and A.~{\'A}lvarez, ``Improving aphasic speech recognition by using novel semi-supervised learning methods on aphasiabank for english and spanish,'' \emph{Applied Sciences}, vol.~11, no.~19, p. 8872, 2021.

\bibitem{weiner2017manual}
J.~Weiner, M.~Engelbart, and T.~Schultz, ``Manual and automatic transcriptions in dementia detection from speech.'' in \emph{Interspeech}, 2017, pp. 3117--3121.

\bibitem{zhu21e_interspeech}
Y.~Zhu, A.~Obyat, X.~Liang, J.~A. Batsis, and R.~M. Roth, ``{WavBERT: Exploiting Semantic and Non-Semantic Speech Using Wav2vec and BERT for Dementia Detection},'' in \emph{Proc. Interspeech 2021}, 2021, pp. 3790--3794.

\bibitem{zhang2023survey}
H.~Zhang, H.~Song, S.~Li, M.~Zhou, and D.~Song, ``A survey of controllable text generation using transformer-based pre-trained language models,'' \emph{ACM Computing Surveys}, vol.~56, no.~3, pp. 1--37, 2023.

\bibitem{li2024pre}
J.~Li, T.~Tang, W.~X. Zhao, J.-Y. Nie, and J.-R. Wen, ``Pre-trained language models for text generation: A survey,'' \emph{ACM Computing Surveys}, vol.~56, no.~9, pp. 1--39, 2024.

\bibitem{radford2019language}
A.~Radford, J.~Wu, R.~Child, D.~Luan, D.~Amodei, I.~Sutskever \emph{et~al.}, ``Language models are unsupervised multitask learners,'' \emph{OpenAI blog}, vol.~1, no.~8, p.~9, 2019.

\bibitem{li-etal-2022-gpt}
\BIBentryALTinterwordspacing
C.~Li, D.~Knopman, W.~Xu, T.~Cohen, and S.~Pakhomov, ``{GPT}-{D}: Inducing dementia-related linguistic anomalies by deliberate degradation of artificial neural language models,'' in \emph{Proceedings of the 60th Annual Meeting of the Association for Computational Linguistics (Volume 1: Long Papers)}, S.~Muresan, P.~Nakov, and A.~Villavicencio, Eds.\hskip 1em plus 0.5em minus 0.4em\relax Dublin, Ireland: Association for Computational Linguistics, May 2022, pp. 1866--1877. [Online]. Available: \url{https://aclanthology.org/2022.acl-long.131}
\BIBentrySTDinterwordspacing

\bibitem{yuan2020disfluencies}
J.~Yuan, Y.~Bian, X.~Cai, J.~Huang, Z.~Ye, and K.~Church, ``Disfluencies and fine-tuning pre-trained language models for detection of alzheimer's disease.'' in \emph{Interspeech}, vol. 2020, 2020, pp. 2162--6.

\bibitem{ilias2022explainable}
L.~Ilias and D.~Askounis, ``Explainable identification of dementia from transcripts using transformer networks,'' \emph{IEEE Journal of Biomedical and Health Informatics}, vol.~26, no.~8, pp. 4153--4164, 2022.

\bibitem{Balagopalan2020ToBO}
A.~Balagopalan, B.~Eyre, F.~Rudzicz, and J.~Novikova, ``To bert or not to bert: Comparing speech and language-based approaches for alzheimer’s disease detection,'' in \emph{Interspeech 2020}, 2020, pp. 2167--2171.

\bibitem{nambiar2022comparative}
A.~S. Nambiar, K.~Likhita, K.~S. Pujya, D.~Gupta, S.~Vekkot, and S.~Lalitha, ``Comparative study of deep classifiers for early dementia detection using speech transcripts,'' in \emph{2022 IEEE 19th India Council International Conference (INDICON)}.\hskip 1em plus 0.5em minus 0.4em\relax IEEE, 2022, pp. 1--6.

\bibitem{liu2023approach}
N.~Liu and L.~Wang, ``An approach for assisting diagnosis of alzheimer's disease based on natural language processing,'' \emph{Frontiers in Aging Neuroscience}, vol.~15, 2023.

\bibitem{roshanzamir2021transformer}
A.~Roshanzamir, H.~Aghajan, and M.~Soleymani~Baghshah, ``Transformer-based deep neural network language models for alzheimer’s disease risk assessment from targeted speech,'' \emph{BMC Medical Informatics and Decision Making}, vol.~21, pp. 1--14, 2021.

\bibitem{info15010002}
\BIBentryALTinterwordspacing
M.~Bouazizi, C.~Zheng, S.~Yang, and T.~Ohtsuki, ``Dementia detection from speech: What if language models are not the answer?'' \emph{Information}, vol.~15, no.~1, 2024. [Online]. Available: \url{https://www.mdpi.com/2078-2489/15/1/2}
\BIBentrySTDinterwordspacing

\bibitem{9870792}
C.~Zheng, M.~Bouazizi, and T.~Ohtsuki, ``An evaluation on information composition in dementia detection based on speech,'' \emph{IEEE Access}, vol.~10, pp. 92\,294--92\,306, 2022.

\bibitem{pan2019automatic}
Y.~Pan, B.~Mirheidari, M.~Reuber, A.~Venneri, D.~Blackburn, and H.~Christensen, ``Automatic hierarchical attention neural network for detecting ad,'' in \emph{Proceedings of Interspeech 2019}.\hskip 1em plus 0.5em minus 0.4em\relax International Speech Communication Association (ISCA), 2019, pp. 4105--4109.

\bibitem{rumelhart1986learning}
D.~E. Rumelhart, G.~E. Hinton, and R.~J. Williams, ``Learning internal representations by error propagation, parallel distributed processing, explorations in the microstructure of cognition, ed. de rumelhart and j. mcclelland. vol. 1. 1986,'' \emph{Biometrika}, vol.~71, no. 599-607, p.~6, 1986.

\bibitem{hochreiter1997long}
S.~Hochreiter and J.~Schmidhuber, ``Long short-term memory,'' \emph{Neural computation}, vol.~9, no.~8, pp. 1735--1780, 1997.

\bibitem{van2020review}
G.~Van~Houdt, C.~Mosquera, and G.~N{\'a}poles, ``A review on the long short-term memory model,'' \emph{Artificial Intelligence Review}, vol.~53, no.~8, pp. 5929--5955, 2020.

\bibitem{cho2014learning}
K.~Cho, B.~Van~Merri{\"e}nboer, C.~Gulcehre, D.~Bahdanau, F.~Bougares, H.~Schwenk, and Y.~Bengio, ``Learning phrase representations using rnn encoder-decoder for statistical machine translation,'' \emph{arXiv preprint arXiv:1406.1078}, 2014.

\bibitem{chung2014empirical}
J.~Chung, C.~Gulcehre, K.~Cho, and Y.~Bengio, ``Empirical evaluation of gated recurrent neural networks on sequence modeling,'' \emph{arXiv preprint arXiv:1412.3555}, 2014.

\bibitem{liu2019bidirectional}
G.~Liu and J.~Guo, ``Bidirectional lstm with attention mechanism and convolutional layer for text classification,'' \emph{Neurocomputing}, vol. 337, pp. 325--338, 2019.

\bibitem{zhou2015c}
C.~Zhou, C.~Sun, Z.~Liu, and F.~Lau, ``A c-lstm neural network for text classification,'' \emph{arXiv preprint arXiv:1511.08630}, 2015.

\bibitem{basiri2021abcdm}
M.~E. Basiri, S.~Nemati, M.~Abdar, E.~Cambria, and U.~R. Acharya, ``Abcdm: An attention-based bidirectional cnn-rnn deep model for sentiment analysis,'' \emph{Future Generation Computer Systems}, vol. 115, pp. 279--294, 2021.

\bibitem{DBLP:journals/corr/Kim14f}
\BIBentryALTinterwordspacing
Y.~Kim, ``Convolutional neural networks for sentence classification,'' \emph{CoRR}, vol. abs/1408.5882, 2014. [Online]. Available: \url{http://arxiv.org/abs/1408.5882}
\BIBentrySTDinterwordspacing

\bibitem{zhang2015sensitivity}
Y.~Zhang and B.~Wallace, ``A sensitivity analysis of (and practitioners' guide to) convolutional neural networks for sentence classification,'' \emph{arXiv preprint arXiv:1510.03820}, 2015.

\bibitem{liu2019robertarobustlyoptimizedbert}
\BIBentryALTinterwordspacing
Y.~Liu, M.~Ott, N.~Goyal, J.~Du, M.~Joshi, D.~Chen, O.~Levy, M.~Lewis, L.~Zettlemoyer, and V.~Stoyanov, ``Roberta: A robustly optimized bert pretraining approach,'' 2019. [Online]. Available: \url{https://arxiv.org/abs/1907.11692}
\BIBentrySTDinterwordspacing

\bibitem{fritsch2019automatic}
J.~Fritsch, S.~Wankerl, and E.~N{\"o}th, ``Automatic diagnosis of alzheimer’s disease using neural network language models,'' in \emph{ICASSP 2019-2019 IEEE International Conference on Acoustics, Speech and Signal Processing (ICASSP)}.\hskip 1em plus 0.5em minus 0.4em\relax IEEE, 2019, pp. 5841--5845.

\bibitem{DBLP:journals/corr/abs-2005-14165}
\BIBentryALTinterwordspacing
T.~B. Brown, B.~Mann, N.~Ryder, M.~Subbiah, J.~Kaplan, P.~Dhariwal, A.~Neelakantan, P.~Shyam, G.~Sastry, A.~Askell, S.~Agarwal, A.~Herbert{-}Voss, G.~Krueger, T.~Henighan, R.~Child, A.~Ramesh, D.~M. Ziegler, J.~Wu, C.~Winter, C.~Hesse, M.~Chen, E.~Sigler, M.~Litwin, S.~Gray, B.~Chess, J.~Clark, C.~Berner, S.~McCandlish, A.~Radford, I.~Sutskever, and D.~Amodei, ``Language models are few-shot learners,'' \emph{CoRR}, vol. abs/2005.14165, 2020. [Online]. Available: \url{https://arxiv.org/abs/2005.14165}
\BIBentrySTDinterwordspacing

\bibitem{achiam2023gpt}
J.~Achiam, S.~Adler, S.~Agarwal, L.~Ahmad, I.~Akkaya, F.~L. Aleman, D.~Almeida, J.~Altenschmidt, S.~Altman, S.~Anadkat \emph{et~al.}, ``Gpt-4 technical report,'' \emph{arXiv preprint arXiv:2303.08774}, 2023.

\bibitem{liu2019text}
Y.~Liu and M.~Lapata, ``Text summarization with pretrained encoders,'' \emph{arXiv preprint arXiv:1908.08345}, 2019.

\bibitem{DBLP:journals/corr/abs-1910-13461}
\BIBentryALTinterwordspacing
M.~Lewis, Y.~Liu, N.~Goyal, M.~Ghazvininejad, A.~Mohamed, O.~Levy, V.~Stoyanov, and L.~Zettlemoyer, ``{BART:} denoising sequence-to-sequence pre-training for natural language generation, translation, and comprehension,'' \emph{CoRR}, vol. abs/1910.13461, 2019. [Online]. Available: \url{http://arxiv.org/abs/1910.13461}
\BIBentrySTDinterwordspacing

\bibitem{DBLP:journals/corr/abs-1910-10683}
\BIBentryALTinterwordspacing
C.~Raffel, N.~Shazeer, A.~Roberts, K.~Lee, S.~Narang, M.~Matena, Y.~Zhou, W.~Li, and P.~J. Liu, ``Exploring the limits of transfer learning with a unified text-to-text transformer,'' \emph{CoRR}, vol. abs/1910.10683, 2019. [Online]. Available: \url{http://arxiv.org/abs/1910.10683}
\BIBentrySTDinterwordspacing

\bibitem{DBLP:journals/corr/abs-2010-11934}
\BIBentryALTinterwordspacing
L.~Xue, N.~Constant, A.~Roberts, M.~Kale, R.~Al{-}Rfou, A.~Siddhant, A.~Barua, and C.~Raffel, ``mt5: {A} massively multilingual pre-trained text-to-text transformer,'' \emph{CoRR}, vol. abs/2010.11934, 2020. [Online]. Available: \url{https://arxiv.org/abs/2010.11934}
\BIBentrySTDinterwordspacing

\bibitem{DBLP:journals/corr/abs-1909-11942}
\BIBentryALTinterwordspacing
Z.~Lan, M.~Chen, S.~Goodman, K.~Gimpel, P.~Sharma, and R.~Soricut, ``{ALBERT:} {A} lite {BERT} for self-supervised learning of language representations,'' \emph{CoRR}, vol. abs/1909.11942, 2019. [Online]. Available: \url{http://arxiv.org/abs/1909.11942}
\BIBentrySTDinterwordspacing

\bibitem{DBLP:journals/corr/abs-1906-08237}
\BIBentryALTinterwordspacing
Z.~Yang, Z.~Dai, Y.~Yang, J.~G. Carbonell, R.~Salakhutdinov, and Q.~V. Le, ``Xlnet: Generalized autoregressive pretraining for language understanding,'' \emph{CoRR}, vol. abs/1906.08237, 2019. [Online]. Available: \url{http://arxiv.org/abs/1906.08237}
\BIBentrySTDinterwordspacing

\bibitem{DBLP:journals/corr/abs-2004-05150}
\BIBentryALTinterwordspacing
I.~Beltagy, M.~E. Peters, and A.~Cohan, ``Longformer: The long-document transformer,'' \emph{CoRR}, vol. abs/2004.05150, 2020. [Online]. Available: \url{https://arxiv.org/abs/2004.05150}
\BIBentrySTDinterwordspacing

\bibitem{DBLP:journals/corr/abs-1901-08746}
\BIBentryALTinterwordspacing
J.~Lee, W.~Yoon, S.~Kim, D.~Kim, S.~Kim, C.~H. So, and J.~Kang, ``Biobert: a pre-trained biomedical language representation model for biomedical text mining,'' \emph{CoRR}, vol. abs/1901.08746, 2019. [Online]. Available: \url{http://arxiv.org/abs/1901.08746}
\BIBentrySTDinterwordspacing

\bibitem{DBLP:journals/corr/abs-1904-03323}
\BIBentryALTinterwordspacing
E.~Alsentzer, J.~R. Murphy, W.~Boag, W.~Weng, D.~Jin, T.~Naumann, and M.~B.~A. McDermott, ``Publicly available clinical {BERT} embeddings,'' \emph{CoRR}, vol. abs/1904.03323, 2019. [Online]. Available: \url{http://arxiv.org/abs/1904.03323}
\BIBentrySTDinterwordspacing

\bibitem{DBLP:journals/corr/abs-2008-02496}
\BIBentryALTinterwordspacing
Z.~Jiang, W.~Yu, D.~Zhou, Y.~Chen, J.~Feng, and S.~Yan, ``Convbert: Improving {BERT} with span-based dynamic convolution,'' \emph{CoRR}, vol. abs/2008.02496, 2020. [Online]. Available: \url{https://arxiv.org/abs/2008.02496}
\BIBentrySTDinterwordspacing

\bibitem{Yancheva2015UsingLF}
M.~Yancheva, K.~C. Fraser, and F.~Rudzicz, ``Using linguistic features longitudinally to predict clinical scores for alzheimer’s disease and related dementias,'' in \emph{Proceedings of SLPAT 2015: 6th workshop on speech and language processing for assistive technologies}, 2015, pp. 134--139.

\bibitem{Zhu2018DetectingCI}
\BIBentryALTinterwordspacing
Z.~Zhu, J.~Novikova, and F.~Rudzicz, ``Detecting cognitive impairments by agreeing on interpretations of linguistic features,'' \emph{ArXiv}, vol. abs/1808.06570, 2018. [Online]. Available: \url{https://api.semanticscholar.org/CorpusID:51938927}
\BIBentrySTDinterwordspacing

\bibitem{Mota2012SpeechGP}
\BIBentryALTinterwordspacing
N.~B. Mota, N.~Vasconcelos, N.~Lemos, A.~C. de~Souza~Pieretti, O.~Kinouchi, G.~A. Cecchi, M.~Copelli, and S.~Ribeiro, ``Speech graphs provide a quantitative measure of thought disorder in psychosis,'' \emph{PLoS ONE}, vol.~7, 2012. [Online]. Available: \url{https://api.semanticscholar.org/CorpusID:9506186}
\BIBentrySTDinterwordspacing

\bibitem{Warriner2013NormsOV}
\BIBentryALTinterwordspacing
A.~B. Warriner, V.~Kuperman, and M.~Brysbaert, ``Norms of valence, arousal, and dominance for 13,915 english lemmas,'' \emph{Behavior Research Methods}, vol.~45, pp. 1191 -- 1207, 2013. [Online]. Available: \url{https://api.semanticscholar.org/CorpusID:16918336}
\BIBentrySTDinterwordspacing

\bibitem{ai2010web}
H.~Ai and X.~Lu, ``A web-based system for automatic measurement of lexical complexity,'' in \emph{27th Annual Symposium of the Computer-Assisted Language Consortium (CALICO-10). Amherst, MA. June}, 2010, pp. 8--12.

\bibitem{Croisile1996ComparativeSO}
\BIBentryALTinterwordspacing
B.~Croisile, B.~Ska, M.-J. Brabant, A.~Duch{\^e}ne, Y.~Lepage, G.~Aimard, and M.~Trillet, ``Comparative study of oral and written picture description in patients with alzheimer's disease,'' \emph{Brain and Language}, vol.~53, pp. 1--19, 1996. [Online]. Available: \url{https://api.semanticscholar.org/CorpusID:36544389}
\BIBentrySTDinterwordspacing

\bibitem{pennington-etal-2014-glove}
\BIBentryALTinterwordspacing
J.~Pennington, R.~Socher, and C.~Manning, ``{G}lo{V}e: Global vectors for word representation,'' in \emph{Proceedings of the 2014 Conference on Empirical Methods in Natural Language Processing ({EMNLP})}, A.~Moschitti, B.~Pang, and W.~Daelemans, Eds.\hskip 1em plus 0.5em minus 0.4em\relax Doha, Qatar: Association for Computational Linguistics, Oct. 2014, pp. 1532--1543. [Online]. Available: \url{https://aclanthology.org/D14-1162}
\BIBentrySTDinterwordspacing

\bibitem{DBLP:journals/corr/abs-1901-02860}
\BIBentryALTinterwordspacing
Z.~Dai, Z.~Yang, Y.~Yang, J.~G. Carbonell, Q.~V. Le, and R.~Salakhutdinov, ``Transformer-xl: Attentive language models beyond a fixed-length context,'' \emph{CoRR}, vol. abs/1901.02860, 2019. [Online]. Available: \url{http://arxiv.org/abs/1901.02860}
\BIBentrySTDinterwordspacing

\bibitem{Radford2018ImprovingLU}
\BIBentryALTinterwordspacing
A.~Radford and K.~Narasimhan, ``Improving language understanding by generative pre-training,'' 2018. [Online]. Available: \url{https://api.semanticscholar.org/CorpusID:49313245}
\BIBentrySTDinterwordspacing

\bibitem{fraser2016linguistic}
K.~C. Fraser, J.~A. Meltzer, and F.~Rudzicz, ``Linguistic features identify alzheimer’s disease in narrative speech,'' \emph{Journal of Alzheimer's Disease}, vol.~49, no.~2, pp. 407--422, 2016.

\bibitem{di-palo-parde-2019-enriching}
\BIBentryALTinterwordspacing
F.~Di~Palo and N.~Parde, ``Enriching neural models with targeted features for dementia detection,'' in \emph{Proceedings of the 57th Annual Meeting of the Association for Computational Linguistics: Student Research Workshop}, F.~Alva-Manchego, E.~Choi, and D.~Khashabi, Eds.\hskip 1em plus 0.5em minus 0.4em\relax Florence, Italy: Association for Computational Linguistics, Jul. 2019, pp. 302--308. [Online]. Available: \url{https://aclanthology.org/P19-2042}
\BIBentrySTDinterwordspacing

\bibitem{DBLP:journals/corr/abs-1907-12412}
\BIBentryALTinterwordspacing
Y.~Sun, S.~Wang, Y.~Li, S.~Feng, H.~Tian, H.~Wu, and H.~Wang, ``{ERNIE} 2.0: {A} continual pre-training framework for language understanding,'' \emph{CoRR}, vol. abs/1907.12412, 2019. [Online]. Available: \url{http://arxiv.org/abs/1907.12412}
\BIBentrySTDinterwordspacing

\bibitem{bird2009natural}
S.~Bird, E.~Klein, and E.~Loper, \emph{Natural language processing with Python: analyzing text with the natural language toolkit}.\hskip 1em plus 0.5em minus 0.4em\relax " O'Reilly Media, Inc.", 2009.

\bibitem{garrard2005effects}
P.~Garrard, L.~M. Maloney, J.~R. Hodges, and K.~Patterson, ``The effects of very early alzheimer's disease on the characteristics of writing by a renowned author,'' \emph{Brain}, vol. 128, no.~2, pp. 250--260, 2005.

\bibitem{berisha2015tracking}
V.~Berisha, S.~Wang, A.~LaCross, and J.~Liss, ``Tracking discourse complexity preceding alzheimer's disease diagnosis: A case study comparing the press conferences of presidents ronald reagan and george herbert walker bush,'' \emph{Journal of Alzheimer's Disease}, vol.~45, no.~3, pp. 959--963, 2015.

\bibitem{bucks2000analysis}
R.~S. Bucks, S.~Singh, J.~M. Cuerden, and G.~K. Wilcock, ``Analysis of spontaneous, conversational speech in dementia of alzheimer type: Evaluation of an objective technique for analysing lexical performance,'' \emph{Aphasiology}, vol.~14, no.~1, pp. 71--91, 2000.

\bibitem{mikolov2013efficient}
T.~Mikolov, K.~Chen, G.~Corrado, and J.~Dean, ``Efficient estimation of word representations in vector space,'' \emph{arXiv preprint arXiv:1301.3781}, 2013.

\bibitem{cañete2023spanishpretrainedbertmodel}
\BIBentryALTinterwordspacing
J.~Cañete, G.~Chaperon, R.~Fuentes, J.-H. Ho, H.~Kang, and J.~Pérez, ``Spanish pre-trained bert model and evaluation data,'' 2023. [Online]. Available: \url{https://arxiv.org/abs/2308.02976}
\BIBentrySTDinterwordspacing

\bibitem{thirunavukarasu2023large}
A.~J. Thirunavukarasu, D.~S.~J. Ting, K.~Elangovan, L.~Gutierrez, T.~F. Tan, and D.~S.~W. Ting, ``Large language models in medicine,'' \emph{Nature medicine}, vol.~29, no.~8, pp. 1930--1940, 2023.

\bibitem{zhao2023survey}
W.~X. Zhao, K.~Zhou, J.~Li, T.~Tang, X.~Wang, Y.~Hou, Y.~Min, B.~Zhang, J.~Zhang, Z.~Dong \emph{et~al.}, ``A survey of large language models,'' \emph{arXiv preprint arXiv:2303.18223}, 2023.

\bibitem{chang2024survey}
Y.~Chang, X.~Wang, J.~Wang, Y.~Wu, L.~Yang, K.~Zhu, H.~Chen, X.~Yi, C.~Wang, Y.~Wang \emph{et~al.}, ``A survey on evaluation of large language models,'' \emph{ACM Transactions on Intelligent Systems and Technology}, vol.~15, no.~3, pp. 1--45, 2024.

\bibitem{openai2024chatgpt}
\BIBentryALTinterwordspacing
OpenAI, ``Chatgpt: Openai’s gpt-4 language model,'' 2024. [Online]. Available: \url{https://www.openai.com/chatgpt}
\BIBentrySTDinterwordspacing

\bibitem{DBLP:journals/corr/LeM14}
\BIBentryALTinterwordspacing
Q.~V. Le and T.~Mikolov, ``Distributed representations of sentences and documents,'' \emph{CoRR}, vol. abs/1405.4053, 2014. [Online]. Available: \url{http://arxiv.org/abs/1405.4053}
\BIBentrySTDinterwordspacing

\bibitem{reimers2019sentence}
N.~Reimers and I.~Gurevych, ``Sentence-bert: Sentence embeddings using siamese bert-networks,'' \emph{arXiv preprint arXiv:1908.10084}, 2019.

\bibitem{joulin2016bag}
A.~Joulin, E.~Grave, P.~Bojanowski, and T.~Mikolov, ``Bag of tricks for efficient text classification,'' \emph{arXiv preprint arXiv:1607.01759}, 2016.

\bibitem{akbik2018contextual}
A.~Akbik, D.~Blythe, and R.~Vollgraf, ``Contextual string embeddings for sequence labeling,'' in \emph{Proceedings of the 27th international conference on computational linguistics}, 2018, pp. 1638--1649.

\bibitem{9507270}
N.~D. Cilia, T.~D’Alessandro, C.~De~Stefano, F.~Fontanella, and M.~Molinara, ``From online handwriting to synthetic images for alzheimer's disease detection using a deep transfer learning approach,'' \emph{IEEE Journal of Biomedical and Health Informatics}, vol.~25, no.~12, pp. 4243--4254, 2021.

\bibitem{SUN2024121929}
\BIBentryALTinterwordspacing
J.~Sun, H.~H. Dodge, and M.~H. Mahoor, ``Mc-vivit: Multi-branch classifier-vivit to detect mild cognitive impairment in older adults using facial videos,'' \emph{Expert Systems with Applications}, vol. 238, p. 121929, 2024. [Online]. Available: \url{https://www.sciencedirect.com/science/article/pii/S0957417423024314}
\BIBentrySTDinterwordspacing

\bibitem{10.1007/978-3-030-66096-3_20}
X.~Liang, A.~Angelopoulou, E.~Kapetanios, B.~Woll, R.~Al~Batat, and T.~Woolfe, ``A multi-modal machine learning approach and toolkit to automate recognition of early stages of dementia among british sign language users,'' in \emph{Computer Vision -- ECCV 2020 Workshops}, A.~Bartoli and A.~Fusiello, Eds.\hskip 1em plus 0.5em minus 0.4em\relax Cham: Springer International Publishing, 2020, pp. 278--293.

\bibitem{hu2019graph}
K.~Hu, Z.~Wang, W.~Wang, K.~A.~E. Martens, L.~Wang, T.~Tan, S.~J. Lewis, and D.~D. Feng, ``Graph sequence recurrent neural network for vision-based freezing of gait detection,'' \emph{IEEE Transactions on Image Processing}, vol.~29, pp. 1890--1901, 2019.

\bibitem{9438658}
A.~Das, X.~Niu, A.~Dantcheva, S.~L. Happy, H.~Han, R.~Zeghari, P.~Robert, S.~Shan, F.~Bremond, and X.~Chen, ``A spatio-temporal approach for apathy classification,'' \emph{IEEE Transactions on Circuits and Systems for Video Technology}, vol.~32, no.~5, pp. 2561--2573, 2022.

\bibitem{vessio2019dynamic}
G.~Vessio, ``Dynamic handwriting analysis for neurodegenerative disease assessment: a literary review,'' \emph{Applied Sciences}, vol.~9, no.~21, p. 4666, 2019.

\bibitem{onofri2013dysgraphia}
E.~Onofri, M.~Mercuri, M.~Salesi, S.~Ferrara, G.~M. Troili, C.~Simeone, M.~R. Ricciardi, S.~Ricci, and T.~Archer, ``Dysgraphia in relation to cognitive performance in patients with alzheimer’s disease,'' \emph{Journal of Intellectual Disability-Diagnosis and Treatment}, vol.~1, no.~2, pp. 113--124, 2013.

\bibitem{muller2017diagnostic}
S.~M{\"u}ller, O.~Preische, P.~Heymann, U.~Elbing, and C.~Laske, ``Diagnostic value of a tablet-based drawing task for discrimination of patients in the early course of alzheimer’s disease from healthy individuals,'' \emph{Journal of Alzheimer's Disease}, vol.~55, no.~4, pp. 1463--1469, 2017.

\bibitem{szegedy2016rethinking}
C.~Szegedy, V.~Vanhoucke, S.~Ioffe, J.~Shlens, and Z.~Wojna, ``Rethinking the inception architecture for computer vision,'' in \emph{Proceedings of the IEEE conference on computer vision and pattern recognition}, 2016, pp. 2818--2826.

\bibitem{CILIA2022104822}
\BIBentryALTinterwordspacing
N.~D. Cilia, G.~{De Gregorio}, C.~{De Stefano}, F.~Fontanella, A.~Marcelli, and A.~Parziale, ``Diagnosing alzheimer’s disease from on-line handwriting: A novel dataset and performance benchmarking,'' \emph{Engineering Applications of Artificial Intelligence}, vol. 111, p. 104822, 2022. [Online]. Available: \url{https://www.sciencedirect.com/science/article/pii/S0952197622000902}
\BIBentrySTDinterwordspacing

\bibitem{10.1007/978-3-031-37660-3_44}
N.~D. Cilia, T.~D'Alessandro, C.~De~Stefano, F.~Fontanella, and E.~Nardone, ``Predicting alzheimer's disease: A stroke-based handwriting analysis approach based on machine learning,'' in \emph{Pattern Recognition, Computer Vision, and Image Processing. ICPR 2022 International Workshops and Challenges}, J.-J. Rousseau and B.~Kapralos, Eds.\hskip 1em plus 0.5em minus 0.4em\relax Cham: Springer Nature Switzerland, 2023, pp. 632--643.

\bibitem{CILIA2024107891}
\BIBentryALTinterwordspacing
N.~D. Cilia, C.~{De Stefano}, F.~Fontanella, and S.~M. Siniscalchi, ``How word semantics and phonology affect handwriting of alzheimer’s patients: A machine learning based analysis,'' \emph{Computers in Biology and Medicine}, vol. 169, p. 107891, 2024. [Online]. Available: \url{https://www.sciencedirect.com/science/article/pii/S0010482523013562}
\BIBentrySTDinterwordspacing

\bibitem{nam2020analyzing}
U.~Nam, K.~Lee, H.~Ko, J.-Y. Lee, and E.~C. Lee, ``Analyzing facial and eye movements to screen for alzheimer’s disease,'' \emph{Sensors}, vol.~20, no.~18, p. 5349, 2020.

\bibitem{dourado2019facial}
M.~C.~N. Dourado, B.~Torres Mendon{\c{c}}a~de Melo~F{\'a}del, J.~P. Sim{\~o}es~Neto, G.~Alves, and C.~Alves, ``Facial expression recognition patterns in mild and moderate alzheimer’s disease,'' \emph{Journal of Alzheimer's Disease}, vol.~69, no.~2, pp. 539--549, 2019.

\bibitem{jin2020diagnosing}
B.~Jin, Y.~Qu, L.~Zhang, and Z.~Gao, ``Diagnosing parkinson disease through facial expression recognition: video analysis,'' \emph{Journal of medical Internet research}, vol.~22, no.~7, p. e18697, 2020.

\bibitem{tanaka2019detecting}
H.~Tanaka, H.~Adachi, H.~Kazui, M.~Ikeda, T.~Kudo, and S.~Nakamura, ``Detecting dementia from face in human-agent interaction,'' in \emph{Adjunct of the 2019 international conference on multimodal interaction}, 2019, pp. 1--6.

\bibitem{DBLP:journals/corr/abs-2106-13230}
\BIBentryALTinterwordspacing
Z.~Liu, J.~Ning, Y.~Cao, Y.~Wei, Z.~Zhang, S.~Lin, and H.~Hu, ``Video swin transformer,'' \emph{CoRR}, vol. abs/2106.13230, 2021. [Online]. Available: \url{https://arxiv.org/abs/2106.13230}
\BIBentrySTDinterwordspacing

\bibitem{hely2008sydney}
M.~A. Hely, W.~G. Reid, M.~A. Adena, G.~M. Halliday, and J.~G. Morris, ``The sydney multicenter study of parkinson's disease: the inevitability of dementia at 20 years,'' \emph{Movement disorders}, vol.~23, no.~6, pp. 837--844, 2008.

\bibitem{macht2007predictors}
M.~Macht, Y.~Kaussner, J.~C. M{\"o}ller, K.~Stiasny-Kolster, K.~M. Eggert, H.-P. Kr{\"u}ger, and H.~Ellgring, ``Predictors of freezing in parkinson's disease: a survey of 6,620 patients,'' \emph{Movement disorders}, vol.~22, no.~7, pp. 953--956, 2007.

\bibitem{bloem2004falls}
B.~R. Bloem, J.~M. Hausdorff, J.~E. Visser, and N.~Giladi, ``Falls and freezing of gait in parkinson's disease: a review of two interconnected, episodic phenomena,'' \emph{Movement disorders: official journal of the Movement Disorder Society}, vol.~19, no.~8, pp. 871--884, 2004.

\bibitem{lewis2009pathophysiological}
S.~J. Lewis and R.~A. Barker, ``A pathophysiological model of freezing of gait in parkinson's disease,'' \emph{Parkinsonism \& related disorders}, vol.~15, no.~5, pp. 333--338, 2009.

\bibitem{khan2013motion}
T.~Khan, J.~Westin, and M.~Dougherty, ``Motion cue analysis for parkinsonian gait recognition,'' \emph{The open biomedical engineering journal}, vol.~7, p.~1, 2013.

\bibitem{nieto2016vision}
M.~Nieto-Hidalgo, F.~J. Ferr{\'a}ndez-Pastor, R.~J. Valdivieso-Sarabia, J.~Mora-Pascual, and J.~M. Garc{\'\i}a-Chamizo, ``A vision based proposal for classification of normal and abnormal gait using rgb camera,'' \emph{Journal of biomedical informatics}, vol.~63, pp. 82--89, 2016.

\bibitem{molitor2015eye}
R.~J. Molitor, P.~C. Ko, and B.~A. Ally, ``Eye movements in alzheimer's disease,'' \emph{Journal of Alzheimer's disease}, vol.~44, no.~1, pp. 1--12, 2015.

\bibitem{oyama2019novel}
A.~Oyama, S.~Takeda, Y.~Ito, T.~Nakajima, Y.~Takami, Y.~Takeya, K.~Yamamoto, K.~Sugimoto, H.~Shimizu, M.~Shimamura \emph{et~al.}, ``Novel method for rapid assessment of cognitive impairment using high-performance eye-tracking technology,'' \emph{Scientific reports}, vol.~9, no.~1, p. 12932, 2019.

\bibitem{lagun2011detecting}
D.~Lagun, C.~Manzanares, S.~M. Zola, E.~A. Buffalo, and E.~Agichtein, ``Detecting cognitive impairment by eye movement analysis using automatic classification algorithms,'' \emph{Journal of neuroscience methods}, vol. 201, no.~1, pp. 196--203, 2011.

\bibitem{mengoudi2020augmenting}
K.~Mengoudi, D.~Ravi, K.~X. Yong, S.~Primativo, I.~M. Pavisic, E.~Brotherhood, K.~Lu, J.~M. Schott, S.~J. Crutch, and D.~C. Alexander, ``Augmenting dementia cognitive assessment with instruction-less eye-tracking tests,'' \emph{IEEE journal of biomedical and health informatics}, vol.~24, no.~11, pp. 3066--3075, 2020.

\bibitem{lazarus1976multimodal}
A.~A. Lazarus \emph{et~al.}, ``Multimodal behavior therapy: I.'' 1976.

\bibitem{xu2023multimodal}
P.~Xu, X.~Zhu, and D.~A. Clifton, ``Multimodal learning with transformers: A survey,'' \emph{IEEE Transactions on Pattern Analysis and Machine Intelligence}, vol.~45, no.~10, pp. 12\,113--12\,132, 2023.

\bibitem{guo2019deep}
W.~Guo, J.~Wang, and S.~Wang, ``Deep multimodal representation learning: A survey,'' \emph{Ieee Access}, vol.~7, pp. 63\,373--63\,394, 2019.

\bibitem{baltruvsaitis2018multimodal}
T.~Baltru{\v{s}}aitis, C.~Ahuja, and L.-P. Morency, ``Multimodal machine learning: A survey and taxonomy,'' \emph{IEEE transactions on pattern analysis and machine intelligence}, vol.~41, no.~2, pp. 423--443, 2018.

\bibitem{altinok2024explainable}
D.~Altinok, ``Explainable multimodal fusion for dementia detection from text and speech,'' in \emph{International Conference on Text, Speech, and Dialogue}.\hskip 1em plus 0.5em minus 0.4em\relax Springer, 2024, pp. 236--251.

\bibitem{10.3389/fnagi.2022.830943}
\BIBentryALTinterwordspacing
L.~Ilias and D.~Askounis, ``Multimodal deep learning models for detecting dementia from speech and transcripts,'' \emph{Frontiers in Aging Neuroscience}, vol.~14, 2022. [Online]. Available: \url{https://www.frontiersin.org/articles/10.3389/fnagi.2022.830943}
\BIBentrySTDinterwordspacing

\bibitem{9926818}
L.~Ilias, D.~Askounis, and J.~Psarras, ``A multimodal approach for dementia detection from spontaneous speech with tensor fusion layer,'' in \emph{2022 IEEE-EMBS International Conference on Biomedical and Health Informatics (BHI)}, 2022, pp. 1--5.

\bibitem{DBLP:journals/corr/abs-2103-00020}
\BIBentryALTinterwordspacing
A.~Radford, J.~W. Kim, C.~Hallacy, A.~Ramesh, G.~Goh, S.~Agarwal, G.~Sastry, A.~Askell, P.~Mishkin, J.~Clark, G.~Krueger, and I.~Sutskever, ``Learning transferable visual models from natural language supervision,'' \emph{CoRR}, vol. abs/2103.00020, 2021. [Online]. Available: \url{https://arxiv.org/abs/2103.00020}
\BIBentrySTDinterwordspacing

\bibitem{touvron2023llamaopenefficientfoundation}
\BIBentryALTinterwordspacing
H.~Touvron, T.~Lavril, G.~Izacard, X.~Martinet, M.-A. Lachaux, T.~Lacroix, B.~Rozière, N.~Goyal, E.~Hambro, F.~Azhar, A.~Rodriguez, A.~Joulin, E.~Grave, and G.~Lample, ``Llama: Open and efficient foundation language models,'' 2023. [Online]. Available: \url{https://arxiv.org/abs/2302.13971}
\BIBentrySTDinterwordspacing

\bibitem{li2023blip2bootstrappinglanguageimagepretraining}
\BIBentryALTinterwordspacing
J.~Li, D.~Li, S.~Savarese, and S.~Hoi, ``Blip-2: Bootstrapping language-image pre-training with frozen image encoders and large language models,'' 2023. [Online]. Available: \url{https://arxiv.org/abs/2301.12597}
\BIBentrySTDinterwordspacing

\bibitem{sun2019videobertjointmodelvideo}
\BIBentryALTinterwordspacing
C.~Sun, A.~Myers, C.~Vondrick, K.~Murphy, and C.~Schmid, ``Videobert: A joint model for video and language representation learning,'' 2019. [Online]. Available: \url{https://arxiv.org/abs/1904.01766}
\BIBentrySTDinterwordspacing

\bibitem{li2019visualbert}
L.~H. Li, M.~Yatskar, D.~Yin, C.-J. Hsieh, and K.-W. Chang, ``Visualbert: A simple and performant baseline for vision and language,'' \emph{arXiv preprint arXiv:1908.03557}, 2019.

\bibitem{DBLP:journals/corr/abs-1908-02265}
\BIBentryALTinterwordspacing
J.~Lu, D.~Batra, D.~Parikh, and S.~Lee, ``Vilbert: Pretraining task-agnostic visiolinguistic representations for vision-and-language tasks,'' \emph{CoRR}, vol. abs/1908.02265, 2019. [Online]. Available: \url{http://arxiv.org/abs/1908.02265}
\BIBentrySTDinterwordspacing

\bibitem{zheng2021fused}
R.~Zheng, J.~Chen, M.~Ma, and L.~Huang, ``Fused acoustic and text encoding for multimodal bilingual pretraining and speech translation,'' in \emph{International Conference on Machine Learning}.\hskip 1em plus 0.5em minus 0.4em\relax PMLR, 2021, pp. 12\,736--12\,746.

\bibitem{shi2022learning}
B.~Shi, W.-N. Hsu, K.~Lakhotia, and A.~Mohamed, ``Learning audio-visual speech representation by masked multimodal cluster prediction,'' \emph{arXiv preprint arXiv:2201.02184}, 2022.

\bibitem{guo2020graphcodebert}
D.~Guo, S.~Ren, S.~Lu, Z.~Feng, D.~Tang, S.~Liu, L.~Zhou, N.~Duan, A.~Svyatkovskiy, S.~Fu \emph{et~al.}, ``Graphcodebert: Pre-training code representations with data flow,'' \emph{arXiv preprint arXiv:2009.08366}, 2020.

\bibitem{9156959}
K.~Gavrilyuk, R.~Sanford, M.~Javan, and C.~G.~M. Snoek, ``{ Actor-Transformers for Group Activity Recognition },'' in \emph{2020 IEEE/CVF Conference on Computer Vision and Pattern Recognition (CVPR)}.\hskip 1em plus 0.5em minus 0.4em\relax Los Alamitos, CA, USA: IEEE Computer Society, Jun. 2020, pp. 836--845.

\bibitem{lin2021interbertvisionandlanguageinteractionmultimodal}
\BIBentryALTinterwordspacing
J.~Lin, A.~Yang, Y.~Zhang, J.~Liu, J.~Zhou, and H.~Yang, ``Interbert: Vision-and-language interaction for multi-modal pretraining,'' 2021. [Online]. Available: \url{https://arxiv.org/abs/2003.13198}
\BIBentrySTDinterwordspacing

\bibitem{tsai2019multimodaltransformerunalignedmultimodal}
\BIBentryALTinterwordspacing
Y.-H.~H. Tsai, S.~Bai, P.~P. Liang, J.~Z. Kolter, L.-P. Morency, and R.~Salakhutdinov, ``Multimodal transformer for unaligned multimodal language sequences,'' 2019. [Online]. Available: \url{https://arxiv.org/abs/1906.00295}
\BIBentrySTDinterwordspacing

\bibitem{xu2020cross}
X.~Xu, T.~Wang, Y.~Yang, L.~Zuo, F.~Shen, and H.~T. Shen, ``Cross-modal attention with semantic consistence for image--text matching,'' \emph{IEEE transactions on neural networks and learning systems}, vol.~31, no.~12, pp. 5412--5425, 2020.

\bibitem{9383618}
A.~Khare, S.~Parthasarathy, and S.~Sundaram, ``Self-supervised learning with cross-modal transformers for emotion recognition,'' in \emph{2021 IEEE Spoken Language Technology Workshop (SLT)}, 2021, pp. 381--388.

\bibitem{murahari2020largescalepretrainingvisualdialog}
\BIBentryALTinterwordspacing
V.~Murahari, D.~Batra, D.~Parikh, and A.~Das, ``Large-scale pretraining for visual dialog: A simple state-of-the-art baseline,'' 2020. [Online]. Available: \url{https://arxiv.org/abs/1912.02379}
\BIBentrySTDinterwordspacing

\bibitem{arevalo2020gated}
J.~Arevalo, T.~Solorio, M.~Montes-y Gomez, and F.~A. Gonz{\'a}lez, ``Gated multimodal networks,'' \emph{Neural Computing and Applications}, vol.~32, pp. 10\,209--10\,228, 2020.

\bibitem{ZHANG2021104042}
\BIBentryALTinterwordspacing
Y.~Zhang, D.~Sidibé, O.~Morel, and F.~Mériaudeau, ``Deep multimodal fusion for semantic image segmentation: A survey,'' \emph{Image and Vision Computing}, vol. 105, p. 104042, 2021. [Online]. Available: \url{https://www.sciencedirect.com/science/article/pii/S0262885620301748}
\BIBentrySTDinterwordspacing

\bibitem{10.1093/bib/bbab569}
\BIBentryALTinterwordspacing
S.~R. Stahlschmidt, B.~Ulfenborg, and J.~Synnergren, ``{Multimodal deep learning for biomedical data fusion: a review},'' \emph{Briefings in Bioinformatics}, vol.~23, no.~2, p. bbab569, 01 2022. [Online]. Available: \url{https://doi.org/10.1093/bib/bbab569}
\BIBentrySTDinterwordspacing

\bibitem{tsanas2012novel}
A.~Tsanas, M.~A. Little, P.~E. McSharry, J.~Spielman, and L.~O. Ramig, ``Novel speech signal processing algorithms for high-accuracy classification of parkinson's disease,'' \emph{IEEE transactions on biomedical engineering}, vol.~59, no.~5, pp. 1264--1271, 2012.

\bibitem{ALMEIDA201955}
\BIBentryALTinterwordspacing
J.~S. Almeida, P.~P. {Rebouças Filho}, T.~Carneiro, W.~Wei, R.~Damaševičius, R.~Maskeliūnas, and V.~H.~C. {de Albuquerque}, ``Detecting parkinson’s disease with sustained phonation and speech signals using machine learning techniques,'' \emph{Pattern Recognition Letters}, vol. 125, pp. 55--62, 2019. [Online]. Available: \url{https://www.sciencedirect.com/science/article/pii/S0167865519301163}
\BIBentrySTDinterwordspacing

\bibitem{GORRIZ2023101945}
\BIBentryALTinterwordspacing
J.~Górriz \emph{et~al.}, ``Computational approaches to explainable artificial intelligence: Advances in theory, applications and trends,'' \emph{Information Fusion}, vol. 100, p. 101945, 2023. [Online]. Available: \url{https://www.sciencedirect.com/science/article/pii/S1566253523002610}
\BIBentrySTDinterwordspacing

\bibitem{holzinger2019causability}
A.~Holzinger, G.~Langs, H.~Denk, K.~Zatloukal, and H.~M{\"u}ller, ``Causability and explainability of artificial intelligence in medicine,'' \emph{Wiley Interdisciplinary Reviews: Data Mining and Knowledge Discovery}, vol.~9, no.~4, p. e1312, 2019.

\bibitem{vilone2021notions}
G.~Vilone and L.~Longo, ``Notions of explainability and evaluation approaches for explainable artificial intelligence,'' \emph{Information Fusion}, vol.~76, pp. 89--106, 2021.

\bibitem{ribeiro2016should}
M.~T. Ribeiro, S.~Singh, and C.~Guestrin, ``"why should i trust you?" explaining the predictions of any classifier,'' in \emph{Proceedings of the 22nd ACM SIGKDD international conference on knowledge discovery and data mining}, 2016, pp. 1135--1144.

\bibitem{almor1999alzheimer}
A.~Almor, D.~Kempler, M.~C. MacDonald, E.~S. Andersen, and L.~K. Tyler, ``Why do alzheimer patients have difficulty with pronouns? working memory, semantics, and reference in comprehension and production in alzheimer's disease,'' \emph{Brain and language}, vol.~67, no.~3, pp. 202--227, 1999.

\bibitem{ortiz2025cognialign}
D.~Ortiz-Perez, M.~Benavent-Lledo, J.~Rodriguez-Juan, J.~Garcia-Rodriguez, and D.~Tomás, ``Cognialign: Word-level multimodal speech alignment with gated cross-attention for alzheimer’s detection,'' \emph{Knowledge-Based Systems}, vol. 329, p. 114264, 2025.

\bibitem{wang2021actionclipnewparadigmvideo}
\BIBentryALTinterwordspacing
M.~Wang, J.~Xing, and Y.~Liu, ``Actionclip: A new paradigm for video action recognition,'' 2021. [Online]. Available: \url{https://arxiv.org/abs/2109.08472}
\BIBentrySTDinterwordspacing

\bibitem{wu2023bidirectionalcrossmodalknowledgeexploration}
\BIBentryALTinterwordspacing
W.~Wu, X.~Wang, H.~Luo, J.~Wang, Y.~Yang, and W.~Ouyang, ``Bidirectional cross-modal knowledge exploration for video recognition with pre-trained vision-language models,'' 2023. [Online]. Available: \url{https://arxiv.org/abs/2301.00182}
\BIBentrySTDinterwordspacing

\bibitem{tao2018digital}
F.~Tao, H.~Zhang, A.~Liu, and A.~Y. Nee, ``Digital twin in industry: State-of-the-art,'' \emph{IEEE Transactions on industrial informatics}, vol.~15, no.~4, pp. 2405--2415, 2018.

\bibitem{bruynseels2018digital}
K.~Bruynseels, F.~Santoni~de Sio, and J.~Van~den Hoven, ``Digital twins in health care: ethical implications of an emerging engineering paradigm,'' \emph{Frontiers in genetics}, vol.~9, p.~31, 2018.

\end{thebibliography}

\end{document}